\documentclass[10pt,twocolumn,letterpaper]{article}

\usepackage{iccv}
\usepackage{times}
\usepackage{epsfig}
\usepackage{tikz}
\usepackage{amsmath}
\usepackage{amssymb}

\usepackage{ctable}

\usepackage[pagebackref=true,breaklinks=true,letterpaper=true,colorlinks,bookmarks=false]{hyperref}

\iccvfinalcopy 

\def\iccvPaperID{7364} 

\ificcvfinal\fi

\begin{document}

\title{Learning Signed Distance Field for Multi-view Surface Reconstruction}

\author{Jingyang Zhang \qquad Yao Yao \qquad Long Quan\\
The Hong Kong University of Science and Technology \\
{\tt\small \{jzhangbs,yyaoag,quan\}@cse.ust.hk}
}

\maketitle
\ificcvfinal\thispagestyle{empty}\fi

\newcommand{\fakepara}[1]{\noindent\textbf{#1}\space\space\space\space}

\begin{abstract}
	Recent works on implicit neural representations have shown promising results for multi-view surface reconstruction. 
	However, most approaches are limited to relatively simple geometries and usually require clean object masks for reconstructing complex and concave objects. 
	In this work, we introduce a novel neural surface reconstruction framework that leverages the knowledge of stereo matching and feature consistency to optimize the implicit surface representation. 
	More specifically, we apply a signed distance field (SDF) and a surface light field to represent the scene geometry and appearance respectively. The SDF is directly supervised by geometry from stereo matching, and is refined by optimizing the multi-view feature consistency and the fidelity of rendered images. 
	Our method is able to improve the robustness of geometry estimation and support reconstruction of complex scene topologies.
	Extensive experiments have been conducted on DTU, EPFL and Tanks and Temples datasets. Compared to previous state-of-the-art methods, our method achieves better mesh reconstruction in wide open scenes without masks as input.
\end{abstract}

\section{Introduction}

Surface reconstruction from calibrated multi-view images is one of the key problems in 3D computer vision. Traditionally,  surface reconstruction can be divided into two substeps: 1) depth maps and point clouds are reconstructed from images via multi-view stereo (MVS) algorithms \cite{campbell2008using,furukawa2009accurate,tola2012efficient,schonberger2016pixelwise,xu2019multi}; 2) a surface, usually represented as a triangular mesh, is extracted from dense points by maximizing the conformity to points \cite{labatut2007efficient,kazhdan2013screened,lorensen1987marching}. Optionally, a surface refinement step can be applied to recover geometry details through multi-view photo-consistency \cite{pons2007multi,vu2011high,delaunoy2011gradient,delaunoy2014photometric,furukawa2006carved}.  While this pipeline has been proven to be effective and robust in various scenarios, the reconstructed geometry may be suboptimal due to the accumulated loss in representation conversions from images to points, then to mesh.  For example,  errors introduced in point cloud reconstruction would pass to the surface reconstruction, causing wrong mesh topology and difficult to be recovered.  Although recent learning-based MVS  \cite{kar2017learning,huang2018deepmvs,xue2019mvscrf,paschalidou2018raynet,yao2018mvsnet} and mesh extraction \cite{Park_2019_CVPR,mescheder2019occupancy,sitzmann2019scene,peng2020convolutional} methods have been proposed to boost the reconstruction quality of each substep independently,  it is still desirable to reconstruct the optimal surface from images in an end-to-end manner.

\begin{figure}
	\centering
	\begin{tabular}{@{\hskip4pt}c@{\hskip4pt}@{\hskip4pt}c@{\hskip4pt}}
		\includegraphics[width=0.45\linewidth,trim={150 0 300 0},clip]{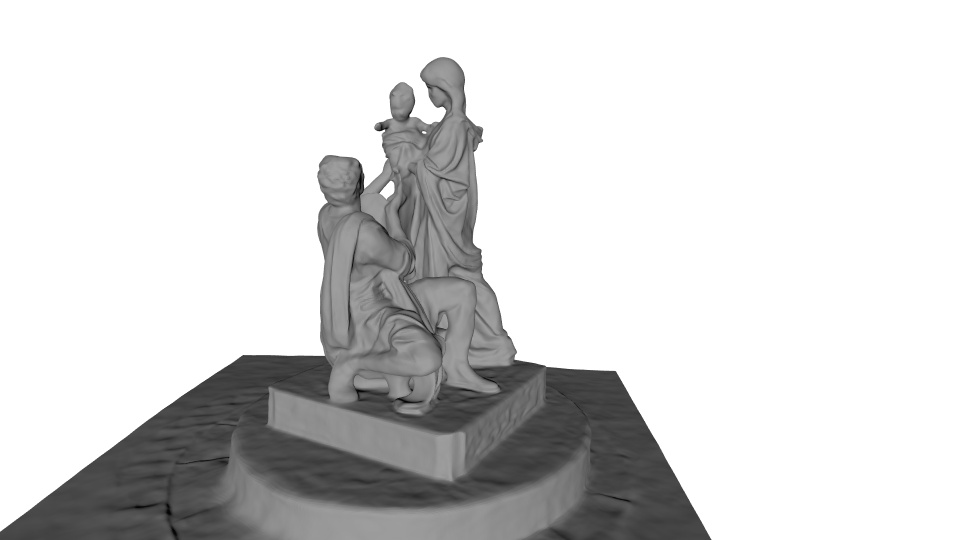} & 
		\includegraphics[width=0.45\linewidth,trim={100 0 200 0},clip]{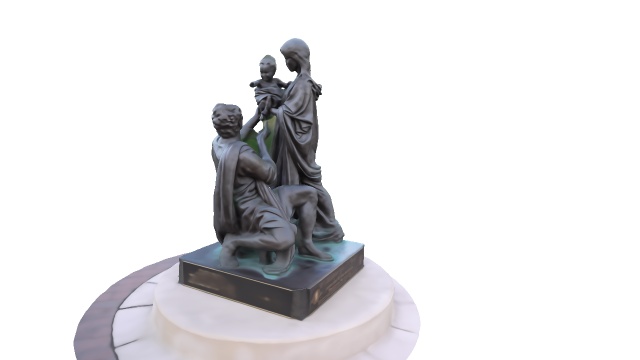} \\
		\includegraphics[width=0.45\linewidth,trim={180 0 210 0},clip]{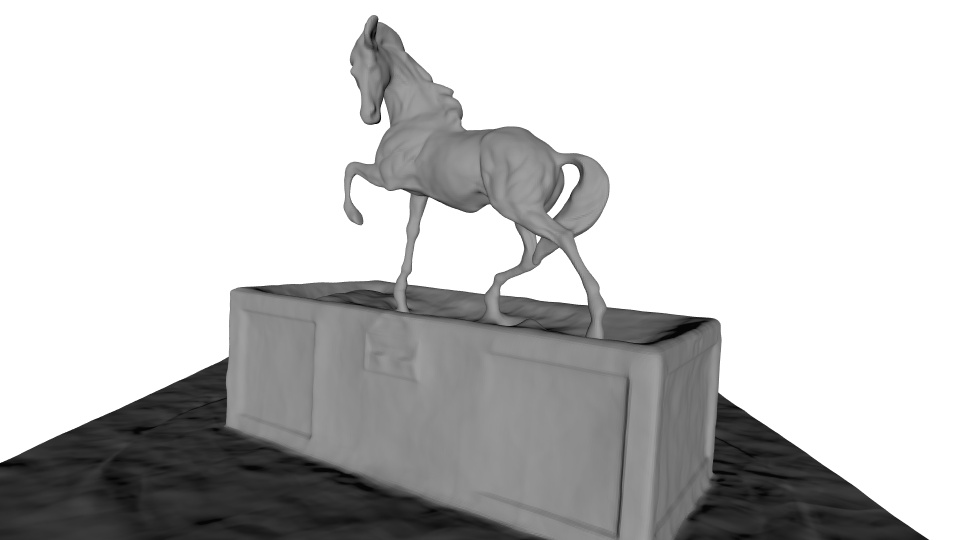} & 
		\includegraphics[width=0.45\linewidth,trim={120 0 140 0},clip]{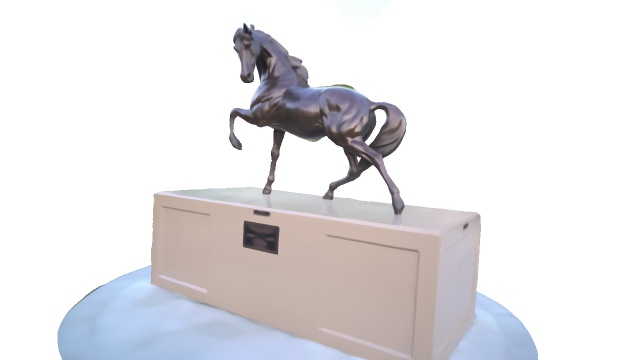} \\
		Mesh & Rendered Image
	\end{tabular}
	\caption{Our reconstructions on \textit{Family} and \textit{Horse} of the \textit{Tanks and Temples} dataset. The proposed MVSDF is able to reconstruct correct mesh topologies with fine details for highly texture-less and reflective surfaces. }
	\label{fig:teaser}
	\vspace{-3mm}
\end{figure}

Alternatively,  recent works on neural representations show that mesh surface can be directly constructed from images through implicit representations and differentiable rendering \cite{niemeyer2020differentiable,liu2019learning,saito2019pifu,liu2020dist,yariv2020multiview,sitzmann2019scene,mildenhall2020nerf}. Surface geometry and color information of the scene are usually represented as implicit functions,  which are directly modeled by multi-layer perceptrons (MLPs) in the network and optimized through differentiable rendering. The triangle mesh can be extracted from the implicit field via the Marching Cube algorithm \cite{lorensen1987marching,mescheder2019occupancy}.  Compared with classical meshing pipelines,  these methods are able to reconstruct the scene geometry in an end-to-end manner and generate synthesized images at the same time. 
However, as all scene parameters are jointly optimized at the same time, geometry is only a by-product of the entire differential rendering pipeline, and ambiguities exist in geometry and appearance\cite{zhang2020nerf++}.
To mitigate the problem, implicit differentiable renderer (IDR) \cite{yariv2020multiview} applies manually labeled object masks as input,  but it is not feasible for a large number of images and is sometimes not well defined for real-world image inputs. 

In this paper, we present MVSDF, a novel neural surface reconstruction framework that combines implicit neural surface estimation with recent advanced MVS networks. 
On the one hand, we follow the implicit differentiable renderer\cite{yariv2020multiview} to represent the surface as zero level set of a signed distance field (SDF) and the appearance as a surface light field, which are jointly optimized through render loss.
On the other hand, we introduce deep image features and depth maps from learning-based MVS \cite{yao2018mvsnet,zhang2020visibility,yu2020fast} to assist the implicit SDF estimation.  
The SDF is supervised by inferred depth values from the MVS network, and is further refined by maximizing the multi-view feature consistency at the surface points of the SDF.  
We find that the surface topology can be greatly improved with the guidance from MVS depth maps, and our method can be applied to complex geometries even without input object masks.
Also, compared to render loss in IDR, the multi-view feature consistency imposes a photometric constraint at an early stage of the differentiable rendering pipeline, which significantly improves the geometry accuracy and helps to preserve high-fidelity details in final reconstructions.

Our method has been evaluated on \textit{DTU} \cite{jensen2014large}, \textit{EPFL} \cite{strecha2008benchmarking} and \textit{Tanks and Temples} \cite{knapitsch2017tanks} datasets. 
We compare our method with classical meshing pipelines and recent differentiable rendering based networks on both mesh reconstruction and view synthesis quality.  Both quantitative and qualitative results demonstrate that our method is able to recover complex geometries even without object masks as input. 

\section{Related Works}

\fakepara{Multi-view Stereo}
Multi-view stereo \cite{seitz2006comparison,goesele2006multi,furukawa2015multi} is a well-developed approach to recover the dense representation of the scene from overlapping images. The fundamental principle of MVS is that a surface point should be visually consistent in all visible views. Traditionally, MVS evaluates the matching cost of image patches for all depth hypotheses and finds the one that best describes the input images. Depth hypotheses can be uniformly sampled from predefined camera frustum \cite{collins1996space} or propagated from neighboring pixels and adjacent views \cite{lhuillier2005quasi,furukawa2009accurate,bleyer2011patchmatch,galliani2015massively,schonberger2016pixelwise,xu2019multi,kuhn2020deepc}. Usually depth map outputs are fused into a unified point cloud, and further converted into a mesh surface \cite{labatut2007efficient,kazhdan2013screened,lorensen1987marching}.  However, such conversions may be lossy. For example, depth maps may be over-filtered so that holes would appear in the final reconstruction.  Moreover, surface details may be over-smoothed during the conversion from point cloud to mesh. 

Recently, deep learning techniques have been applied to multi-view stereo \cite{kar2017learning,huang2018deepmvs,yao2018mvsnet,yu2020fast,zhang2020visibility}. In the network, hand-crafted image features are converted into deep features and engineered cost regularizations are replaced by learned ones \cite{paschalidou2018raynet,yao2018mvsnet,yao2019recurrent,gu2020cascade,wang2020patchmatchnet}.  Although the overall depth quality is improved, it is still difficult to match pixels in texture-less or non-Lambertian regions. In this work, we aim to improve the reconstruction of texture-less regions by the interpolation capability of implicit functions, and non-Lambertian regions by explicit view-dependent appearance modeling. 

\fakepara{Implicit Neural Surface}
Implicit neural representations have also gained popularity in 3D scene reconstructions. Occupancy fields \cite{mescheder2019occupancy,peng2020convolutional} are proposed to model the object surface using the per-point occupancy information, while DeepSDF \cite{Park_2019_CVPR} applies a signed distance field to describe the 3D geometry of the scene. These methods usually take 3D point cloud as input and optimize corresponding implicit fields using ground truth labels, which can be viewed as learned mesh reconstruction from dense point cloud.  The marching cube algorithm \cite{lorensen1987marching} is usually applied to extract the mesh surface from neural implicit functions. Compared with widely used mesh surfaces and discretized volumes, implicit neural functions are able to model continuous surfaces with fix-sized multi-layer perceptrons, making it more natural and efficient to represent complex geometries with arbitrary topologies. 

\begin{figure*}
	\centering
	\includegraphics[width=\linewidth]{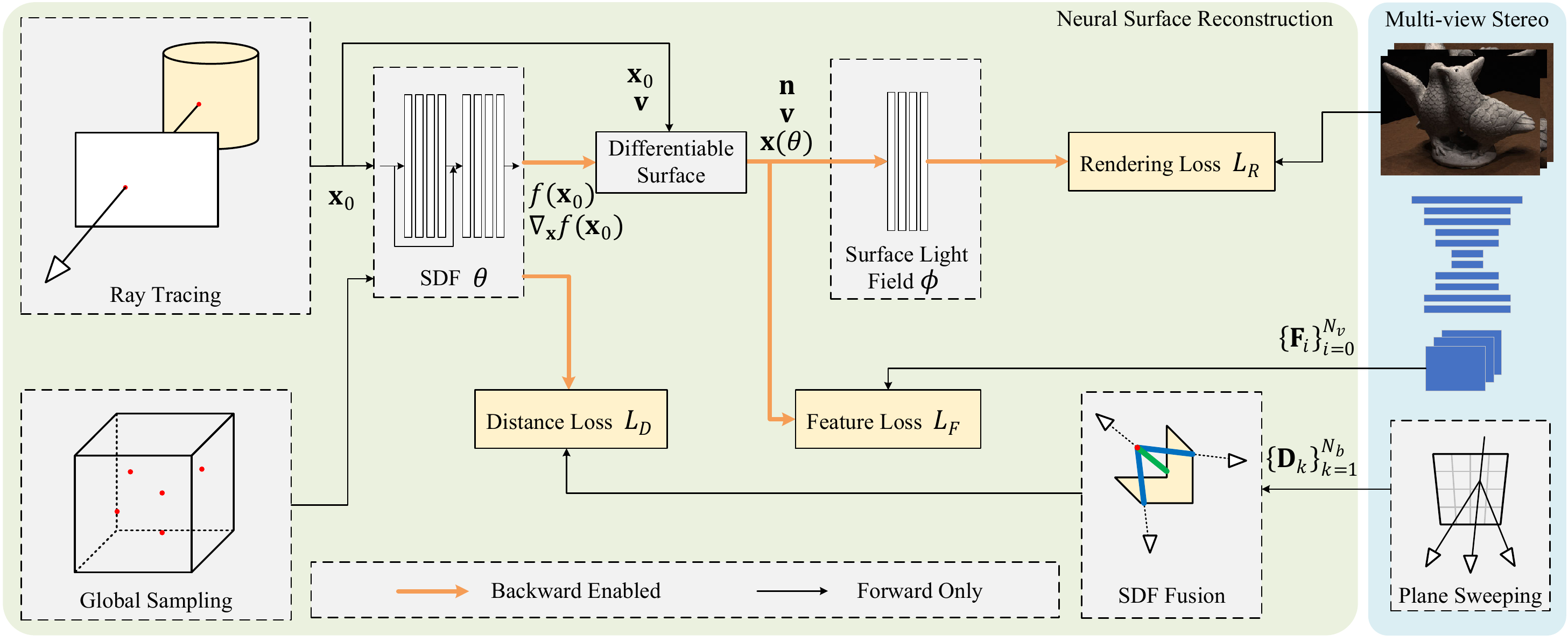}
	\caption{\textbf{Illustration of the proposed framework.} In the network, points are sampled from the space and their signed distance values are directly supervised by MVS depth outputs (Sec.~\ref{sec:geo}). Then, surface points are calculated by ray tracing and refined by deep feature consistency (Sec.~\ref{sec:feature}). Finally, the SDF geometry and surface light field are jointly optimized by the render loss (Sec.~\ref{sec:render}). }
	\label{fig:method}
	\vspace{-3mm}
\end{figure*}

\fakepara{Surface Reconstruction by Differentiable Rendering}
Apart from MVS, rendering and view synthesis based methods \cite{kato2020differentiable} provide an alternative to estimate scene geometries by minimizing the difference between rendered and input images. The geometry is either represented using soft representations such as density/transparency fields \cite{sitzmann2019scene,mildenhall2020nerf,liu2020neural,martin2020nerf,zhang2020nerf++}, or explicit representations such as occupancy fields \cite{niemeyer2020differentiable} and signed distance fields \cite{liu2019learning,liu2020dist,yariv2020multiview}. 

Our method is most related to IDR \cite{yariv2020multiview} which uses SDF and surface light field as the scene representation. These two implicit networks are jointly trained by the render loss, and image masks are applied for constrained SDF optimization.  However, the reconstruction quality of IDR highly depends on the accuracy of the input masks, and inaccurate masks may result in either missing or extra mesh surfaces.  As auto object segmentation methods \cite{zhao2019pyramid,qin2019basnet} cannot always be perfect, IDR applies manually labeled masks to ensure the reconstruction quality. In this work, we introduce multi-view stereo and feature consistency as our geometry constraints to improve the surface quality and relax the requirement of image masks. 

\section{Method}

\subsection{Geometry and Appearance Representations}
In our network, the surface $ S_{\theta} $ is explicitly modeled as the zero level set of a SDF, which is represented by a MLP $ f $ in the network.  
We define $ \theta $ the learnable parameters of $ f $. The MLP will take a query location $ \mathbf{x} $ as input and output a distance from the query to the closest surface point. 
\begin{equation}
\begin{aligned}
S_{\theta} = \{\mathbf{x} \in \mathbb{R}^3 | f(\mathbf{x}; \theta) = 0 \}
\end{aligned}
\end{equation}
Inspired by IDR \cite{yariv2020multiview}, our scene appearance is represented by a surface light field using another MLP $g$ with learnable parameters $ \phi $. The surface light field takes the query surface point $ \mathbf{x} $, its normal vector $ \mathbf{n} $ and the unit vector of the viewing ray $ \mathbf{v} $ as input and outputs the RGB color $\mathbf{c}$ of the query. 
\begin{equation}
\begin{aligned}
\mathbf{c} = g(\mathbf{x}, \mathbf{n}, \mathbf{v}; \phi)
\end{aligned}
\label{eq:slf}
\end{equation}
During the rendering, the intersection point of the viewing ray and the surface is obtained by sphere tracing, and the point normal can be calculated as the analytical gradient of the implicit surface $\mathbf{n} = \nabla_{\mathbf{x}} f(\mathbf{x}; \theta) $. 

\fakepara{Differentiable Surface Intersection}
The sphere tracing is not a differentiable operation in the network. Following previous works \cite{niemeyer2020differentiable, yariv2020multiview}, we construct the first order approximation of the function from network parameters to the intersection location. For the current network parameters $ \theta_0 $, viewing ray $ \mathbf{v} $ and the intersection point $ \mathbf{x}_0 $ on this ray, we take implicit differentiation on the equation $ f(\mathbf{x}; \theta) \equiv 0 $, and the surface intersection can be expressed as a function of $ \theta $:
\begin{equation}
\begin{aligned}
\mathbf{x}(\theta) = \mathbf{x}_0 - \frac{f(\mathbf{x}_0;\theta) - f(\mathbf{x}_0;\theta_0)}{\nabla_{\mathbf{x}} f(\mathbf{x}_0;\theta_0) \cdot \mathbf{v}}\mathbf{v}
\end{aligned}
\label{eq:intersection}
\end{equation}
where $ f(\mathbf{x}_0;\theta_0) $ and $ \nabla_{\mathbf{x}} f(\mathbf{x}_0;\theta_0) $ are constants. 

\subsection{Geometry Supervision}\label{sec:geo}

Multi-view stereo algorithms are able to provide high-quality depth maps as a dense representation of the scene. In this section, we describe how to use MVS depth maps to supervise our SDF optimization.

\fakepara{Multi-view Depth Map Estimation}
In our network, the MVS module aims to generate deep image features and qualified depth maps for all input images. We apply the open-sourced Vis-MVSNet \cite{zhang2020visibility} as our depth generation module. For a reference image $\mathbf{I}_0$ and its $N_v$ neighboring source images $ \{\mathbf{I}_i\}_{i=1}^{N_v} $, a standard UNet is first applied to extract deep image feature maps $ \{\mathbf{F}_i\}_{i=0}^{N_v} $. Then, all feature maps will be warped into the camera frustum of $\mathbf{I}_0$ and construct a 3D cost volume $ \mathbf{C} $. We further regularize the cost volume by 3D CNNs and obtain the probability distribution of the depth samples by \textit{softmax}. Finally, the depth $ \mathbf{D}_0 $ is regressed from the probability volume by taking the depth \textit{expectation}. In addition, for a pixel $\mathbf{p}$ in the depth map, we evaluate its probability sum $ \mathbf{P}(\mathbf{p}) $ around the predicted depth value as an indicator of the depth confidence \cite{yao2018mvsnet}. Pixels with low confidence will be filtered out to generate a cleaned depth map.  

\begin{figure}
	\centering
	\includegraphics[width=0.5\linewidth]{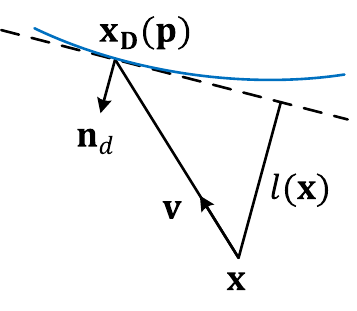}
	\caption{Approximated signed distance. }
	\label{fig:dist}
	\vspace{-3mm}
\end{figure}

\fakepara{Direct SDF Supervision}
Previous work \cite{niemeyer2020differentiable} proposes to train the implicit network by minimizing the difference between the traced depth map and the ground truth one. However, such strategy can only affect network outputs near the current surface estimation. 
To ensure the SDF to be correctly recovered in the whole space, we instead randomly sample points from the whole space and compute the distance from the sample point to the MVS depth map. 

Specifically, given a sample point $\mathbf{x}$ and a depth map $\mathbf{D}$, we first project $\mathbf{x}$ to the depth map at location $\mathbf{p}$. Then we backproject the MVS depth $ \mathbf{D}(\mathbf{p}) $ at the same location to the space as $\mathbf{x}_{\mathbf{D}}(\mathbf{p})$. As is shown in Fig.~\ref{fig:dist}, the signed distance from $\mathbf{x}$ to MVS surface is approximated as
\begin{equation}
	l(\mathbf{x}) = \text{sgn}[( \mathbf{x}_{\mathbf{D}}(\mathbf{p}) - \mathbf{x} )\cdot \mathbf{v}] (- \mathbf{n}_d\cdot \mathbf{v}) \| \mathbf{x}_{\mathbf{D}}(\mathbf{p}) - \mathbf{x} \|
\end{equation}
where $ \mathbf{n}_d $ is the normal calculated from depth.
Also, if the probability sum $\mathbf{P}(\mathbf{p})$ is smaller than a threshold $T_{prob}$, we consider this pixel as in the background and will exclude the corresponding point from the distance computation. This approximated signed distance can be used to supervise the SDF training, and we define the distance loss $ L_D $ as:
\begin{equation}
\begin{aligned}
L_D(\theta) = \frac{1}{|\mathbb{S}|} \sum_{\mathbf{x}\in \mathbb{S}} | f(\mathbf{x};\theta) - l(\mathbf{x}) |
\end{aligned}
\label{eq:l_d}
\end{equation}
where $\mathbb{S}$ is the set of valid sample points.

\fakepara{Signed Distance Fusion in a Mini-batch}
One problem of Eq.~\ref{eq:l_d} is that the approximated signed distance $l(\mathbf{x})$ calculated from a single depth map is usually not reliable. First, a sample point in the free space may be occluded in a given view. Second, the approximated $l(\mathbf{x})$ may be inaccurate when non-planar surfaces occur. To improve the accuracy of $l(\mathbf{x})$, we group $N_b$ views in a mini-batch during training, and $l(\mathbf{x})$ will be refined by fusing multiple observations from $N_b$ depth maps within the mini-batch. 

For a query point $\mathbf{x}$, we first calculate its approximated signed distances $\{l_k(\mathbf{x})\}_{k = 1}^{N_b}$ in each depth map. According to the sign of $l_k(\mathbf{x})$, we define that a point is outside the surface if at least $T_{out}$ distances from $\{l_k(\mathbf{x})\}_{k = 1}^{N_b}$ are positive.  After the query point is determined to be inside or outside, we collect the per-view distance with the same sign and take the minimum depth distance as the absolute value of the fused distance $l(\mathbf{x})$.  We find that such simple fusion strategy can effectively filter out erroneous observations from single depth map, and the fused $l(\mathbf{x})$ is accurate enough to be used to guide the SDF optimization.  

\subsection{Local Geometry Refinement}\label{sec:refine}
The geometry supervision in Sec.~\ref{sec:geo} can correctly recover the surface topology. However, as depth maps from MVS networks are usually noisy, it is rather difficult to restore surface details in the final mesh reconstruction. To this end, we propose to optimize the feature consistency and the rendered image consistency during the network training.

\fakepara{Feature Consistency}\label{sec:feature}
In traditional MVS or mesh reconstruction pipelines, dense point clouds or mesh surfaces are usually refined via multi-view photo-consistency optimization \cite{vu2011high,delaunoy2011gradient,delaunoy2014photometric,li2016efficient,yao2019recurrent}. The photo-consistency of a surface point is defined as the matching cost (e.g., ZNCC) among multiple views.  In our work, note that deep image features have already been extracted in Vis-MVSNet. Inspired by \cite{yu2020fast} , we instead minimize the multi-view deep feature consistency. 

Suppose a surface point $\mathbf{x}$ is obtained via ray tracing in view 0, we denote its projections in view 0 and its neighboring views as $ \{\mathbf{p}_i\}_{i=0}^{N_v}$. As these projections refer to the same 3D point in space,  their deep image features should be consistent. The feature loss is defined as:
\begin{equation}
\begin{aligned}
L_F(\theta) = \frac{1}{N_vN_c} \sum_{i=1}^{N_v} | \mathbf{F}_0(\mathbf{p}_0) - \mathbf{F}_i(\mathbf{p}_i) |
\end{aligned}
\end{equation}
where $ N_c $ is the number of feature channels, $\mathbf{p}_i = \mathbf{K}_i (\mathbf{R}_i \mathbf{x} + \mathbf{t}_i)$ is the projection of $\mathbf{x}$ in view $i$ and $[\mathbf{K}_i, \mathbf{R}_i, \mathbf{t}_i]$ the corresponding camera parameters.  Deep image feature at pixel $ \mathbf{p}_i$, denoted as $\mathbf{F}_i(\mathbf{p_i})$, is obtained by bilinear interpolation. 
To optimize the SDF via the feature consistency loss, we derived the gradient of $L_F(\theta)$ with respect to the network parameter $\theta$ as:
\begin{equation}
\begin{aligned}
\frac{\partial L_F(\theta)}{\partial \theta} &= \frac{\partial L_F}{\partial \mathbf{x}} \cdot  \frac{\partial \mathbf{x}}{\partial \theta} =  (\sum_{i=0}^{N_v} \frac{\partial L_{F}}{\partial \mathbf{F}_i} \frac{\partial \mathbf{F}_i}{\partial \mathbf{p}_i} \frac{\partial \mathbf{p}_i}{\partial \mathbf{x}}) \cdot \frac{\partial \mathbf{x}}{\partial \theta} 
\end{aligned}
\end{equation}
where $\partial \mathbf{F}_i / \partial \mathbf{p}_i$ is the gradient of the feature map. The last term $\partial \mathbf{x} / \partial \theta$ can be calculated from the derivative of Eq.~\ref{eq:intersection}.

\begin{figure*}
	\centering
	\begin{tabular}{@{\hskip2pt}c@{\hskip2pt}@{\hskip2pt}c@{\hskip2pt}@{\hskip2pt}c@{\hskip2pt}@{\hskip2pt}c@{\hskip2pt}@{\hskip2pt}c@{\hskip2pt}}
		\begin{tikzpicture}\node[above right, inner sep=0](image) at (0,0) {\includegraphics[width=0.19\linewidth]{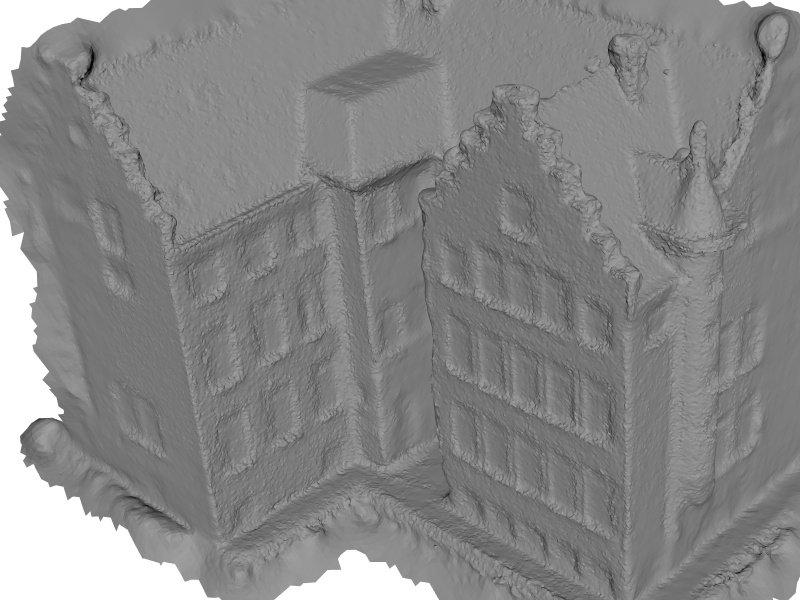}};\draw[very thick,red] (1.7,0) rectangle (2.7,2);\draw[very thick,green] (0.6,1.5) rectangle (1.5,2.3); \end{tikzpicture} &
		\begin{tikzpicture}\node[above right, inner sep=0](image) at (0,0) {\includegraphics[width=0.19\linewidth]{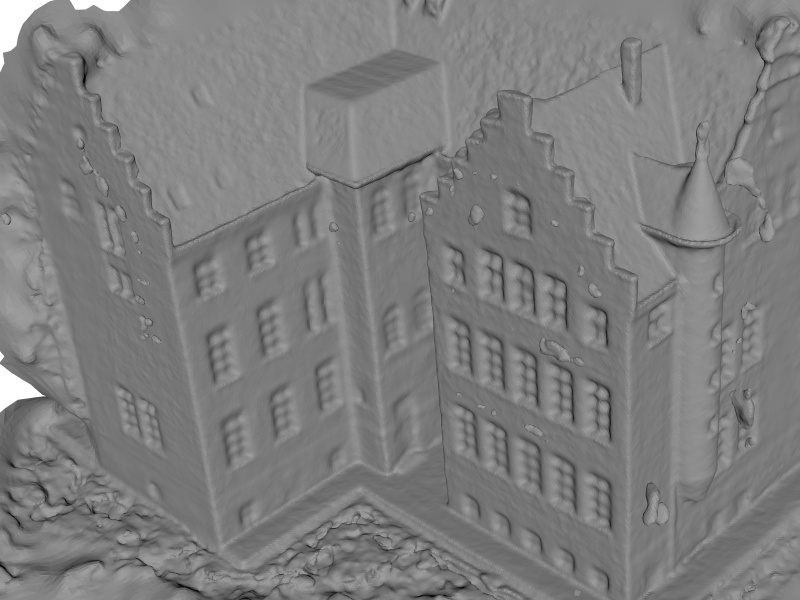}};\draw[very thick,green] (1.7,0) rectangle (2.7,2);\draw[very thick,green] (0.6,1.5) rectangle (1.5,2.3); \end{tikzpicture} &
		\begin{tikzpicture}\node[above right, inner sep=0](image) at (0,0) {\includegraphics[width=0.19\linewidth]{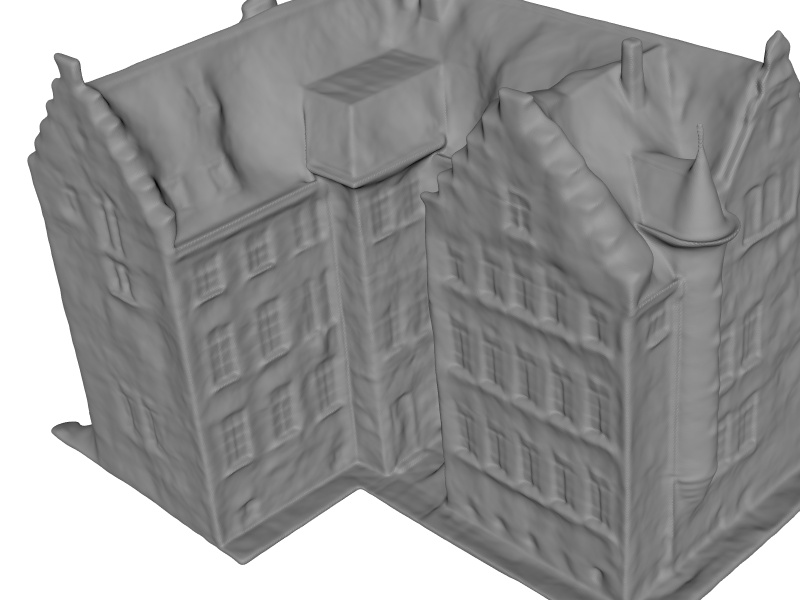}};\draw[very thick,green] (1.7,0) rectangle (2.7,2);\draw[very thick,red] (0.6,1.5) rectangle (1.5,2.3); \end{tikzpicture} &
		\begin{tikzpicture}\node[above right, inner sep=0](image) at (0,0) {\includegraphics[width=0.19\linewidth]{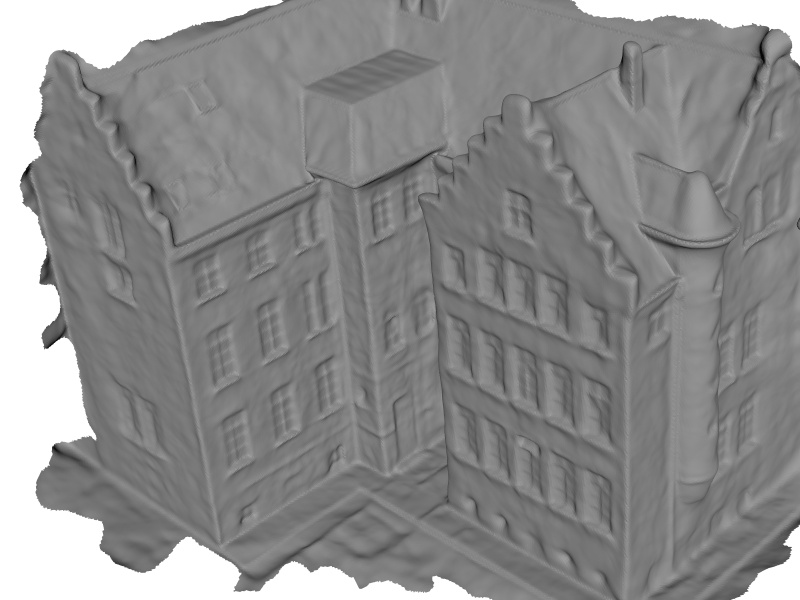}};\draw[very thick,green] (1.7,0) rectangle (2.7,2);\draw[very thick,green] (0.6,1.5) rectangle (1.5,2.3); \end{tikzpicture} & 
		\begin{tikzpicture}\node[above right, inner sep=0](image) at (0,0) {\includegraphics[width=0.19\linewidth]{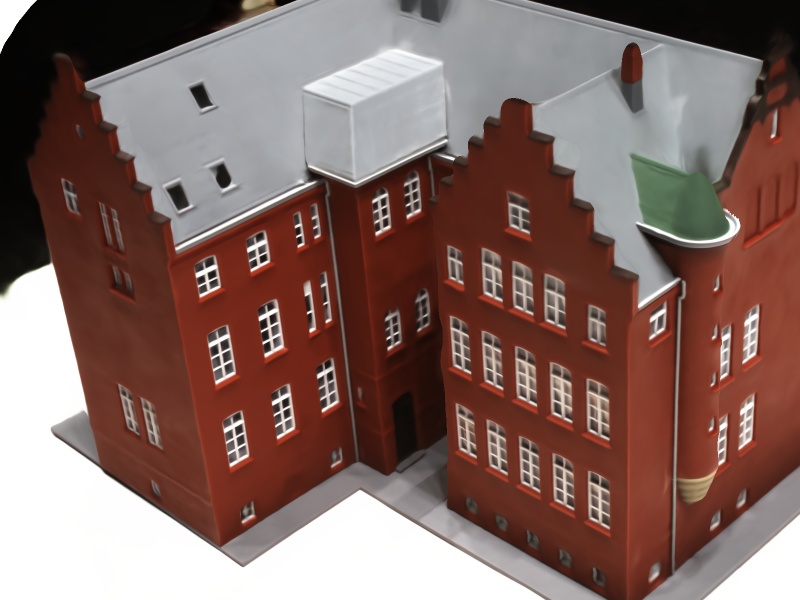}}; \end{tikzpicture} \\

		\begin{tikzpicture}\node[above right, inner sep=0](image) at (0,0) {\includegraphics[width=0.19\linewidth]{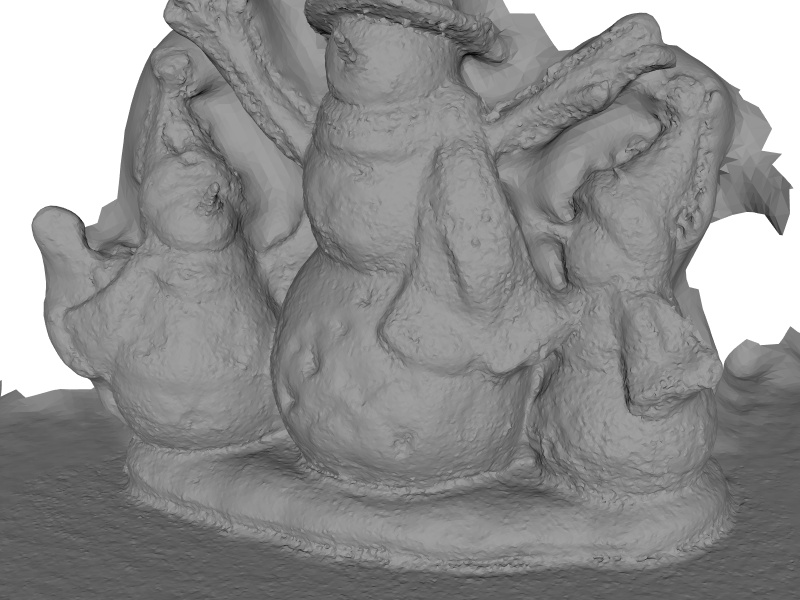}};\draw[very thick,red] (0.4,0.4) rectangle (2.4,1.4); \end{tikzpicture} &
		\begin{tikzpicture}\node[above right, inner sep=0](image) at (0,0) {\includegraphics[width=0.19\linewidth]{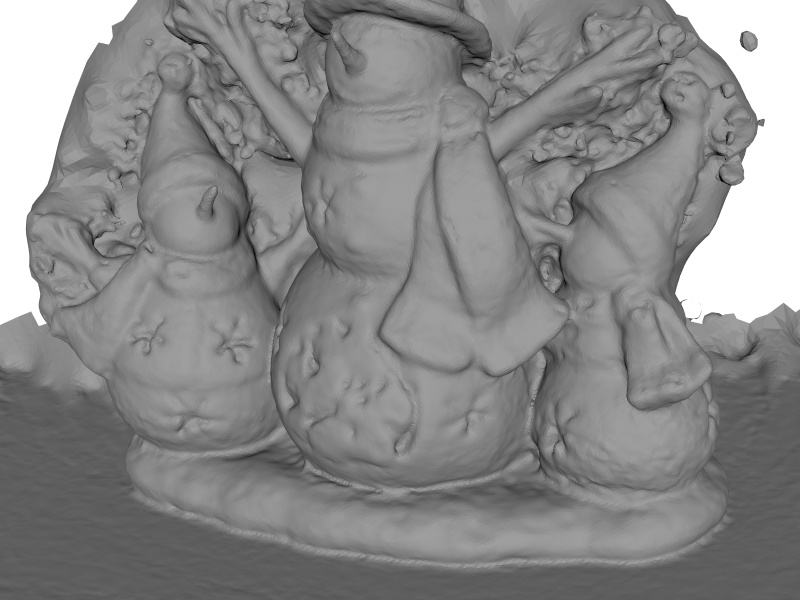}};\draw[very thick,red] (0.4,0.4) rectangle (2.4,1.4); \end{tikzpicture} &
		\begin{tikzpicture}\node[above right, inner sep=0](image) at (0,0) {\includegraphics[width=0.19\linewidth]{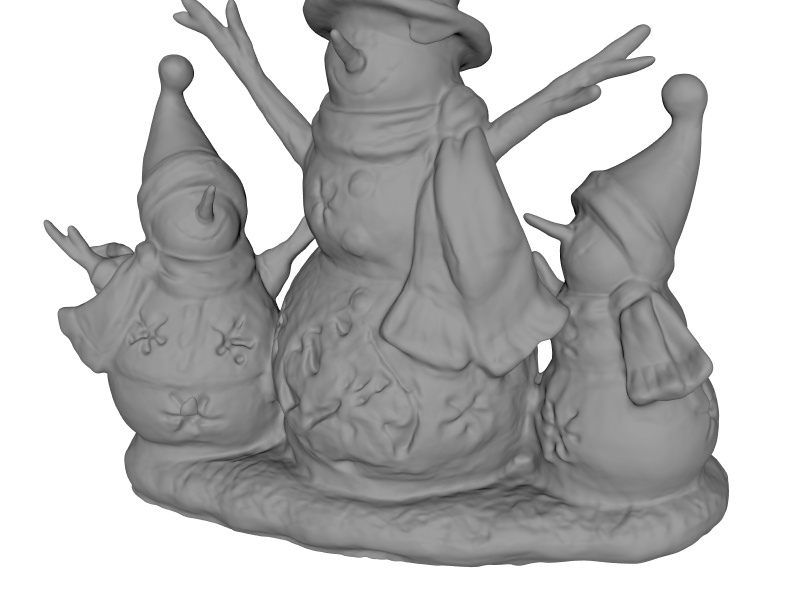}};\draw[very thick,green] (0.4,0.4) rectangle (2.4,1.4); \end{tikzpicture} &
		\begin{tikzpicture}\node[above right, inner sep=0](image) at (0,0) {\includegraphics[width=0.19\linewidth]{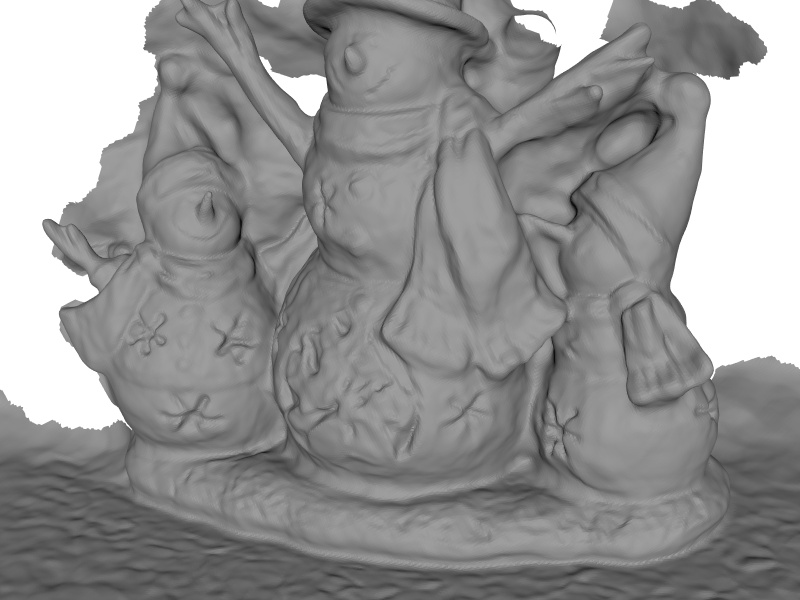}};\draw[very thick,green] (0.4,0.4) rectangle (2.4,1.4); \end{tikzpicture} & 
		\begin{tikzpicture}\node[above right, inner sep=0](image) at (0,0) {\includegraphics[width=0.19\linewidth]{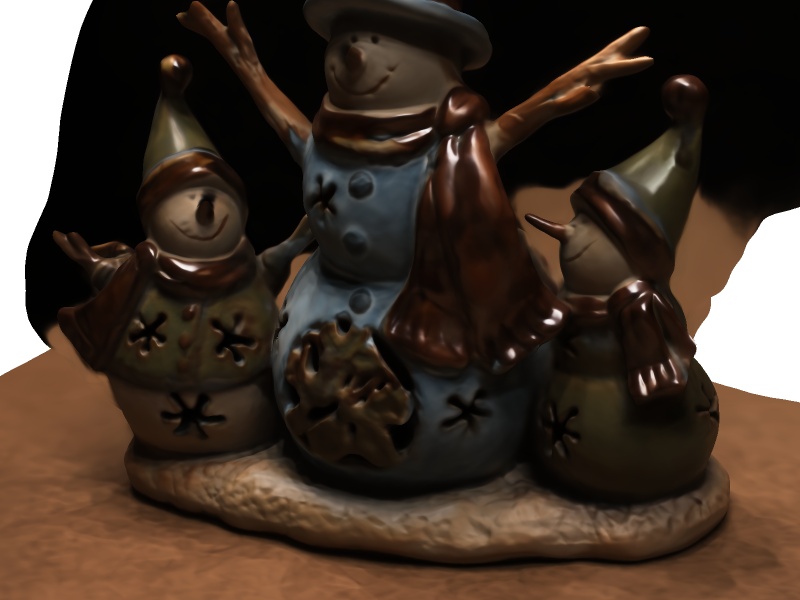}}; \end{tikzpicture} \\
		
		\begin{tikzpicture}\node[above right, inner sep=0](image) at (0,0) {\includegraphics[width=0.19\linewidth]{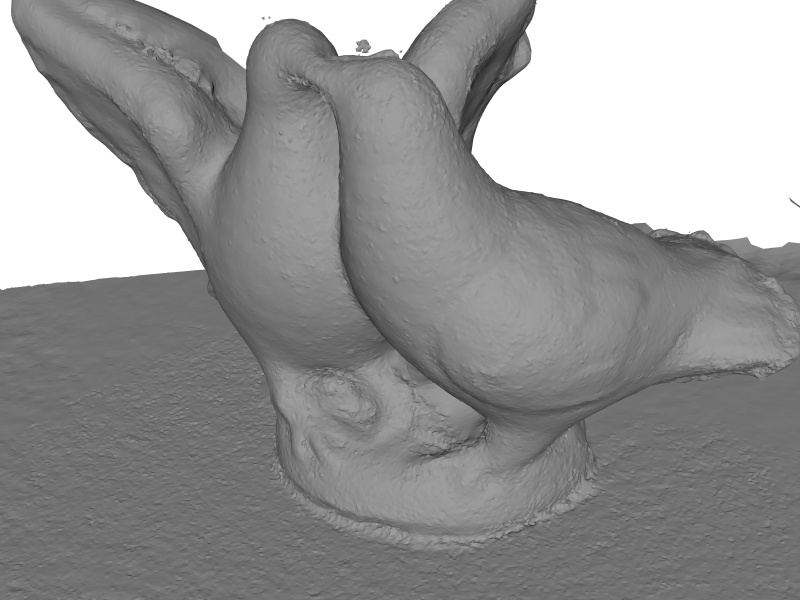}};\draw[very thick,red] (0.8,1) rectangle (2.3,2); \end{tikzpicture} &
		\begin{tikzpicture}\node[above right, inner sep=0](image) at (0,0) {\includegraphics[width=0.19\linewidth]{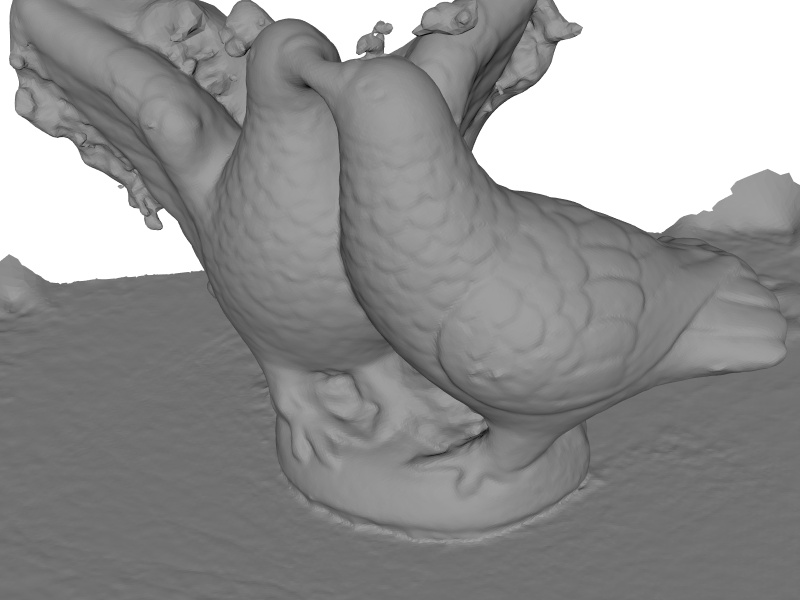}};\draw[very thick,red] (0.8,1) rectangle (2.3,2); \end{tikzpicture} &
		\begin{tikzpicture}\node[above right, inner sep=0](image) at (0,0) {\includegraphics[width=0.19\linewidth]{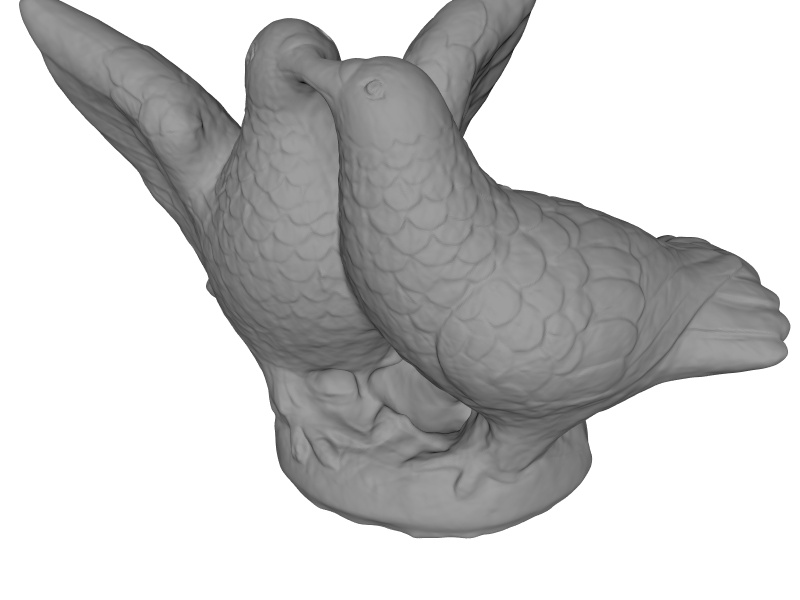}};\draw[very thick,green] (0.8,1) rectangle (2.3,2); \end{tikzpicture} &
		\begin{tikzpicture}\node[above right, inner sep=0](image) at (0,0) {\includegraphics[width=0.19\linewidth]{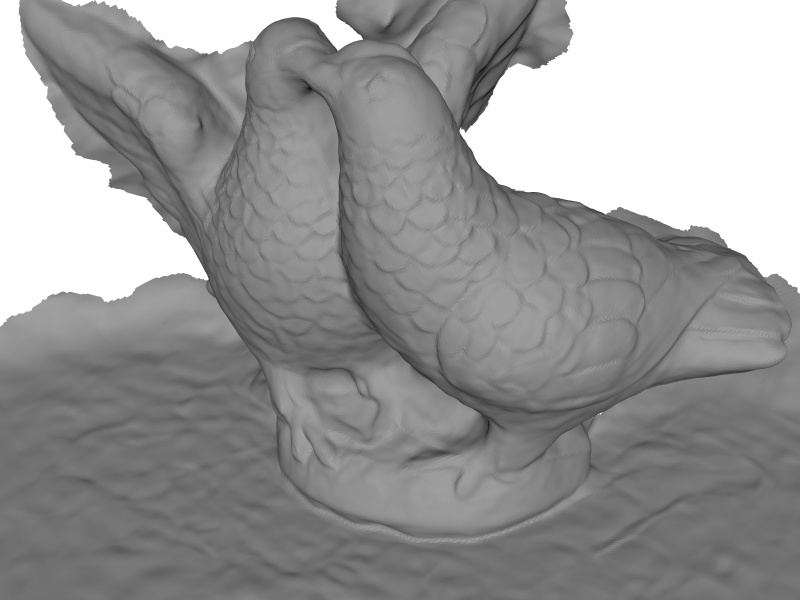}};\draw[very thick,green] (0.8,1) rectangle (2.3,2); \end{tikzpicture} & 
		\begin{tikzpicture}\node[above right, inner sep=0](image) at (0,0) {\includegraphics[width=0.19\linewidth]{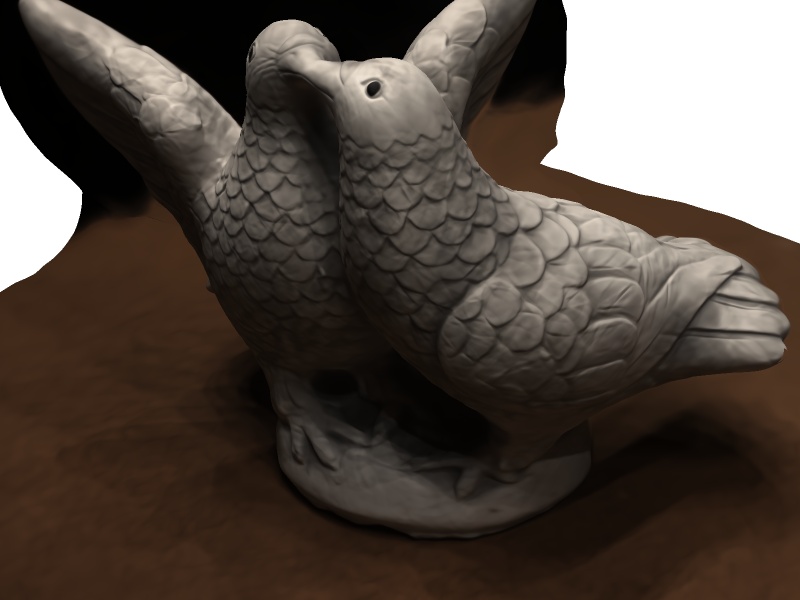}}; \end{tikzpicture} \\

		Colmap & Vis-MVSNet & IDR (perfect mask) & MVSDF (Ours) & MVSDF (Ours) Render
	\end{tabular}
	\caption{\textbf{Qualitative results on DTU dataset.} Our method produces both high quality meshes and rendered images without requiring masks as input. }
	\label{fig:dtu}
	\vspace{-3mm}
\end{figure*}
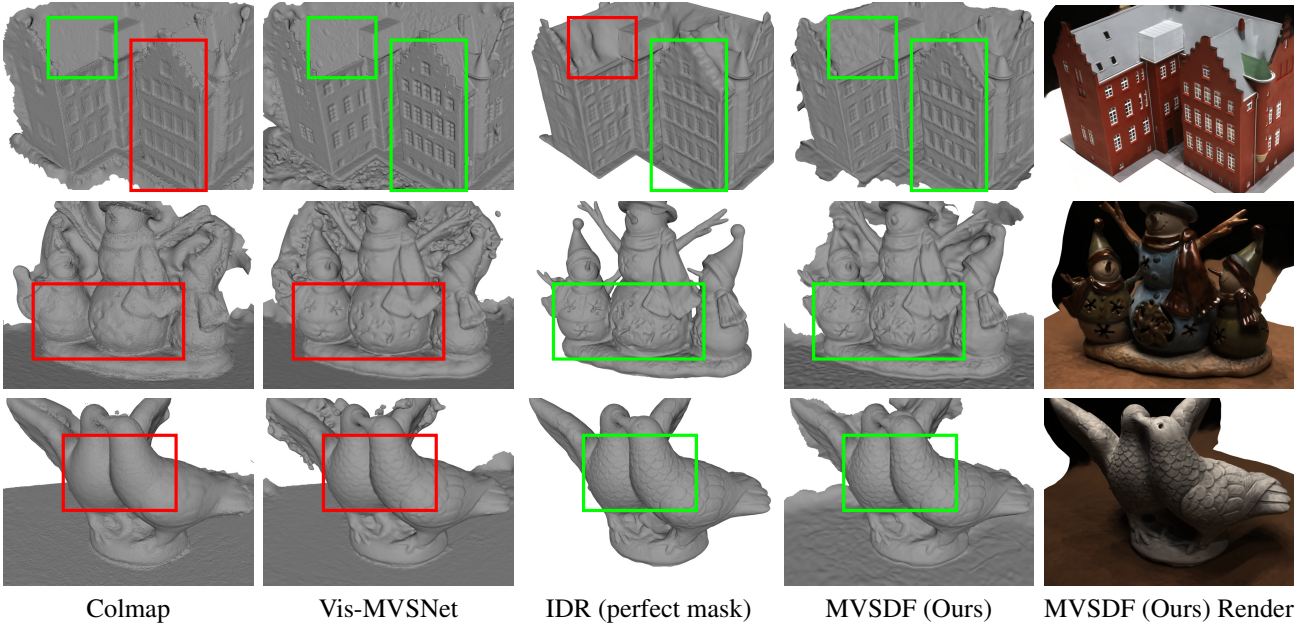

\begin{figure}
	\centering
	\begin{tabular}{@{\hskip4pt}c@{\hskip4pt}@{\hskip4pt}c@{\hskip4pt}}
		\includegraphics[width=0.45\linewidth]{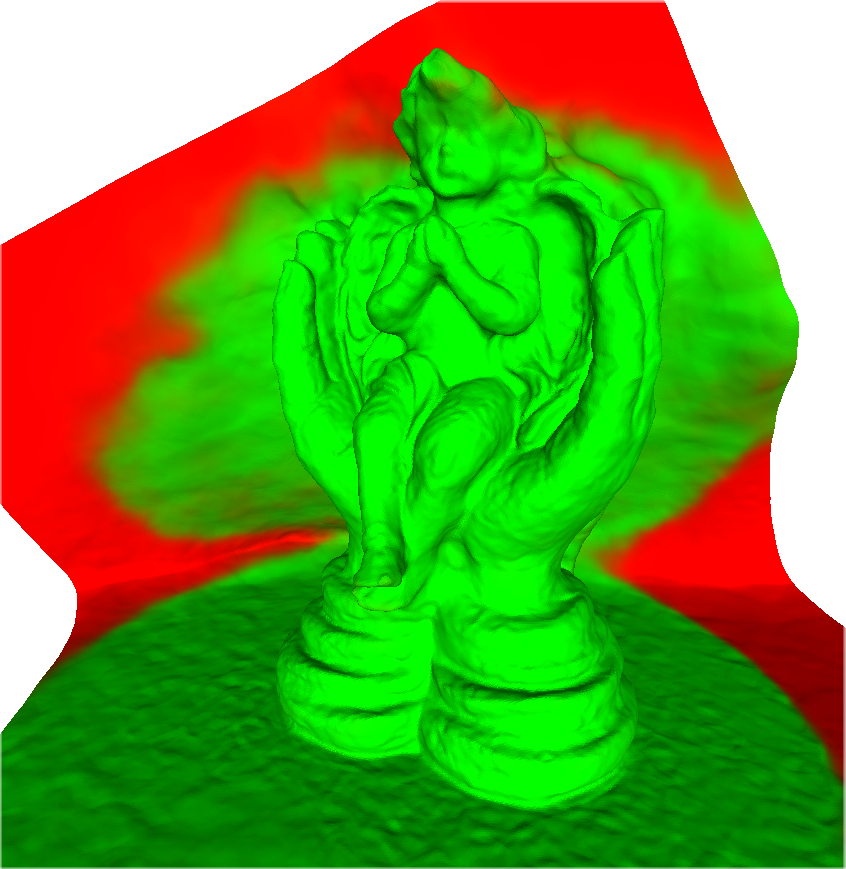} & 
		\includegraphics[width=0.45\linewidth]{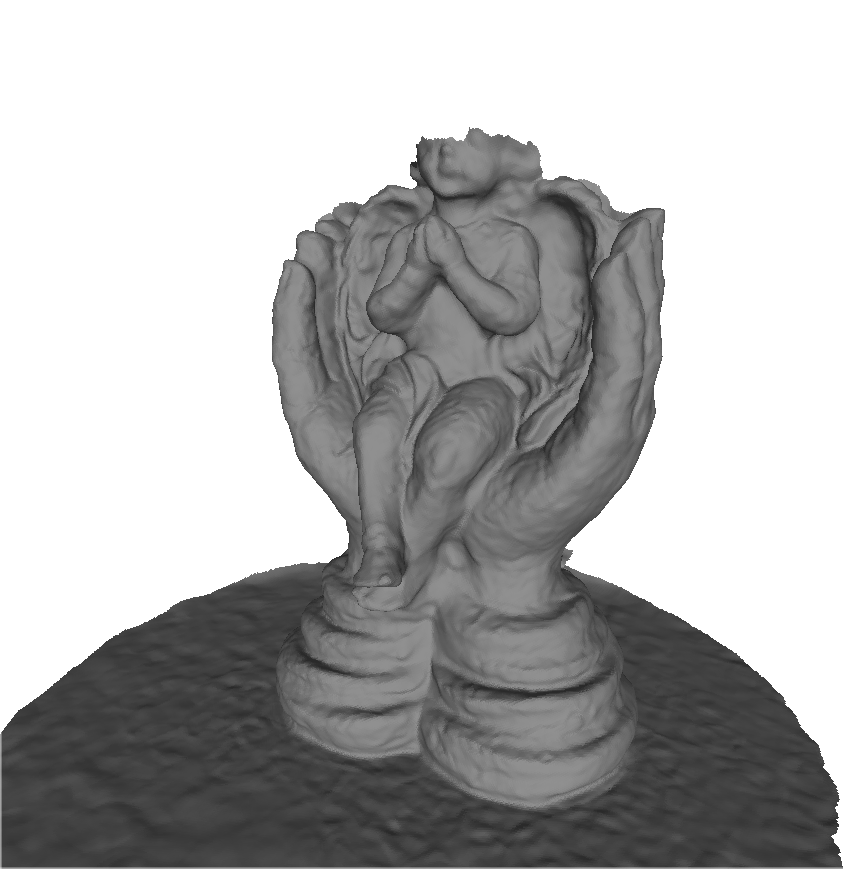} \\
	\end{tabular}
	\caption{\textbf{Illustration of surface indicator and mesh trimming.} Extra surfaces are trimmed with a graph-cut based algorithm (Sec.~\ref{sec:trim}) according to learned surface indicators (Sec.~\ref{sec:indicator}, green indicates accurate surface).}
	\label{fig:trim}
	\vspace{-3mm}
\end{figure}

Compared with the render loss to be discussed in the next paragraph, the proposed feature loss introduces a photo-consistency constraint in an early stage of the whole differentiable rendering pipeline, which reduces the geometry and appearance ambiguity during the joint optimization.  We show in ablation study that $L_F(\theta)$ can effectively increase the mesh reconstruction quality (see Tab.~\ref{tab:abl}).  

\fakepara{Rendered Image Consistency}\label{sec:render}
The rendered image consistency is widely used in recent differentiable rendering pipelines \cite{niemeyer2020differentiable,mildenhall2020nerf,yariv2020multiview}. For pixel $\mathbf{p}$ in the image, we can trace its surface intersection $\mathbf{x}$ in the space. The rendered color $ \mathbf{c}(\mathbf{p}) $ of pixel $\mathbf{p}$ can be directly fetched from the surface light field by feeding $ \mathbf{x}(\theta), \nabla_{\mathbf{x}}f(\mathbf{x};\theta) $ and $ \mathbf{v} $ into function $ g $. 
The render loss is then calculated as the L1 distance from the rendered color to the input image color:
\begin{equation}
\begin{aligned}
L_R(\theta, \phi) = \frac{1}{|\mathbb{S}_{I}|} \sum_{\mathbf{p} \in \mathbb{S}_{I}} | \mathbf{c}(\mathbf{p}) - \mathbf{I}(\mathbf{p}) |
\end{aligned}
\end{equation}
where $\mathbb{S}_{I}$ denote the set of image pixels whose view ray can intersect the surface in the space.

The render loss can jointly optimize the geometry $\theta$ and the appearance $\phi$. Compared with $L_F$, $L_R$ is more sensitive to local color changes, and plays an important role in recovering high-fidelity surface details. 

\subsection{Valid Surface Indicator}\label{sec:indicator}
If input images cannot fully cover the object of interest, the surface of unseen areas will be not well defined, and will tend to produce extrapolated surfaces in the background areas. To distinguish such invalid surface, we use another indicator function to mark whether a space point can be traced from some input views. Specifically, function $h(\mathbf{x};\gamma)$ represents an indicator that $\mathbf{x}$ is in the valid surface. During each training iteration, indicators of successfully traced locations $\{\mathbf{x^{+}}\}$ are set to 1. To prevent $h(\mathbf{x};\gamma)$ reporting $1$ everywhere, we also randomly sample points $ \{\mathbf{x}^-\} $ in the space and set the background indicator to $ 0 $. We then apply the binary cross entropy as our indicator loss.
\begin{equation}
\begin{aligned}
L_P(\gamma) = \sum_{\mathbf{x}^+} -\log h(\mathbf{x}^+;\gamma) + \sum_{\mathbf{x}^-} -\log (1-h(\mathbf{x}^-;\gamma))
\end{aligned}
\end{equation}

Note that our MVS depth map is filtered using the corresponding probability map and we will not apply ray tracing on those filtered pixels. As a result, filtered regions in MVS depth maps will tend to be assigned with the background indicator of 0. In other words, we could identify the invalid surface area according to filtered MVS depth maps.

\begin{table*}[]
	\centering
	\resizebox{\linewidth}{!}{%
		\begin{tabular}{l||cc|ccc||cc|ccc}
			\specialrule{.2em}{.1em}{.1em}
			& \multicolumn{5}{c||}{Chamfer (mm)}         & \multicolumn{5}{c}{PSNR}                    \\ \cline{2-11}
			& Colmap \cite{schonberger2016pixelwise} & Vis-MVSNet \cite{zhang2020visibility} & DVR \cite{niemeyer2020differentiable} & IDR \cite{yariv2020multiview} & MVSDF (Ours) & Colmap \cite{schonberger2016pixelwise} & Vis-MVSNet \cite{zhang2020visibility} & DVR \cite{niemeyer2020differentiable} & IDR \cite{yariv2020multiview}   & MVSDF (Ours)  \\ \hline
			24   & 0.99   & 0.98       & 4.10 & 1.63 & \textbf{0.83} & 18.44  & 18.35      & 16.23 & 23.29 & \textbf{25.02} \\
			37   & 2.35   & 2.10       & 4.54 & 1.87 & \textbf{1.76} & 14.37  & 14.71      & 13.93 & \textbf{21.36} & 19.47 \\
			40   & 0.73   & 0.93       & 4.24 & \textbf{0.63} & 0.88 & 19.24  & 18.60      & 18.15 & 24.39 & \textbf{25.96} \\
			55   & 0.53   & 0.46       & 2.61 & 0.48 & \textbf{0.44} & 18.27  & 19.07      & 17.14 & 22.96 & \textbf{24.14} \\
			63   & 1.56   & 1.89       & 4.34 & \textbf{1.04} & 1.11 & 19.92  & 17.55      & 17.84 & \textbf{23.22} & 22.16 \\
			65   & 1.01   & \textbf{0.67}       & 2.81 & 0.79 & 0.90 & 13.80  & 17.17      & 17.23 & 23.94 & \textbf{26.89} \\
			69   & 0.89   & \textbf{0.67}       & 2.53 & 0.77 & 0.75 & 21.23  & 21.81      & 16.33 & 20.34 & \textbf{26.38} \\
			83   & 1.14   & \textbf{1.08}       & 2.93 & 1.33 & 1.26 & 22.67  & 23.11      & 18.10 & 21.87 & \textbf{25.79} \\
			97   & 0.91   & \textbf{0.67}       & 3.03 & 1.16 & 1.02 & 18.19  & 18.68      & 16.61 & 22.95 & \textbf{26.22} \\
			105  & 1.46   & 0.95       & 3.24 & \textbf{0.76} & 1.35 & 20.43  & 21.68      & 18.39 & 22.71 & \textbf{27.29} \\
			106  & 0.79   & \textbf{0.66}       & 2.51 & 0.67 & 0.87 & 20.73  & 21.03      & 17.39 & 22.81 & \textbf{27.78} \\
			110  & 1.08   & 0.85       & 4.80 & 0.90 & \textbf{0.84} & 17.93  & 18.41      & 14.43 & 21.26 & \textbf{23.82} \\
			114  & 0.44   & \textbf{0.30}       & 3.09 & 0.42 & 0.34 & 19.08  & 19.42      & 17.08 & 25.35 & \textbf{27.79} \\
			118  & 0.68   & \textbf{0.45}       & 1.63 & 0.51 & 0.47 & 22.05  & 23.85      & 19.08 & 23.54 & \textbf{28.60} \\
			122  & 0.73   & 0.51       & 1.58 & 0.53 & \textbf{0.46} & 22.04  & 24.29      & 21.03 & 27.98 & \textbf{31.49} \\ \hline
			Mean & 1.02   & \textbf{0.88}       & 3.20 & 0.90 & \textbf{0.88} & 19.23  & 19.85      & 17.26 & 23.20 & \textbf{25.92} \\
			\specialrule{.2em}{.1em}{.1em}
	\end{tabular}}
	\caption{\textbf{Quantitative results on DTU dataset.} Our method achieves the best mean Chamfer distance as Vis-MVSNet \cite{zhang2020visibility} and the highest PSNR score among all methods.}
	\label{tab:dtu}
	\vspace{-3mm}
\end{table*}

\subsection{Loss}\label{sec:loss}
Apart from the aforementioned losses, we further regularize our SDF by a Eikonal loss \cite{gropp2020implicit} that restricts the expectation of the gradient magnitude to be $1$. 
\begin{equation}
\begin{aligned}
L_E(\theta) = \mathbb{E}_{x\in \mathbb{R}^3}(\|\nabla_{\mathbf{x}} f(\mathbf{x};\theta)\| - 1)^2
\end{aligned}
\end{equation}
The final loss is expressed as a weighted sum of all the aforementioned losses. 
\begin{equation}
\begin{aligned}
L = & w_R L_R(\theta, \phi) + w_F L_F(\theta) + w_D L_D(\theta) \\&+ w_E L_E(\theta) + w_P L_P(\gamma)
\end{aligned}
\end{equation}
where the weight $w$ will be changed over the network training. 
Our training process is divided into three stages: 1) in the first stage $ w_D $ is set to be dominant so as to determine the initial topology; 2) in the second stage, the significance of $ w_F $ is increased to recover finer structures in the surface; 3) in the final stage, both $ w_D $ and $ w_F $ is decreased so that the render loss can restore fine-scale details of the surface. 

\section{Experiments}

\subsection{Implementation}
\fakepara{Network Architecture}
The SDF is implemented by an 8-layer MLP with 512 hidden units and a skip connection in the middle. Positional encoding \cite{mildenhall2020nerf} is applied to input position to capture the high frequency information. This MLP simultaneously outputs the distance, the surface indicator probability and a descriptor of the location as an input of the surface light field function. Similarly, the surface light field is implemented by a 4-layer MLP with 512 hidden units. The function takes the point location, its descriptor, normal and viewing ray as input. Only the viewing direction is enhanced by positional encoding as the point descriptor has already included the rich positional information. In MVS module, we use one reference and two source images ($N_v = 2$) as input to Vis-MVSNet, and will output deep feature ($ N_c=32 $) maps of all images and the reference depth map. The MVS module is pretrained on \textit{BlendedMVS} \cite{yao2020blendedmvs} dataset and the parameters are fixed during training. 

\fakepara{Training}
For each input scene, the network is end-to-end trained for 10800 steps with a batch size of $ N_b = 8$. 
In each training step, 4096 pixels are uniformly sampled from each of the 8 images in the mini-batch for tracing surface intersections. Additionally, the same number of 3D points are sampled from the space to calculate the distance loss and the Eikonal loss. To recover correct topologies of thin structures, we need to sample more points near the object surface. This can be achieved by jittering surface points obtained from MVS depth maps. In distance fusion, the minimal number for outside decision $ T_{out}=2 $. The initial learning rate is $ 10^{-3} $ and is scaled down by $ 10 $ when reaching $ 4/6 $ and $ 5/6 $ of the whole training process. As mentioned in Sec.~\ref{sec:loss}, weights of the losses are set according to the training stages. Please refer to the supplementary material for detailed setting. 


The memory consumption is related to batch size, number of samples per images and number of source images. For \textit{DTU} dataset, our training setting takes $\sim$20 GB VRAM. And the whole training process takes 5.5 hours for one scan with 49 images on an NVidia RTX Titan. 

\begin{figure*}
	\centering
	\begin{tabular}{@{\hskip2pt}c@{\hskip2pt}@{\hskip2pt}c@{\hskip2pt}@{\hskip2pt}c@{\hskip2pt}@{\hskip2pt}c@{\hskip2pt}@{\hskip2pt}c@{\hskip2pt}}
		\begin{tikzpicture}\node[above right, inner sep=0](image) at (0,0) {\includegraphics[width=0.19\linewidth]{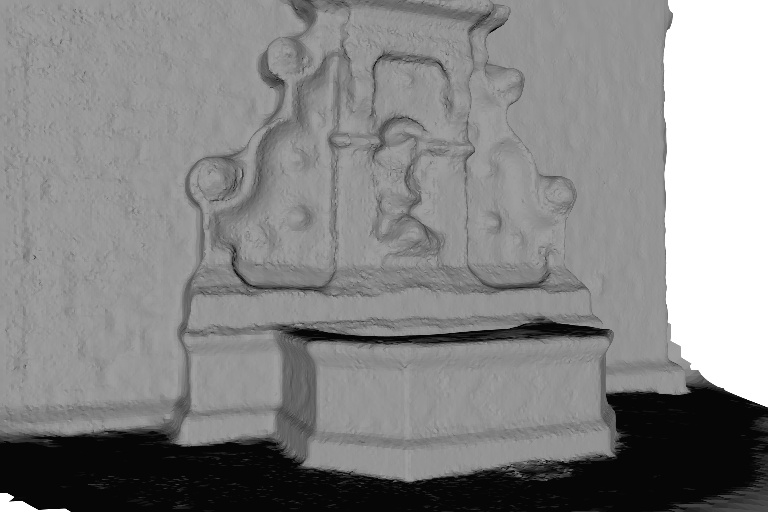}}; \end{tikzpicture} &
		\begin{tikzpicture}\node[above right, inner sep=0](image) at (0,0) {\includegraphics[width=0.19\linewidth]{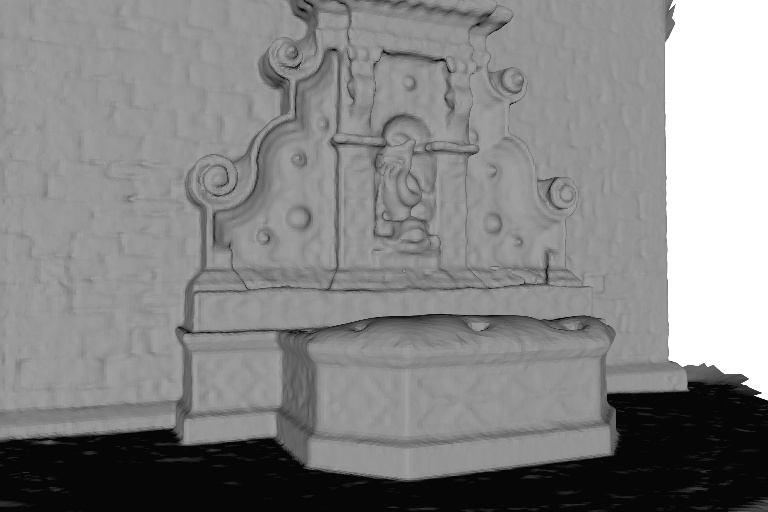}}; \end{tikzpicture} &
		\includegraphics[width=0.19\linewidth]{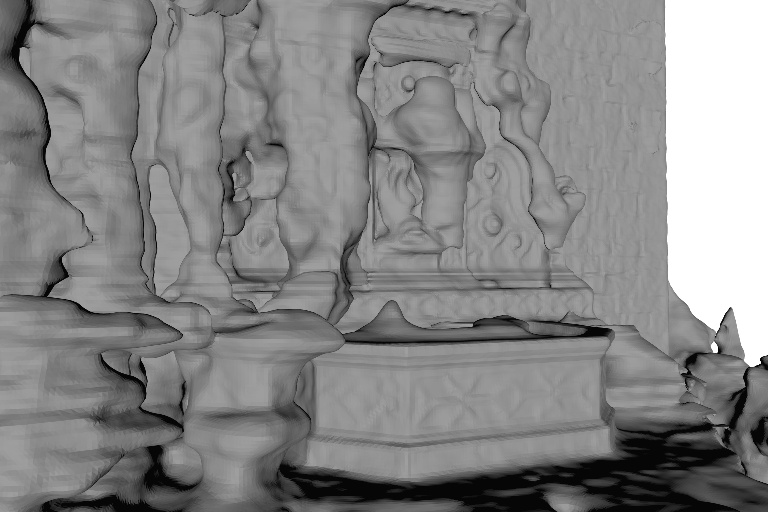} &
		\includegraphics[width=0.19\linewidth]{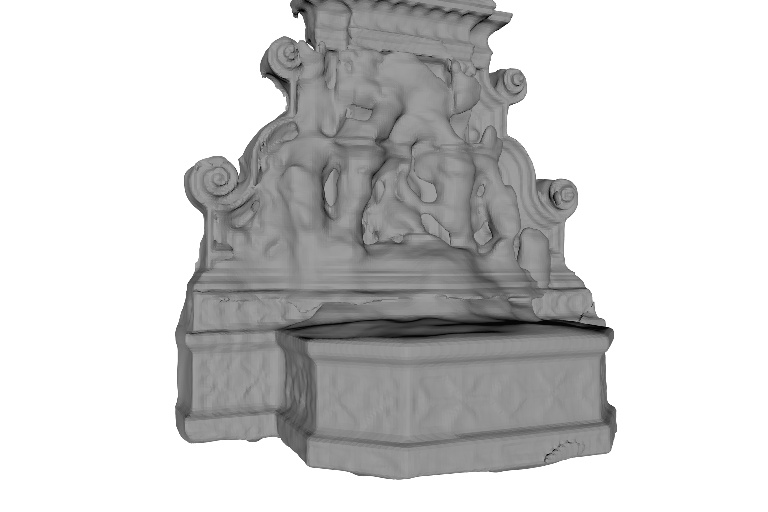} &
		\begin{tikzpicture}\node[above right, inner sep=0](image) at (0,0) {\includegraphics[width=0.19\linewidth]{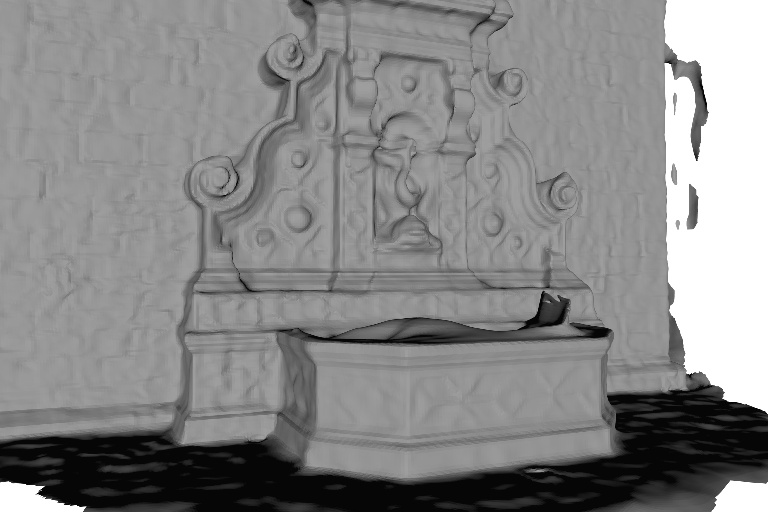}};\draw[very thick,green] (0.8,0.9) rectangle (1.7,1.8); \end{tikzpicture} \\
		
		\includegraphics[width=0.19\linewidth]{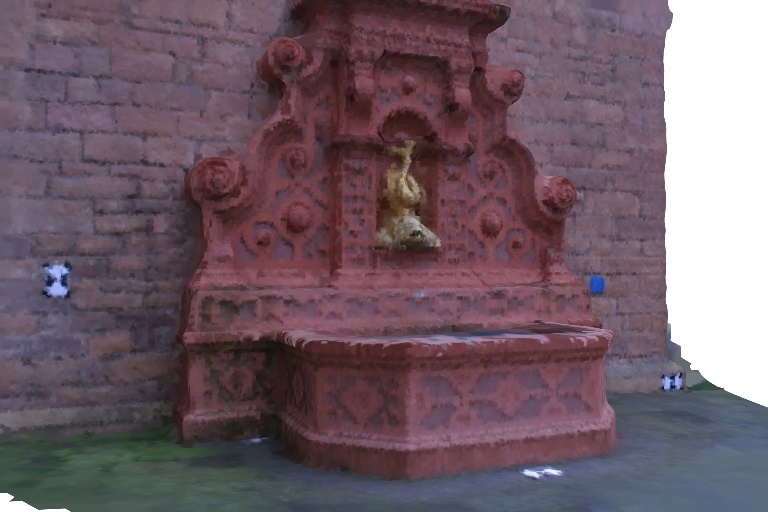} &
		\includegraphics[width=0.19\linewidth]{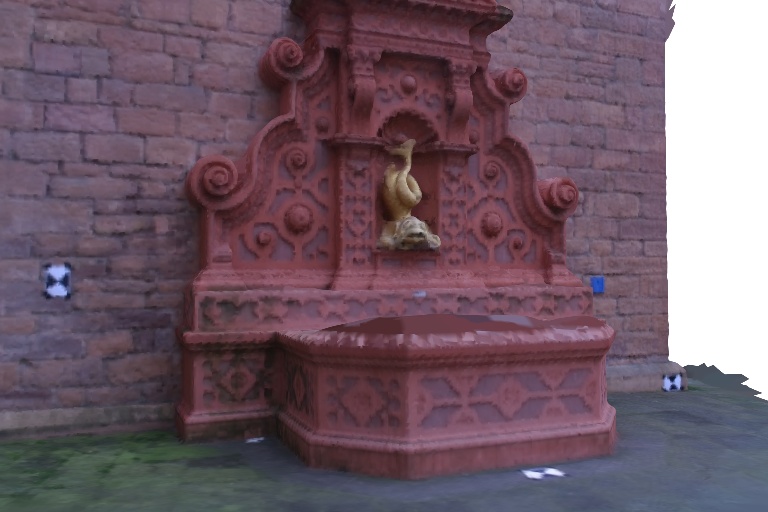} &
		\includegraphics[width=0.19\linewidth]{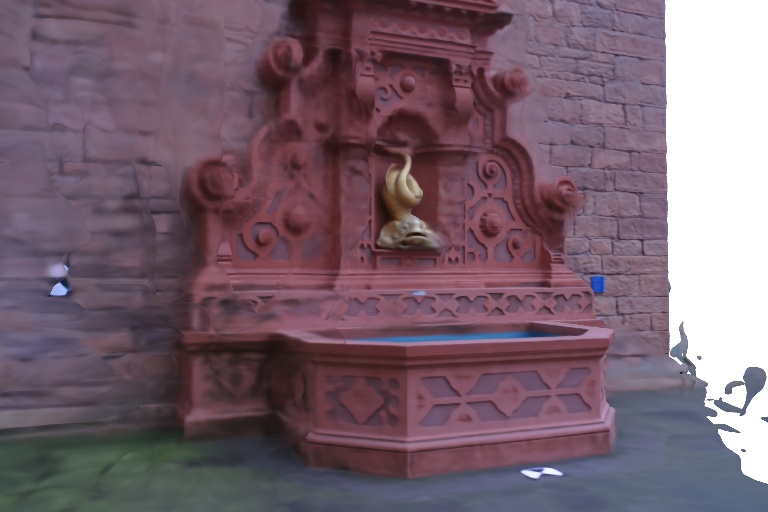} &
		\includegraphics[width=0.19\linewidth]{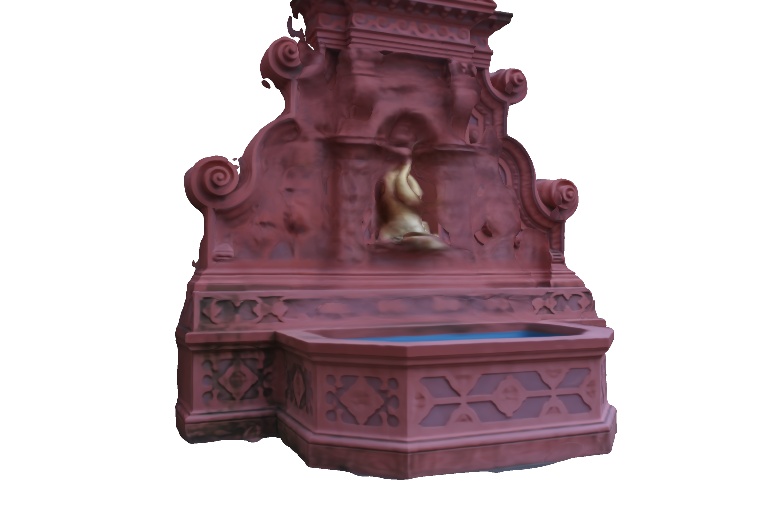} &
		\includegraphics[width=0.19\linewidth]{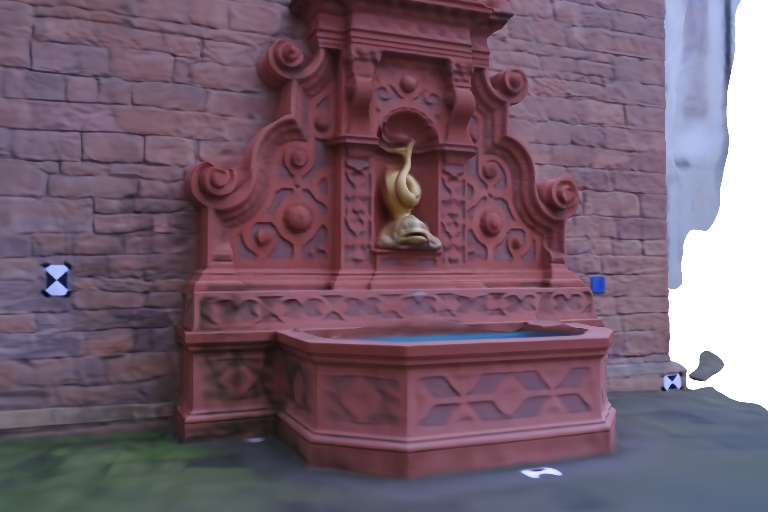} \\
		
		Colmap & Vis-MVSNet & IDR & IDR (w/o wall) & MVSDF (Ours)
	\end{tabular}
	\caption{\textbf{Qualitative results on EPFL dataset.} Our method is able to generate both high quality mesh and rendering results.}
	\label{fig:epfl}
	\vspace{-3mm}
\end{figure*}

\begin{table*}[]
	\centering
	\resizebox{\linewidth}{!}{%
		\begin{tabular}{l||cc|cc||cc|cc}
			\specialrule{.2em}{.1em}{.1em}
			& \multicolumn{4}{c||}{Chamfer ($ \times10^{-2} $)}     & \multicolumn{4}{c}{PSNR}             \\ \cline{2-9} 
			& Colmap & Vis-MVSNet & IDR   & MVSDF (Ours) & Colmap & Vis-MVSNet & IDR   & MVSDF (Ours)  \\ \hline
			Fountain-P11        & 6.35   & \textbf{6.12}       & 18.42 (7.88) & 6.84 & 20.17  & 24.33      & 24.58 (23.43) & \textbf{25.27} \\
			Herzjesu-P8         & 8.99   & 7.47       & 32.19 & \textbf{6.38} & 16.13  & 23.45      & 24.75 & \textbf{28.75} \\ \hline
			Mean                & 7.67   & 6.80       & 25.30 & \textbf{6.61} & 18.15  & 23.89      & 24.67 & \textbf{27.01} \\
			\specialrule{.2em}{.1em}{.1em}
	\end{tabular}}
	\caption{\textbf{Quantitative results on EPFL dataset.} Our method achieves the lowest mean Chamfer distance and the best PSNR score among all methods. Values in the parenthesis represent the results of IDR when the wall is excluded. }
	\label{tab:epfl}
	\vspace{-3mm}
\end{table*}

\fakepara{Mesh Extraction and Trimming}\label{sec:trim}
After network training, a mesh can be extracted from the SDF in a predefined bounding box by the Marching Cube algorithm \cite{lorensen1987marching} with volume size of $512^3$.  For scenes whose camera trajectory does not surround the object (e.g., DTU dataset), extrapolated surfaces would appear in background areas, and we propose to filter these areas according to the surface indicator described in Sec.~\ref{sec:indicator}.  We first evaluate the valid surface indicator for each mesh vertex, and assign each triangle an indicator score as the average score of its three vertices. Next, instead of deleting all triangles with low indicator scores, we propose a graph-cut based method to smoothly filter out those outlier surfaces. 

We define a graph $G = (V,E)$ over the mesh from Marching Cube, where each triangle represents a graph node $v \in V$ and each edge between two adjacent triangles represents a graph edge $e \in E$. A source and a sink node $s,t \in V$ are also defined. Triangles are linked to $s$ with edge weights 1 if their indicator scores are greater than $T_{trim} = 0.94$, or to $t$ otherwise. Adjacent triangles are also linked with weights 10 to encourage smoothness. After a min-cut of the constructed graph is obtained, the triangles linked with $t$ are removed. The proposed trimming algorithm can effectively filter out extrapolated background surfaces, as is illustrated in Fig.~\ref{fig:trim}.

\subsection{Benchmark on DTU Dataset}

We first evaluate our method on the \textit{DTU} MVS dataset. \textit{DTU} dataset contains 128 scans captured in laboratory. For each scan, there are 49 calibrated cameras located on front side of the upper sphere surrounding the captured object. In this paper, we evaluate both the surface mesh and the rendered image using the same set of scans as in \cite{yariv2020multiview}. 

The reconstructed mesh model is evaluated by the Chamfer distance to the ground truth point cloud, and the rendered image is evaluated using the PSNR score to the input image. We compare our method to 1) Colmap \cite{schonberger2016pixelwise}, which represents state-of-the-art traditional MVS algorithms; 2) Vis-MVSNet \cite{zhang2020visibility}, which represents state-of-the-art learning-based algorithms and 3) DVR \cite{niemeyer2020differentiable} and IDR \cite{yariv2020multiview}, which represent recent rendering based surface reconstruction methods. Depth maps from Colmap and Vis-MVSNet are fused into point clouds and converted into surface meshes by the screened Poisson surface reconstruction (sPSR)  \cite{kazhdan2013screened} with trim parameter 5. As Colmap and Vis-MVSNet do not estimate the surface texture, we follow \cite{yariv2020multiview} to assign color from input images when back projecting depth maps to point clouds.

\begin{figure*}
	\centering
	\begin{tabular}{@{\hskip2pt}c@{\hskip2pt}@{\hskip2pt}c@{\hskip2pt}@{\hskip2pt}c@{\hskip2pt}@{\hskip2pt}c@{\hskip2pt}@{\hskip2pt}c@{\hskip2pt}}
		\includegraphics[width=0.19\linewidth,trim={180 0 210 0},clip]{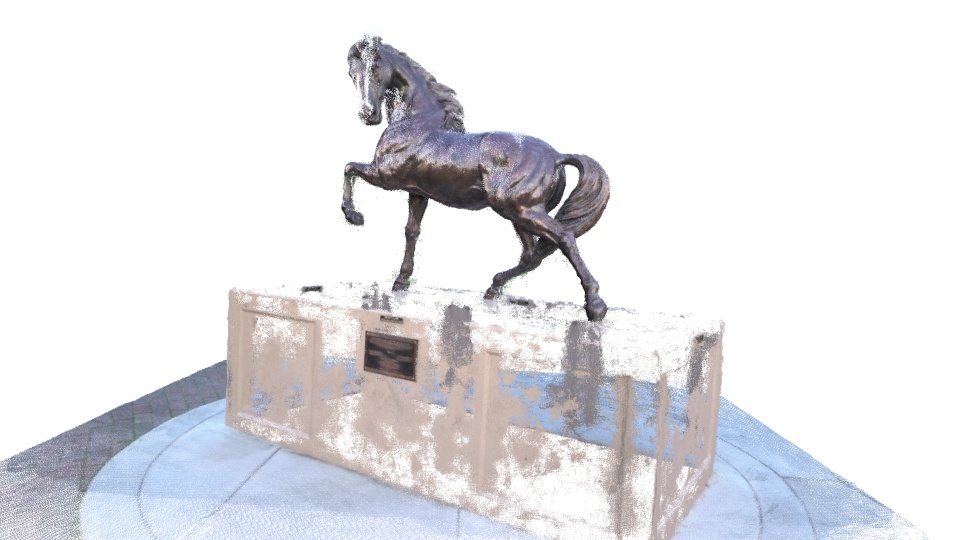} & 
		\includegraphics[width=0.19\linewidth,trim={180 0 210 0},clip]{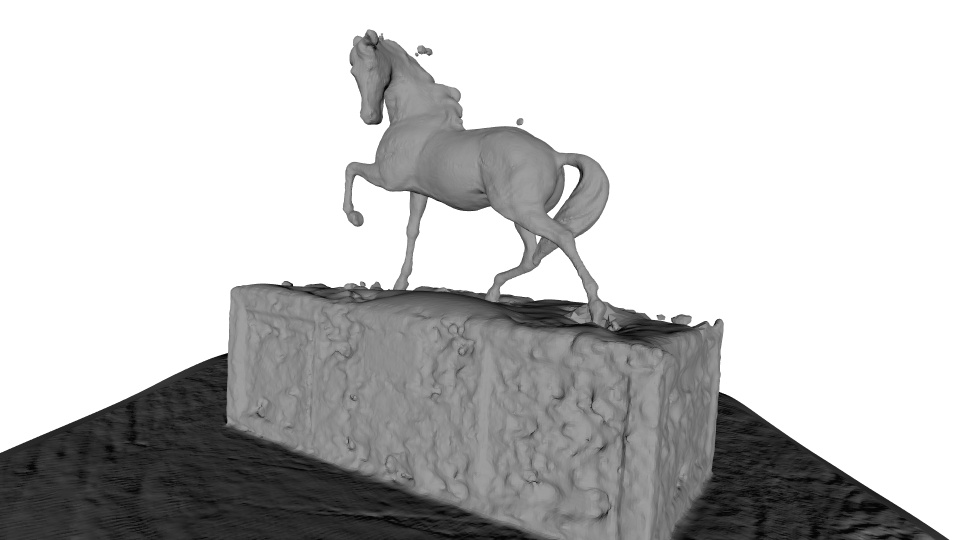} & 
		\includegraphics[width=0.19\linewidth,trim={180 0 210 0},clip]{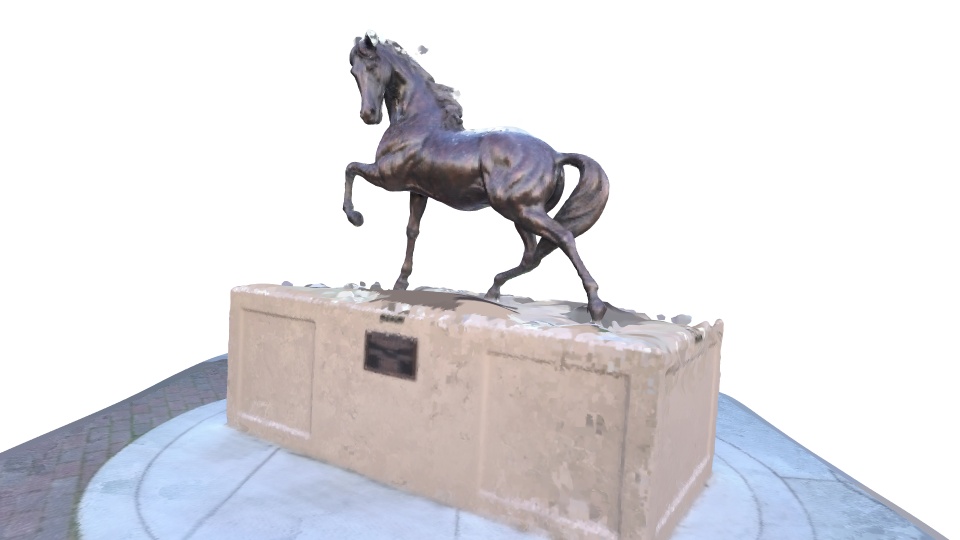} &
		\includegraphics[width=0.19\linewidth,trim={180 0 210 0},clip]{images_tnt_idrx_mesh_Horse_00000063_render.jpg} & 
		\includegraphics[width=0.19\linewidth,trim={120 0 140 0},clip]{images_tnt_idrx_Horse_eval_063.jpg} \\
		Vis-MVSNet Point Cloud & Vis-MVSNet Mesh & Vis-MVSNet Render & MVSDF (Ours) Mesh & MVSDF (Ours) Render
	\end{tabular}
	\caption{\textbf{Qualitative results on Tanks and Temples dataset.} Traditional methods produce holes in texture-less and reflective regions. In contrast, our end-to-end system is able to reconstruct accurate mesh and rendering results in these areas.}
	\label{fig:tnt}
	\vspace{-3mm}
\end{figure*}

\begin{figure}
	\centering
	\begin{tabular}{@{\hskip4pt}c@{\hskip4pt}@{\hskip4pt}c@{\hskip4pt}}
		\includegraphics[height=0.39\linewidth,trim={50 200 70 150},clip]{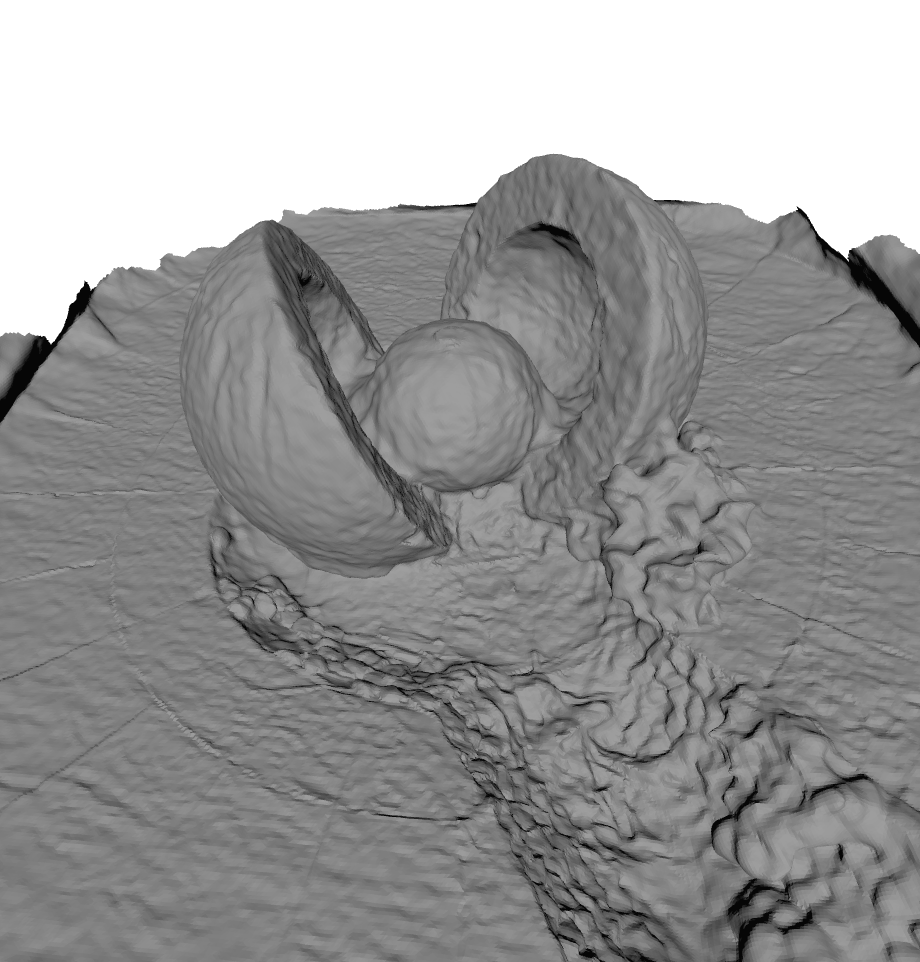} & \includegraphics[height=0.39\linewidth,trim={0 0 0 50},clip]{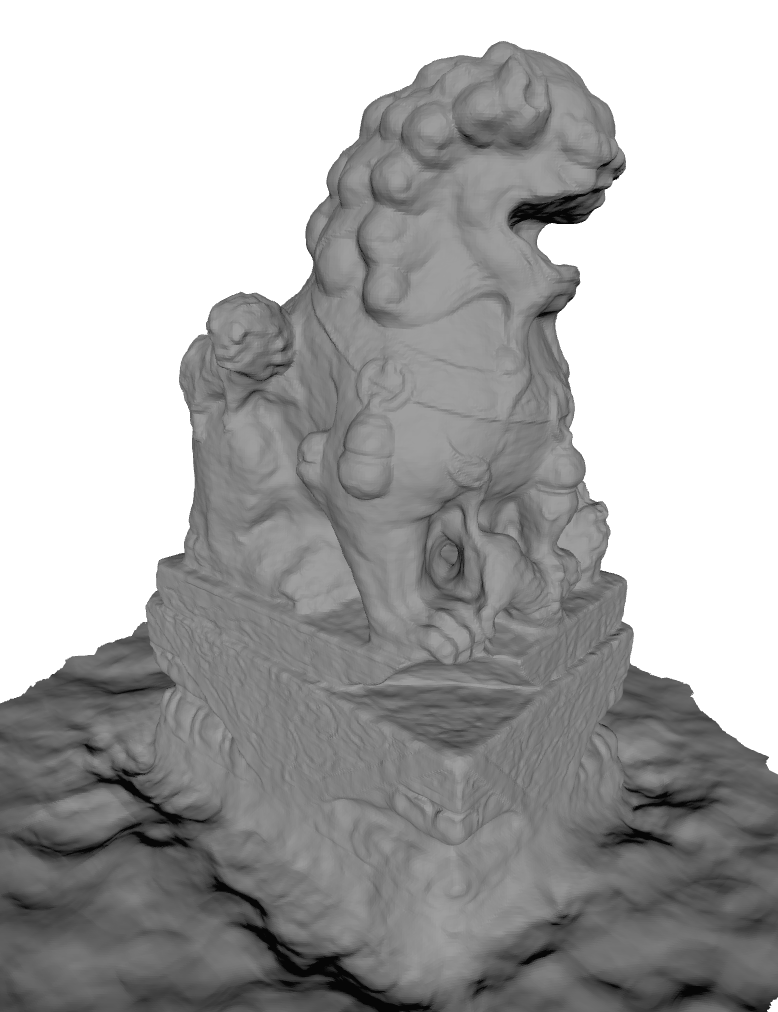} \\
		\includegraphics[height=0.39\linewidth,trim={50 200 70 150},clip]{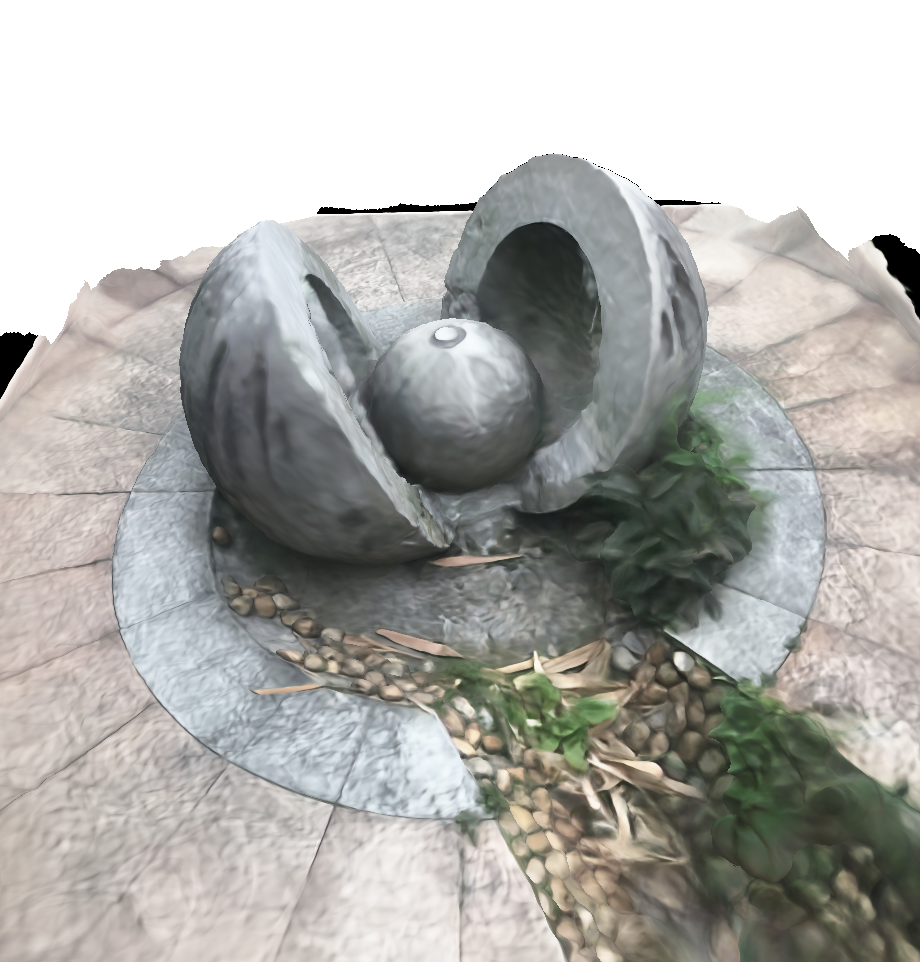} & \includegraphics[height=0.39\linewidth,trim={0 0 0 50},clip]{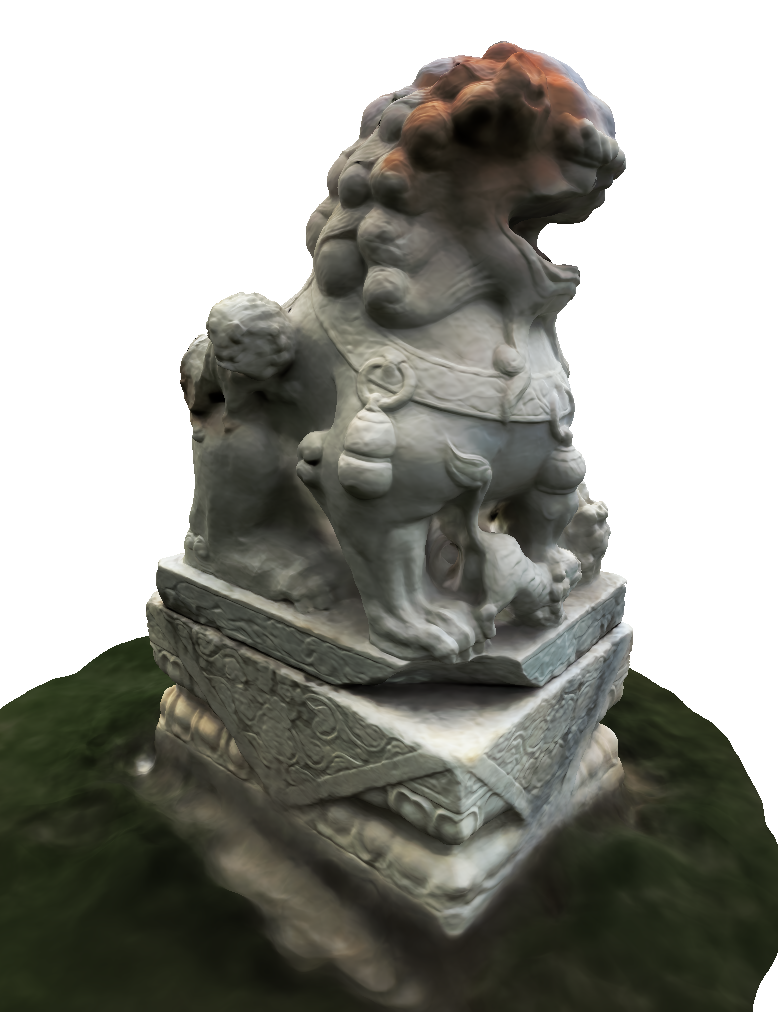}
	\end{tabular}
	\caption{\textbf{Qualitative results on BlendedMVS dataset.} Meshes (upper) and rendered images (lower) of two scenes reconstructed by MVSDF are illustrated. }
	\label{fig:bld}
	\vspace{-3mm}
\end{figure}

Quantitative results are shown in Tab.~\ref{tab:dtu}. The proposed method achieves the best mean Chamfer distance (0.88) and PSNR (25.92) among all methods. 
Qualitative results are shown in Fig.~\ref{fig:dtu}. Both our method and IDR are able to recover high-quality details in mesh surfaces. 
Compared to IDR, we have less distortions in reconstructed mesh surfaces.  It is also noteworthy that our results are reconstructed without any manual masks.

\subsection{Benchmark on EPFL Dataset}

Our method is also evaluated on the \textit{EPFL} dataset. \textit{EPFL} dataset contains 2 outdoor scenes, \textit{Fountain-P11} and \textit{Herzjesu-P8}, with ground truth meshes. We compare our method with Colmap, Vis-MVSNet and IDR. 
As \textit{Fountain-P11} and \textit{Herzjesu-P8} mainly consist of planar surfaces, they can not be well handled by the silhouette based methods. To fairly compare our method with IDR, for \textit{Fountain-P11}, we also test the case where the wall is excluded from the masks. 

Qualitative results are shown in Fig.~\ref{fig:epfl}. Similar to \textit{DTU}, our method is able to produce both high-quality mesh and renderings. 
The mesh from IDR contains inflated surfaces, which is more serious when the wall is included in input masks. The reason is that there is a large gap between the mask visual hull and the real surface so the solution may stuck at local minimum in this unrestricted area. In contrast, the topology is correctly recovered in our reconstruction.  
For quantitative results shown in Tab.~\ref{tab:epfl}, our method achieves the best mean Chamfer distance (6.61) and PSNR (27.01) among all methods. 

\subsection{Additional Qualitative Results}


We additionally provide qualitative results on \textit{Family} and \textit{Horse} in the \textit{Tanks and Temples} \cite{knapitsch2017tanks} (Fig.~\ref{fig:teaser},\ref{fig:tnt}) and two scenes in the \textit{BlendedMVS} \cite{yao2020blendedmvs} dataset (Fig.~\ref{fig:bld}). For \textit{Horse}, as the foundation part is highly texture-less and reflective, the estimated point cloud from Vis-MVSNet is incomplete. Although surfaces can be interpolated in some extent during the mesh reconstruction, the output mesh is bumpy and the rendered image is rather noisy. In contrast, the proposed method is able to produce complete and accurate mesh surface together with realistic view-dependent rendering. 

\subsection{Ablation Study}

In this section, we discuss how different losses in the network would affect the final geometry reconstruction. The following three settings are tested using the \textit{DTU} dataset: 1) \textit{no feature}: the feature loss is removed from the network by setting $w_F = 0$; 2) \textit{no render}: the render loss is removed by setting $w_R = 0$ and 3) \textit{distance only}: both the feature loss and the render loss are disabled $w_R = w_F = 0$.

\begin{figure}
	\centering
	\begin{tabular}{@{\hskip4pt}c@{\hskip4pt}@{\hskip4pt}c@{\hskip4pt}}
		\begin{tikzpicture}\node[above right, inner sep=0](image) at (0,0) {\includegraphics[height=0.3375\linewidth]{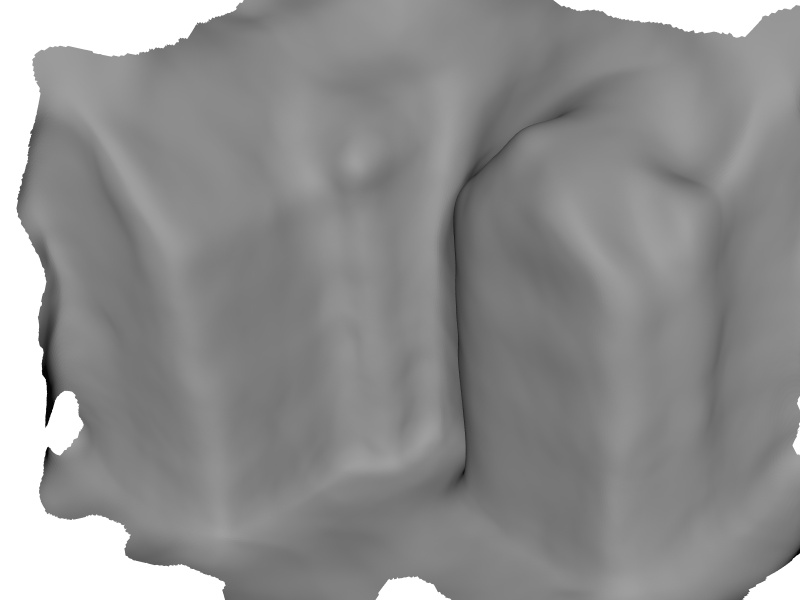}}; \end{tikzpicture} &
		\begin{tikzpicture}\node[above right, inner sep=0](image) at (0,0) {\includegraphics[height=0.3375\linewidth]{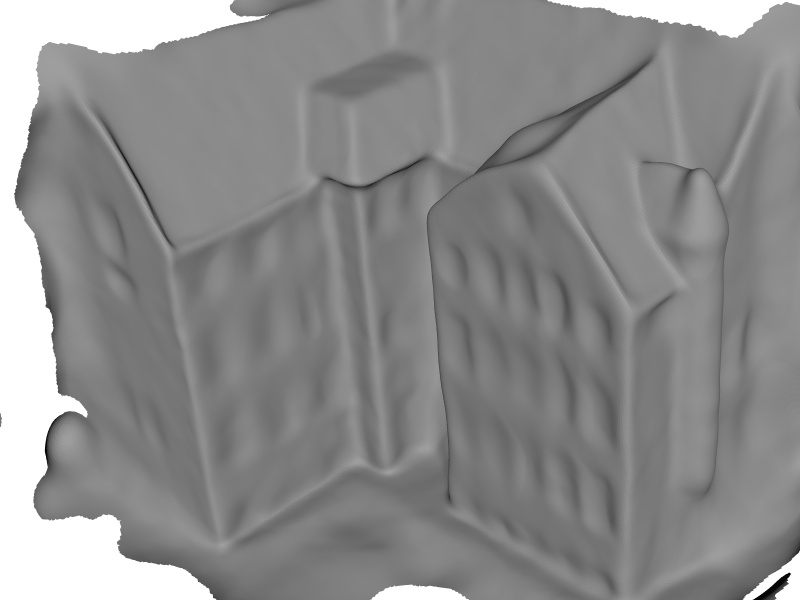}}; \end{tikzpicture} \\
		
		distance only & no render \\
		
		\begin{tikzpicture}\node[above right, inner sep=0](image) at (0,0) {\includegraphics[height=0.3375\linewidth]{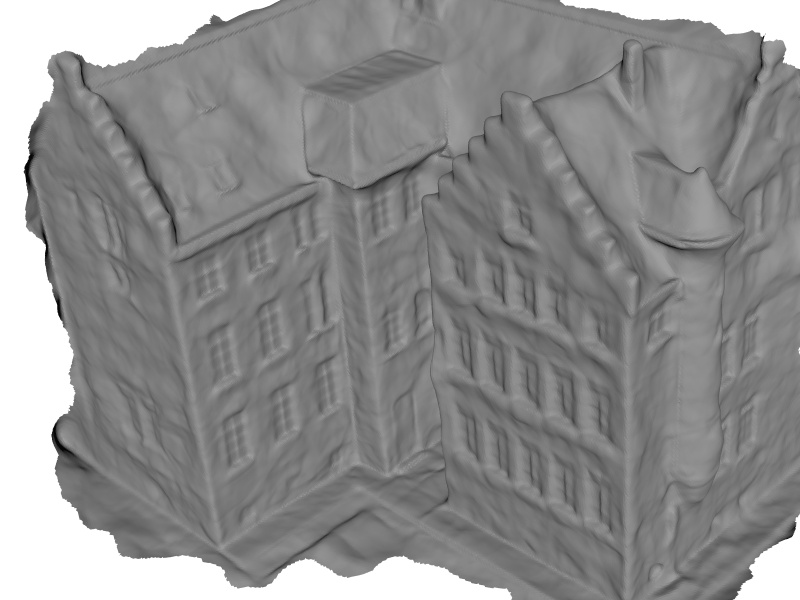}};\draw[very thick,red] (0.9,1.9) rectangle (1.9,2.8); \end{tikzpicture} &
		\begin{tikzpicture}\node[above right, inner sep=0](image) at (0,0) {\includegraphics[height=0.3375\linewidth]{images_dtu_idrx2_mesh_024_00000023_render.jpg}};\draw[very thick,green] (0.9,1.9) rectangle (1.9,2.8); \end{tikzpicture} \\
		
		no feature & full
	\end{tabular}
	\caption{\textbf{Ablation study on training losses.} Both the feature loss and the render loss are able to refine the surface geometry. Moreover, the reconstruction is more robust if the feature loss is applied.}
	\label{fig:abl}
	\vspace{-3mm}
\end{figure}

Mesh surface results of different settings are illustrated in Fig.~\ref{fig:abl}. We find that the feature loss (from \textit{distance only} to \textit{no render}) can successfully refine the surface, but is still coarse compared to the \textit{full} setting. The render loss is able to refine the model to its finest detail level (from \textit{distance only} to \textit{no feature}), however, it is not as robust as the feature loss and will causes erroneous surface in the roof area. Quantitative results are shown in Tab.~\ref{tab:abl}. Both the feature loss and the render loss can significantly boost the reconstruction quality, showing the effectiveness of each component of the proposed method.

\begin{table}[]
	\centering
	\resizebox{\linewidth}{!}{%
		\begin{tabular}{l|ccc|c}
			\specialrule{.2em}{.1em}{.1em}
			& $ L_D $ & $ L_F $ & $ L_R $ & Mean Chamfer (mm) \\ \hline
			distance only & $ \checkmark $  &    &    & 3.56         \\
			no render  & $ \checkmark $  & $ \checkmark $  &    & 1.75         \\
			no feature & $ \checkmark $  &    & $ \checkmark $  & 1.06         \\
			full       & $ \checkmark $  & $ \checkmark $  & $ \checkmark $  & 0.88         \\
			\specialrule{.2em}{.1em}{.1em}
	\end{tabular}}
	\caption{\textbf{Quantitative results of ablation studies on DTU dataset.} Both the feature loss and the render loss can improvement the mesh reconstruction quality.}
	\label{tab:abl}
	\vspace{-3mm}
\end{table}

\section{Conclusion}

In this work, we introduce a novel neural surface reconstruction framework that combines implicit neural surface estimation with recent advanced MVS networks.  In our network, the geometry and appearance are represented as neural implicit functions by MLPs. The geometry is directly supervised by MVS depth maps to recover the surface topology and is locally refined via deep feature consistency and rendered image loss. The proposed method has been extensively evaluated on different datasets.  Both qualitative and quantitative results have shown that our method outperforms previous methods in terms of both geometry accuracy and rendering fidelity, demonstrating the effectiveness of the proposed framework.

\section{Acknowledgments}
This work is supported by Hong Kong RGC GRF 16206819, 16203518, T22-603/15N and Guangzhou Okay Information Technology with the project GZETDZ18EG05.


{\small
\bibliographystyle{ieee_fullname}
\bibliography{egbib_final}

\begin{thebibliography}{10}\itemsep=-1pt

\bibitem{bleyer2011patchmatch}
Michael Bleyer, Christoph Rhemann, and Carsten Rother.
\newblock Patchmatch stereo-stereo matching with slanted support windows.
\newblock In {\em British Machine Vision Conference (BMVC)}, 2011.

\bibitem{campbell2008using}
Neill~DF Campbell, George Vogiatzis, Carlos Hern{\'a}ndez, and Roberto Cipolla.
\newblock Using multiple hypotheses to improve depth-maps for multi-view
  stereo.
\newblock In {\em European Conference on Computer Vision (ECCV)}, 2008.

\bibitem{collins1996space}
Robert~T Collins.
\newblock A space-sweep approach to true multi-image matching.
\newblock In {\em Computer Vision and Pattern Recognition (CVPR)}, 1996.

\bibitem{delaunoy2014photometric}
Ama{\"e}l Delaunoy and Marc Pollefeys.
\newblock Photometric bundle adjustment for dense multi-view 3d modeling.
\newblock In {\em Computer Vision and Pattern Recognition (CVPR)}, 2014.

\bibitem{delaunoy2011gradient}
Ama{\"e}l Delaunoy and Emmanuel Prados.
\newblock Gradient flows for optimizing triangular mesh-based surfaces:
  Applications to 3d reconstruction problems dealing with visibility.
\newblock {\em International Journal of Computer Vision}, 95(2):100--123, 2011.

\bibitem{furukawa2015multi}
Yasutaka Furukawa and Carlos Hern{\'a}ndez.
\newblock Multi-view stereo: A tutorial.
\newblock {\em Foundations and Trends{\textregistered} in Computer Graphics and
  Vision}, 9(1-2):1--148, 2015.

\bibitem{furukawa2006carved}
Yasutaka Furukawa and Jean Ponce.
\newblock Carved visual hulls for image-based modeling.
\newblock In {\em European Conference on Computer Vision (ECCV)}, 2006.

\bibitem{furukawa2009accurate}
Yasutaka Furukawa and Jean Ponce.
\newblock Accurate, dense, and robust multiview stereopsis.
\newblock {\em IEEE transactions on Pattern Analysis and Machine Intelligence},
  32(8):1362--1376, 2009.

\bibitem{galliani2015massively}
Silvano Galliani, Katrin Lasinger, and Konrad Schindler.
\newblock Massively parallel multiview stereopsis by surface normal diffusion.
\newblock In {\em International Conference on Computer Vision (ICCV)}, 2015.

\bibitem{goesele2006multi}
Michael Goesele, Brian Curless, and Steven~M Seitz.
\newblock Multi-view stereo revisited.
\newblock In {\em Computer Vision and Pattern Recognition (CVPR)}, 2006.

\bibitem{gropp2020implicit}
Amos Gropp, Lior Yariv, Niv Haim, Matan Atzmon, and Yaron Lipman.
\newblock Implicit geometric regularization for learning shapes.
\newblock In {\em International Conference on Machine Learning (ICML)}, 2020.

\bibitem{gu2020cascade}
Xiaodong Gu, Zhiwen Fan, Siyu Zhu, Zuozhuo Dai, Feitong Tan, and Ping Tan.
\newblock Cascade cost volume for high-resolution multi-view stereo and stereo
  matching.
\newblock In {\em Computer Vision and Pattern Recognition (CVPR)}, 2020.

\bibitem{huang2018deepmvs}
Po-Han Huang, Kevin Matzen, Johannes Kopf, Narendra Ahuja, and Jia-Bin Huang.
\newblock Deepmvs: Learning multi-view stereopsis.
\newblock In {\em Computer Vision and Pattern Recognition (CVPR)}, 2018.

\bibitem{jensen2014large}
Rasmus Jensen, Anders Dahl, George Vogiatzis, Engil Tola, and Henrik Aan{\ae}s.
\newblock Large scale multi-view stereopsis evaluation.
\newblock In {\em Computer Vision and Pattern Recognition (CVPR)}, 2014.

\bibitem{kar2017learning}
Abhishek Kar, Christian H{\"a}ne, and Jitendra Malik.
\newblock Learning a multi-view stereo machine.
\newblock In {\em Neural Information Processing Systems (NIPS)}, 2017.

\bibitem{kato2020differentiable}
Hiroharu Kato, Deniz Beker, Mihai Morariu, Takahiro Ando, Toru Matsuoka, Wadim
  Kehl, and Adrien Gaidon.
\newblock Differentiable rendering: A survey.
\newblock {\em arXiv preprint arXiv:2006.12057}, 2020.

\bibitem{kazhdan2013screened}
Michael Kazhdan and Hugues Hoppe.
\newblock Screened poisson surface reconstruction.
\newblock {\em ACM Transactions on Graphics (ToG)}, 32(3):1--13, 2013.

\bibitem{knapitsch2017tanks}
Arno Knapitsch, Jaesik Park, Qian-Yi Zhou, and Vladlen Koltun.
\newblock Tanks and temples: Benchmarking large-scale scene reconstruction.
\newblock {\em ACM Transactions on Graphics (ToG)}, 36(4):78, 2017.

\bibitem{kuhn2020deepc}
Andreas Kuhn, Christian Sormann, Mattia Rossi, Oliver Erdler, and Friedrich
  Fraundorfer.
\newblock Deepc-mvs: Deep confidence prediction for multi-view stereo
  reconstruction.
\newblock In {\em International Conference on 3D Vision (3DV)}, 2020.

\bibitem{labatut2007efficient}
Patrick Labatut, Jean-Philippe Pons, and Renaud Keriven.
\newblock Efficient multi-view reconstruction of large-scale scenes using
  interest points, delaunay triangulation and graph cuts.
\newblock In {\em International Conference on Computer Vision (ICCV)}, 2007.

\bibitem{lhuillier2005quasi}
Maxime Lhuillier and Long Quan.
\newblock A quasi-dense approach to surface reconstruction from uncalibrated
  images.
\newblock {\em IEEE transactions on Pattern Analysis and Machine Intelligence},
  27(3):418--433, 2005.

\bibitem{li2016efficient}
Shiwei Li, Sing~Yu Siu, Tian Fang, and Long Quan.
\newblock Efficient multi-view surface refinement with adaptive resolution
  control.
\newblock In {\em European Conference on Computer Vision (ECCV)}, 2016.

\bibitem{liu2020neural}
Lingjie Liu, Jiatao Gu, Kyaw~Zaw Lin, Tat-Seng Chua, and Christian Theobalt.
\newblock Neural sparse voxel fields.
\newblock In {\em Neural Information Processing Systems (NeurIPS)}, 2020.

\bibitem{liu2019learning}
Shichen Liu, Shunsuke Saito, Weikai Chen, and Hao Li.
\newblock Learning to infer implicit surfaces without 3d supervision.
\newblock In {\em Neural Information Processing Systems (NeurIPS)}, 2019.

\bibitem{liu2020dist}
Shaohui Liu, Yinda Zhang, Songyou Peng, Boxin Shi, Marc Pollefeys, and Zhaopeng
  Cui.
\newblock Dist: Rendering deep implicit signed distance function with
  differentiable sphere tracing.
\newblock In {\em Computer Vision and Pattern Recognition (CVPR)}, 2020.

\bibitem{lorensen1987marching}
William~E Lorensen and Harvey~E Cline.
\newblock Marching cubes: A high resolution 3d surface construction algorithm.
\newblock {\em ACM siggraph computer graphics}, 21(4):163--169, 1987.

\bibitem{martin2020nerf}
Ricardo Martin-Brualla, Noha Radwan, Mehdi~SM Sajjadi, Jonathan~T Barron,
  Alexey Dosovitskiy, and Daniel Duckworth.
\newblock Nerf in the wild: Neural radiance fields for unconstrained photo
  collections.
\newblock In {\em Computer Vision and Pattern Recognition (CVPR)}, 2020.

\bibitem{mescheder2019occupancy}
Lars Mescheder, Michael Oechsle, Michael Niemeyer, Sebastian Nowozin, and
  Andreas Geiger.
\newblock Occupancy networks: Learning 3d reconstruction in function space.
\newblock In {\em Computer Vision and Pattern Recognition (CVPR)}, 2019.

\bibitem{mildenhall2020nerf}
Ben Mildenhall, Pratul~P Srinivasan, Matthew Tancik, Jonathan~T Barron, Ravi
  Ramamoorthi, and Ren Ng.
\newblock Nerf: Representing scenes as neural radiance fields for view
  synthesis.
\newblock In {\em European Conference on Computer Vision (ECCV)}, 2020.

\bibitem{niemeyer2020differentiable}
Michael Niemeyer, Lars Mescheder, Michael Oechsle, and Andreas Geiger.
\newblock Differentiable volumetric rendering: Learning implicit 3d
  representations without 3d supervision.
\newblock In {\em Computer Vision and Pattern Recognition (CVPR)}, 2020.

\bibitem{Park_2019_CVPR}
Jeong~Joon Park, Peter Florence, Julian Straub, Richard Newcombe, and Steven
  Lovegrove.
\newblock Deepsdf: Learning continuous signed distance functions for shape
  representation.
\newblock In {\em Computer Vision and Pattern Recognition (CVPR)}, 2019.

\bibitem{paschalidou2018raynet}
Despoina Paschalidou, Osman Ulusoy, Carolin Schmitt, Luc Van~Gool, and Andreas
  Geiger.
\newblock Raynet: Learning volumetric 3d reconstruction with ray potentials.
\newblock In {\em Computer Vision and Pattern Recognition (CVPR)}, 2018.

\bibitem{peng2020convolutional}
Songyou Peng, Michael Niemeyer, Lars Mescheder, Marc Pollefeys, and Andreas
  Geiger.
\newblock Convolutional occupancy networks.
\newblock In {\em European Conference on Computer Vision (ECCV)}, 2020.

\bibitem{pons2007multi}
Jean-Philippe Pons, Renaud Keriven, and Olivier Faugeras.
\newblock Multi-view stereo reconstruction and scene flow estimation with a
  global image-based matching score.
\newblock {\em International Journal of Computer Vision}, 72(2):179--193, 2007.

\bibitem{qin2019basnet}
Xuebin Qin, Zichen Zhang, Chenyang Huang, Chao Gao, Masood Dehghan, and Martin
  Jagersand.
\newblock Basnet: Boundary-aware salient object detection.
\newblock In {\em Computer Vision and Pattern Recognition (CVPR)}, 2019.

\bibitem{saito2019pifu}
Shunsuke Saito, Zeng Huang, Ryota Natsume, Shigeo Morishima, Angjoo Kanazawa,
  and Hao Li.
\newblock Pifu: Pixel-aligned implicit function for high-resolution clothed
  human digitization.
\newblock In {\em International Conference on Computer Vision (ICCV)}, 2019.

\bibitem{schonberger2016pixelwise}
Johannes~L Sch{\"o}nberger, Enliang Zheng, Jan-Michael Frahm, and Marc
  Pollefeys.
\newblock Pixelwise view selection for unstructured multi-view stereo.
\newblock In {\em European Conference on Computer Vision (ECCV)}, 2016.

\bibitem{seitz2006comparison}
Steven~M Seitz, Brian Curless, James Diebel, Daniel Scharstein, and Richard
  Szeliski.
\newblock A comparison and evaluation of multi-view stereo reconstruction
  algorithms.
\newblock In {\em Computer Vision and Pattern Recognition (CVPR)}, 2006.

\bibitem{sitzmann2019scene}
Vincent Sitzmann, Michael Zollh{\"o}fer, and Gordon Wetzstein.
\newblock Scene representation networks: Continuous 3d-structure-aware neural
  scene representations.
\newblock In {\em Neural Information Processing Systems (NeurIPS)}, 2019.

\bibitem{strecha2008benchmarking}
Christoph Strecha, Wolfgang Von~Hansen, Luc Van~Gool, Pascal Fua, and Ulrich
  Thoennessen.
\newblock On benchmarking camera calibration and multi-view stereo for high
  resolution imagery.
\newblock In {\em Computer Vision and Pattern Recognition (CVPR)}, 2008.

\bibitem{tola2012efficient}
Engin Tola, Christoph Strecha, and Pascal Fua.
\newblock Efficient large-scale multi-view stereo for ultra high-resolution
  image sets.
\newblock {\em Machine Vision and Applications}, 23(5):903--920, 2012.

\bibitem{vu2011high}
Hoang-Hiep Vu, Patrick Labatut, Jean-Philippe Pons, and Renaud Keriven.
\newblock High accuracy and visibility-consistent dense multiview stereo.
\newblock {\em IEEE transactions on Pattern Analysis and Machine Intelligence},
  34(5):889--901, 2011.

\bibitem{wang2020patchmatchnet}
Fangjinhua Wang, Silvano Galliani, Christoph Vogel, Pablo Speciale, and Marc
  Pollefeys.
\newblock Patchmatchnet: Learned multi-view patchmatch stereo.
\newblock In {\em Computer Vision and Pattern Recognition (CVPR)}, 2021.

\bibitem{xu2019multi}
Qingshan Xu and Wenbing Tao.
\newblock Multi-scale geometric consistency guided multi-view stereo.
\newblock In {\em Computer Vision and Pattern Recognition (CVPR)}, 2019.

\bibitem{xue2019mvscrf}
Youze Xue, Jiansheng Chen, Weitao Wan, Yiqing Huang, Cheng Yu, Tianpeng Li, and
  Jiayu Bao.
\newblock Mvscrf: Learning multi-view stereo with conditional random fields.
\newblock In {\em International Conference on Computer Vision (ICCV)}, 2019.

\bibitem{yao2018mvsnet}
Yao Yao, Zixin Luo, Shiwei Li, Tian Fang, and Long Quan.
\newblock Mvsnet: Depth inference for unstructured multi-view stereo.
\newblock In {\em European Conference on Computer Vision (ECCV)}, 2018.

\bibitem{yao2019recurrent}
Yao Yao, Zixin Luo, Shiwei Li, Tianwei Shen, Tian Fang, and Long Quan.
\newblock Recurrent mvsnet for high-resolution multi-view stereo depth
  inference.
\newblock In {\em Computer Vision and Pattern Recognition (CVPR)}, 2019.

\bibitem{yao2020blendedmvs}
Yao Yao, Zixin Luo, Shiwei Li, Jingyang Zhang, Yufan Ren, Lei Zhou, Tian Fang,
  and Long Quan.
\newblock Blendedmvs: A large-scale dataset for generalized multi-view stereo
  networks.
\newblock In {\em Computer Vision and Pattern Recognition (CVPR)}, 2020.

\bibitem{yariv2020multiview}
Lior Yariv, Yoni Kasten, Dror Moran, Meirav Galun, Matan Atzmon, Basri Ronen,
  and Yaron Lipman.
\newblock Multiview neural surface reconstruction by disentangling geometry and
  appearance.
\newblock In {\em Neural Information Processing Systems (NeurIPS)}, 2020.

\bibitem{yu2020fast}
Zehao Yu and Shenghua Gao.
\newblock Fast-mvsnet: Sparse-to-dense multi-view stereo with learned
  propagation and gauss-newton refinement.
\newblock In {\em Computer Vision and Pattern Recognition (CVPR)}, 2020.

\bibitem{zhang2020visibility}
Jingyang Zhang, Yao Yao, Shiwei Li, Zixin Luo, and Tian Fang.
\newblock Visibility-aware multi-view stereo network.
\newblock In {\em British Machine Vision Conference (BMVC)}, 2020.

\bibitem{zhang2020nerf++}
Kai Zhang, Gernot Riegler, Noah Snavely, and Vladlen Koltun.
\newblock Nerf++: Analyzing and improving neural radiance fields.
\newblock {\em arXiv preprint arXiv:2010.07492}, 2020.

\bibitem{zhao2019pyramid}
Ting Zhao and Xiangqian Wu.
\newblock Pyramid feature attention network for saliency detection.
\newblock In {\em Computer Vision and Pattern Recognition (CVPR)}, 2019.

\bibitem{Zhou2018}
Qian-Yi Zhou, Jaesik Park, and Vladlen Koltun.
\newblock {Open3D}: {A} modern library for {3D} data processing.
\newblock {\em arXiv:1801.09847}, 2018.

\end{thebibliography}
}

\newpage

\twocolumn[
\begin{center}
	\textbf{\LARGE Supplemental Materials}
	\vspace{5mm}
\end{center}
]
\setcounter{section}{0}
\setcounter{equation}{0}
\setcounter{figure}{0}
\setcounter{table}{0}
\setcounter{page}{1}
\makeatletter

\section{Implementation Details}
We first provide additional implementation details that are not discussed in the main paper due to the space limit.
\subsection{Baselines}
\paragraph{Vis-MVSNet}
We use the official Vis-MVSNet implementation \cite{zhang2020visibility} in our experiments. In MVS step, the source image number is set to 2 and the depth sample number is set to 256. Output depth map size is set to $512 \times 384$ for \textit{DTU}, $600 \times 400$ for \textit{EPFL} and $640 \times 360$ for \textit{Tanks and Temples}. In depth filtering step, the probability thresholds are set to $ 0.9, 0.7, 0.3 $ for \textit{DTU} and \textit{EPFL} and $ 0.9, 0.9, 0.8 $ for \textit{Tanks and Temples}. The number of geometric consistency is 4 for \textit{DTU}, 3 for \textit{EPFL} and 5 for \textit{Tanks and Temples}. 

In order to use sPSR for mesh reconstruction, normal maps are additionally computed from filtered depth maps of Vis-MVSNet. Fused point clouds of both Vis-MVSNet and Colmap are further clipped by the bounding box used in MVSDF, and meshes are reconstructed by sPSR with octree depth as 9 and trim parameter as 5. The implementation of sPSR is provided by Open3D \cite{Zhou2018}.

\vspace{-4mm}\paragraph{IDR}
In order to test IDR on the \textit{EPFL} dataset, we manually create the image masks and use the same bounding box as in MVSDF. We use the official IDR implementation in our experiments \cite{yariv2020multiview}. The network is trained by 10000 epochs. The scheduling of learning rate and alpha value is also scaled accordingly. 

\subsection{MVSDF}
\paragraph{MVS Module}
In our MVS Module, parameters of source image number, depth sample number, output depth map size and probability thresholds are all set to the same as Vis-MVSNet baseline. For the feature loss, we use deep image feature maps from the last scale, whose size is the same as the final depth map. 

\vspace{-4mm}\paragraph{Loss Weights}

During training, weights of Eikonal loss, indicator loss and render loss are all fixed to $ w_E, w_P, w_R = [0.1, 0.01, 0.5] $. Weights of the distance loss and feature loss will be changed in different training stages: 1) in the first $ 1/6 $ of the training, $w_D = 1.0$ and $w_F = 0$; 2) in the next $1/3$, $w_D = w_F = 0.1$; 3) in the last $ 1/2 $, $w_D = w_F = 0.01$. 

We observe that optimization by feature loss and render loss may diverge in the second and third stage of our training process (see the \textbf{Training} part in Sec. 4.1 in the main paper). To ensure that the surface can only be locally refined within certain range, we only decrease the weight of distance loss for sample points whose absolute value of calculated distance $ |l(\mathbf{x})| $ is less than $ 5\% $ of the bounding box size.  


\subsection{Evaluations}
\paragraph{Chamfer Distance}
For \textit{DTU} dataset, the Chamfer distance is calculated by the provided MATLAB code \cite{jensen2014large}. For \textit{EPFL} dataset, we use our own evaluation script. First, we crop the ground truth mesh by the manual image masks. Then both input mesh and ground truth mesh are sampled to point clouds by a predefined sample number. The reported value is the average of Chamfer distance from the input to the ground truth and also from the ground truth to the input. For both directions, we excluded points with distance larger than 0.8. 

\vspace{-4mm}\paragraph{PSNR}
We only evaluate PSNR using pixels located in predefined masks. For \textit{DTU}, we use the perfect mask provided by IDR \cite{yariv2020multiview}. For the other two datasets, we gather the render mask from all methods and take the intersection of them as the predefined mask. 

\begin{table}[]
	\centering
	\resizebox{\linewidth}{!}{%
		\begin{tabular}{l|cc|cccc}
			\specialrule{.2em}{.1em}{.1em}
			& lowres & filtered & distance only & no render & no feature & full \\ \hline
			24   & 0.83   & \textbf{0.79}     & 3.48          & 1.53      & 1.02       & 0.83 \\
			37   & \textbf{1.35}   & 1.65     & 5.67          & 4.38      & 1.80       & 1.76 \\
			40   & 1.11   & \textbf{0.85}     & 3.73          & 1.59      & 1.09       & 0.88 \\
			55   & 0.46   & 0.45     & 2.85          & 0.68      & 0.58       & \textbf{0.44} \\
			63   & 1.22   & \textbf{1.05}     & 3.64          & 1.99      & 1.65       & 1.11 \\
			65   & 1.13   & 1.06     & 4.27          & 2.09      & 1.18       & \textbf{0.90} \\
			69   & 0.84   & 0.80     & 2.76          & 1.15      & 0.80       & \textbf{0.75} \\
			83   & 1.30   & 1.30     & 3.95          & 2.72      & 1.71       & \textbf{1.26} \\
			97   & 1.06   & \textbf{0.98}     & 3.14          & 1.33      & 1.23       & 1.02 \\
			105  & \textbf{1.11}   & 1.15     & 4.64          & 3.01      & 1.31       & 1.35 \\
			106  & \textbf{0.77}   & 1.01     & 3.49          & 1.25      & 0.95       & 0.87 \\
			110  & 0.81   & \textbf{0.71}     & 3.40          & 1.74      & 0.95       & 0.84 \\
			114  & 0.35   & 0.35     & 1.81          & 0.53      & 0.37       & \textbf{0.34} \\
			118  & 0.48   & 0.52     & 3.14          & 1.07      & 0.62       & \textbf{0.47} \\
			122  & 0.49   & 0.53     & 3.37          & 1.22      & 0.65       & \textbf{0.46} \\ \hline
			Mean & 0.89   & \textbf{0.88}     & 3.56          & 1.75      & 1.06       & \textbf{0.88} \\
			\specialrule{.2em}{.1em}{.1em}
	\end{tabular}}
	\caption{Quantitative results of ablation study on DTU dataset.  The proposed method could consistently generate high quality reconstructions in spite of the depth map quality of the MVS module.}
	\label{tab:abl_supp}
\end{table}

\section{Ablation Study on MVS module}

We additionally study on how would the MVS depth map quality affect the geometry accuracy of MVSDF. The following two settings are tested on \textit{DTU} dataset: 1) \textit{lowres}: input depth maps are down sampled to $256\times192$, which represents a MVS module of lower quality and 2) \textit{filtered}: input depth maps are precomputed and geometrically filtered as in \cite{yao2018mvsnet}, which represents a MVS module of higher accuracy. Quantitative results are shown in Tab.~\ref{tab:abl_supp}. We find that both \textit{lowres} and \textit{filtered} settings generates similar results to the proposed setting \textit{full}. In fact, the MVS module is mainly used for recovering the correct initial surface topology, and it could be switched to other MVS algorithms if necessary.

\section{Tanks and Temples Dataset}
We additionally conduct experiments on \textit{Francis}, \textit{M60} and \textit{Panther} of the \textit{Tanks and Temples} dataset and report the PSNR scores of Colmap, Vis-MVSNet and the proposed method in Tab.~\ref{tab:tnt}.  According to the result, our method consistently outperforms other methods on the rendered image quality.

\begin{table}[]
	\centering
	\begin{tabular}{l|ccc}
		\specialrule{.2em}{.1em}{.1em}
		& Colmap & Vis-MVSNet & MVSDF (Ours) \\ \hline
		Family  & 21.50  & 22.09      & \textbf{26.11}        \\
		Francis & 18.25  & 20.07      & \textbf{25.58}        \\
		Horse   & 18.62  & 18.25      & \textbf{26.43}        \\
		M60     & 17.46  & 17.41      & \textbf{20.64}        \\
		Panther & 19.73  & 19.99      & \textbf{23.93}        \\ \hline
		Mean    & 19.11  & 19.56      & \textbf{24.54}       \\
		\specialrule{.2em}{.1em}{.1em}
	\end{tabular}
	\caption{Quantitative results on Tanks and Temples dataset.}
	\label{tab:tnt}
\end{table}

\section{Additional Qualitative Results}

\begin{figure*}
	\centering
	\begin{tabular}{@{\hskip2pt}c@{\hskip2pt}@{\hskip2pt}c@{\hskip2pt}@{\hskip2pt}c@{\hskip2pt}@{\hskip2pt}c@{\hskip2pt}@{\hskip2pt}c@{\hskip2pt}}
		\includegraphics[width=0.19\linewidth]{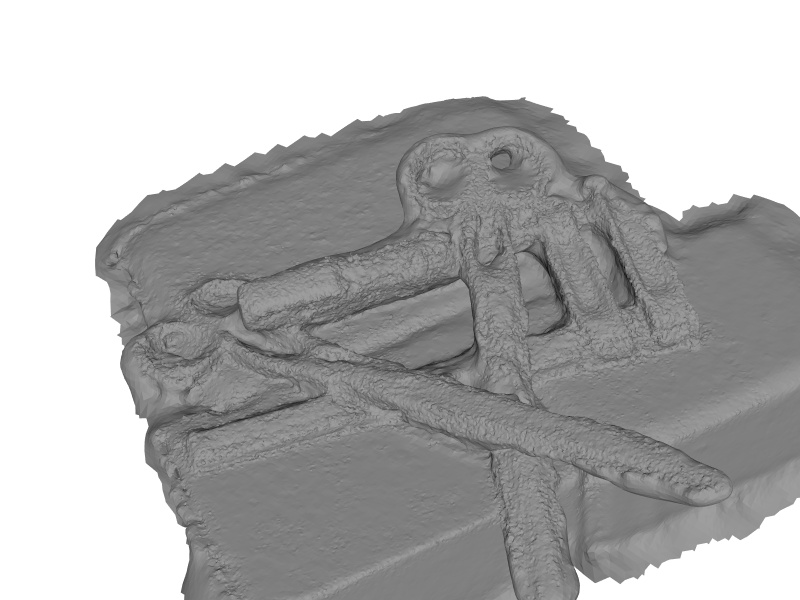} &
		\includegraphics[width=0.19\linewidth]{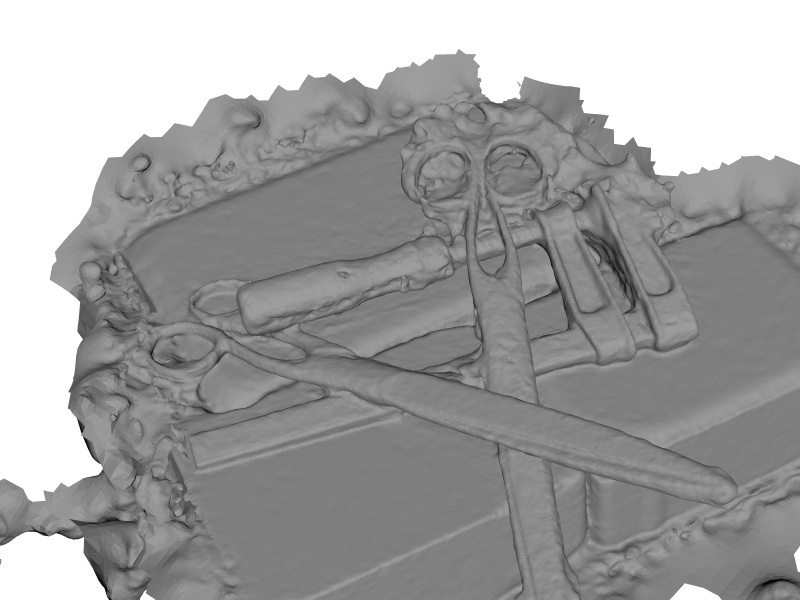} &
		\includegraphics[width=0.19\linewidth]{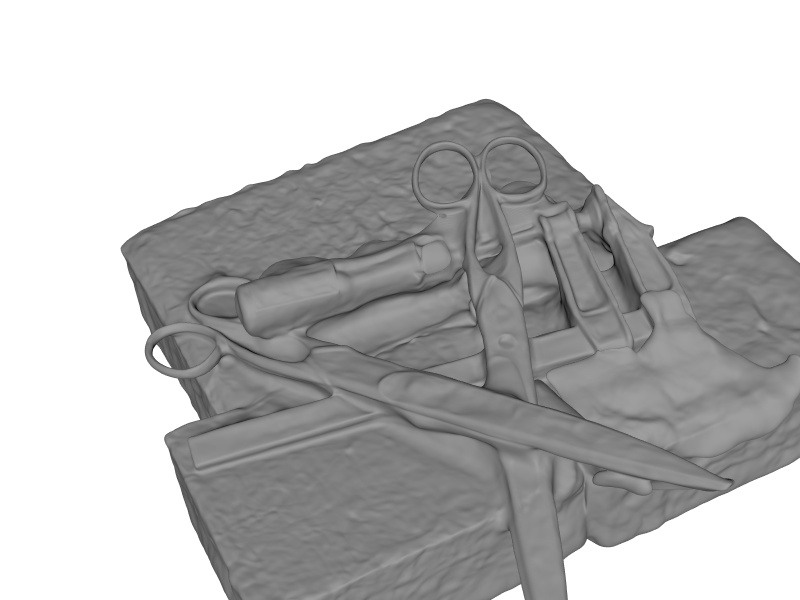} &
		\includegraphics[width=0.19\linewidth]{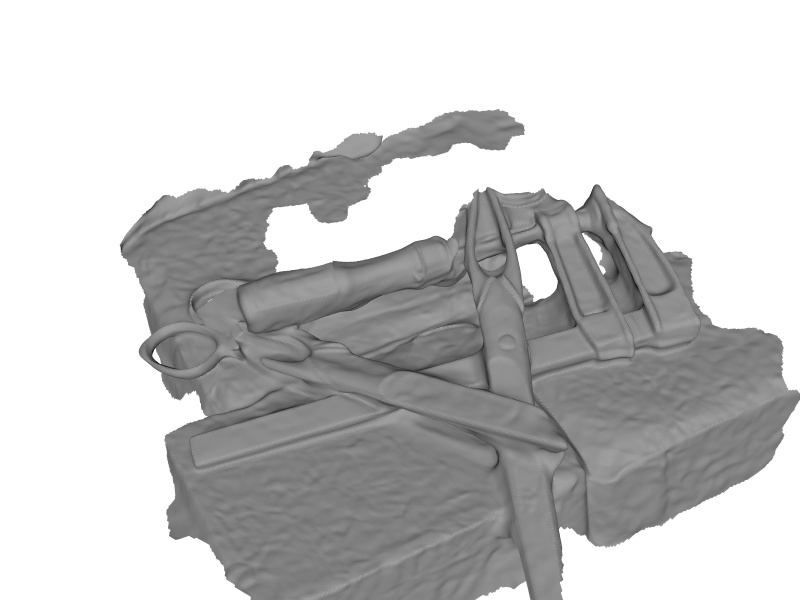} & 
		\includegraphics[width=0.19\linewidth]{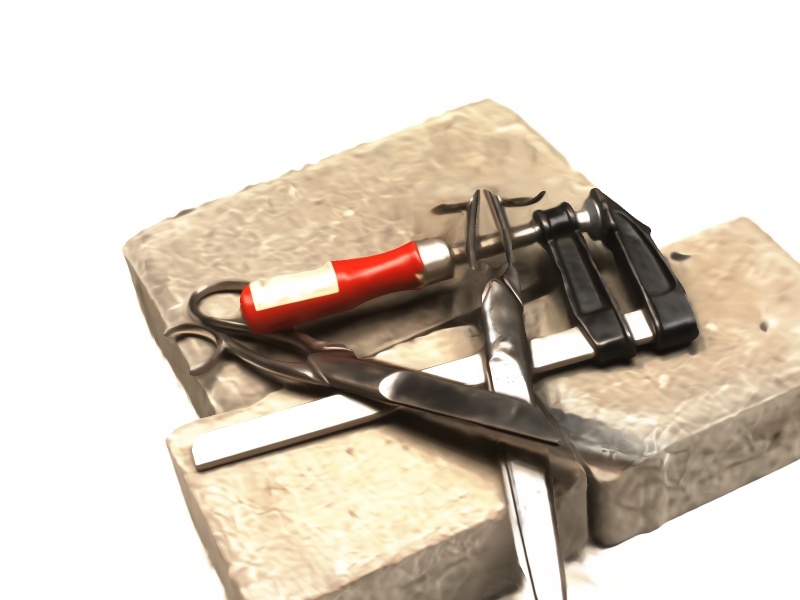} \\
		
		\includegraphics[width=0.19\linewidth]{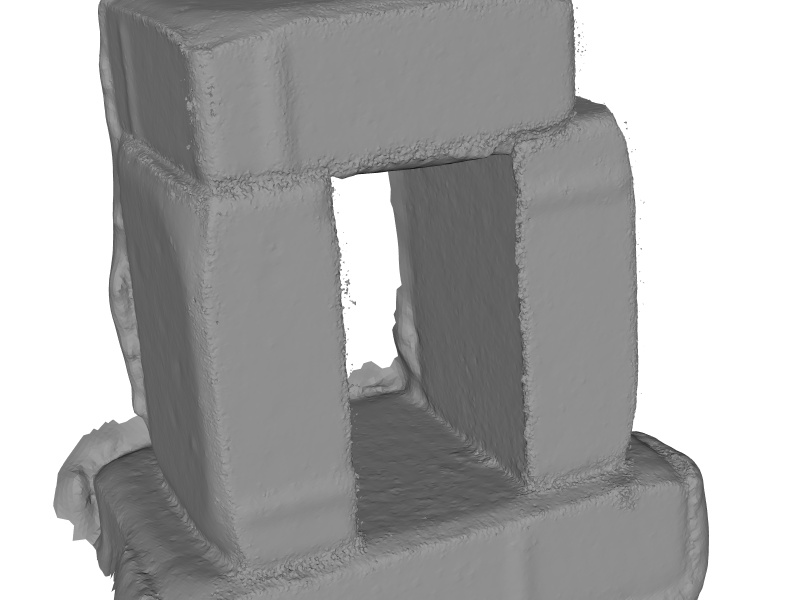} &
		\includegraphics[width=0.19\linewidth]{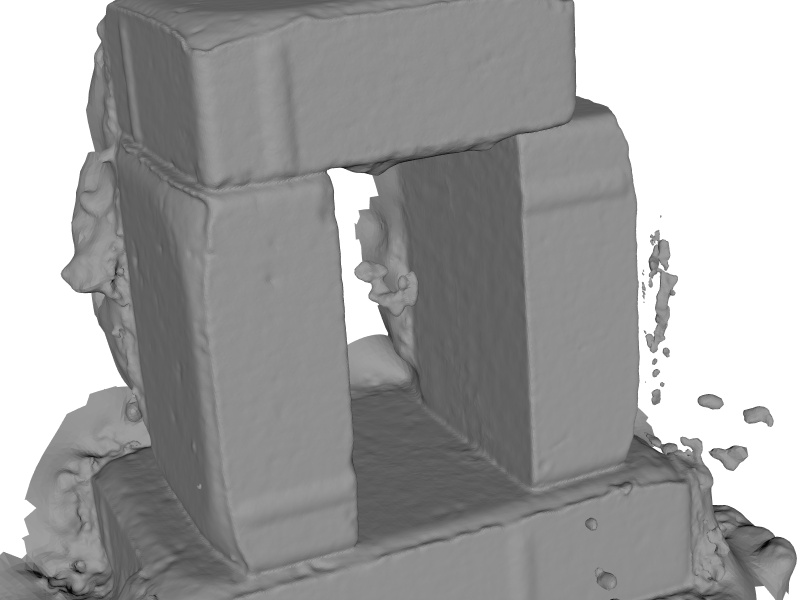} &
		\includegraphics[width=0.19\linewidth]{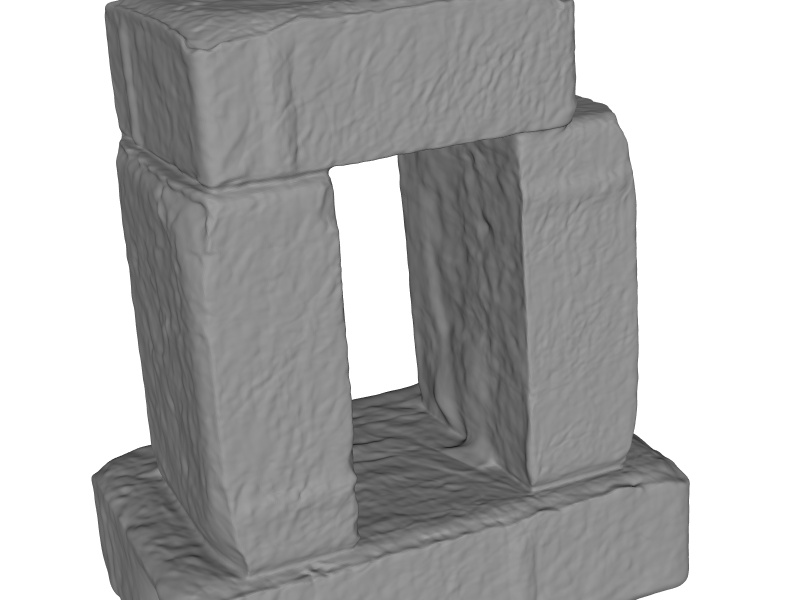} &
		\includegraphics[width=0.19\linewidth]{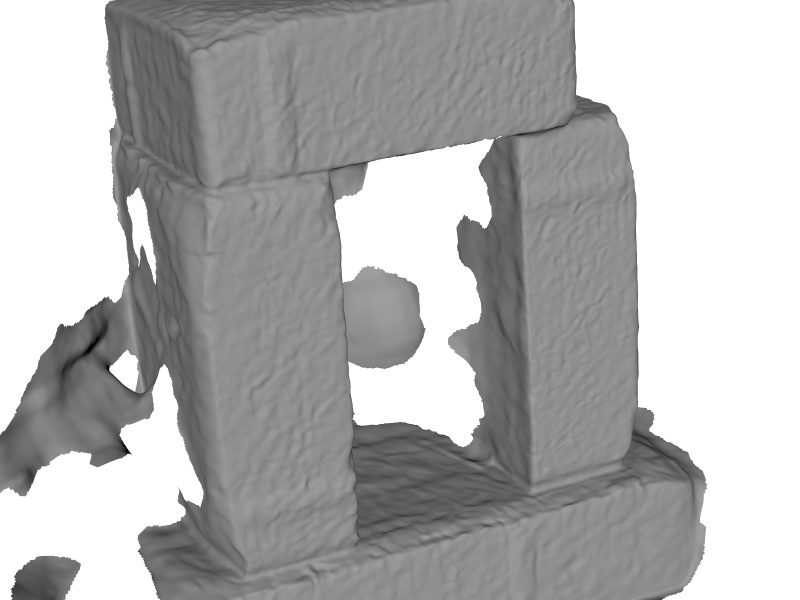} & 
		\includegraphics[width=0.19\linewidth]{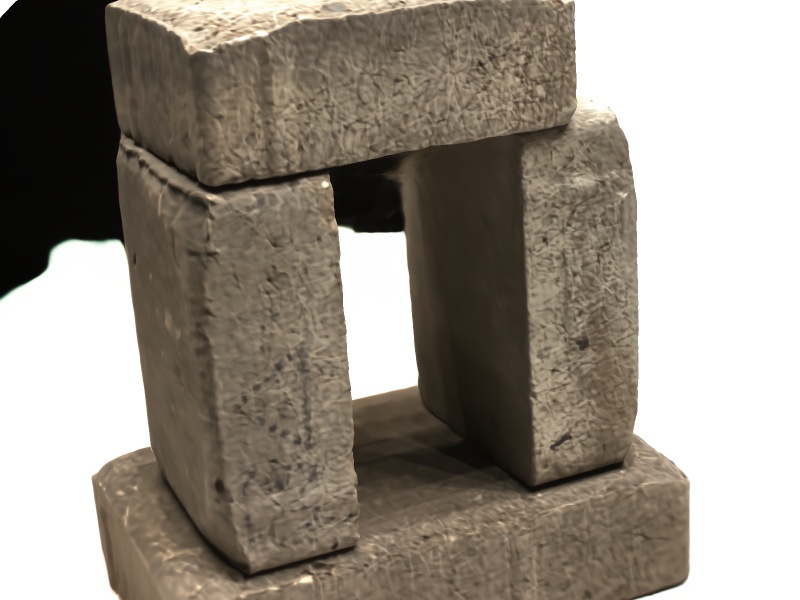} \\
		
		\includegraphics[width=0.19\linewidth]{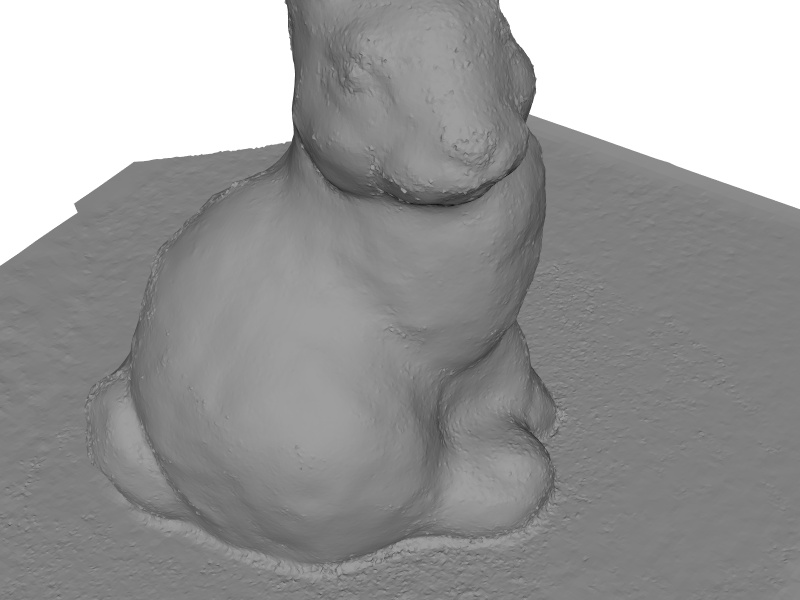} &
		\includegraphics[width=0.19\linewidth]{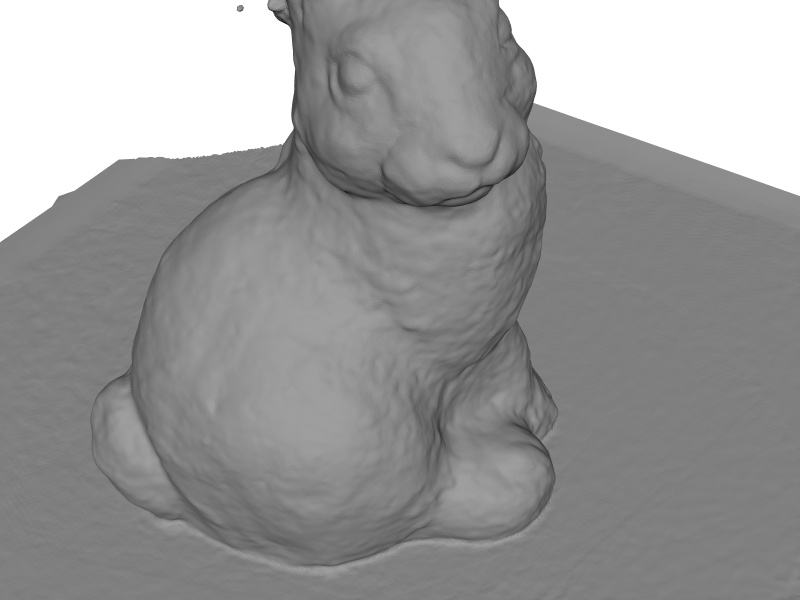} &
		\includegraphics[width=0.19\linewidth]{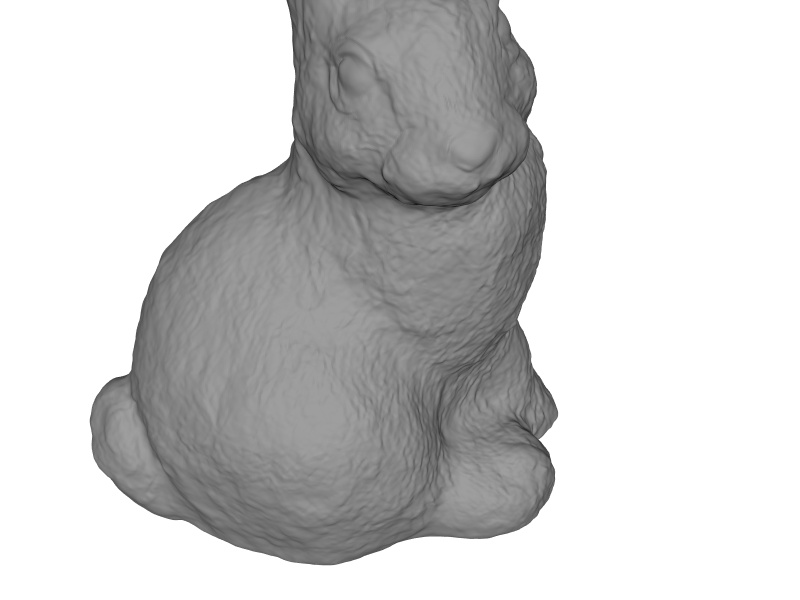} &
		\includegraphics[width=0.19\linewidth]{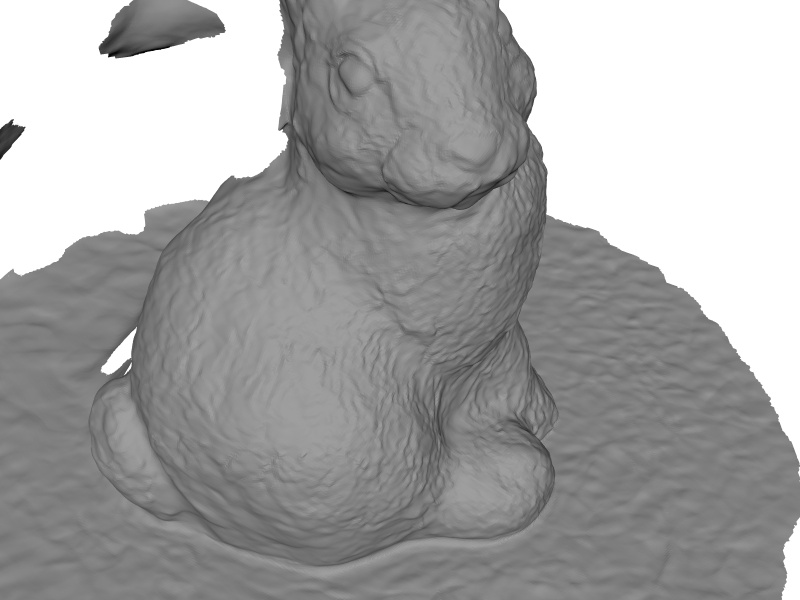} & 
		\includegraphics[width=0.19\linewidth]{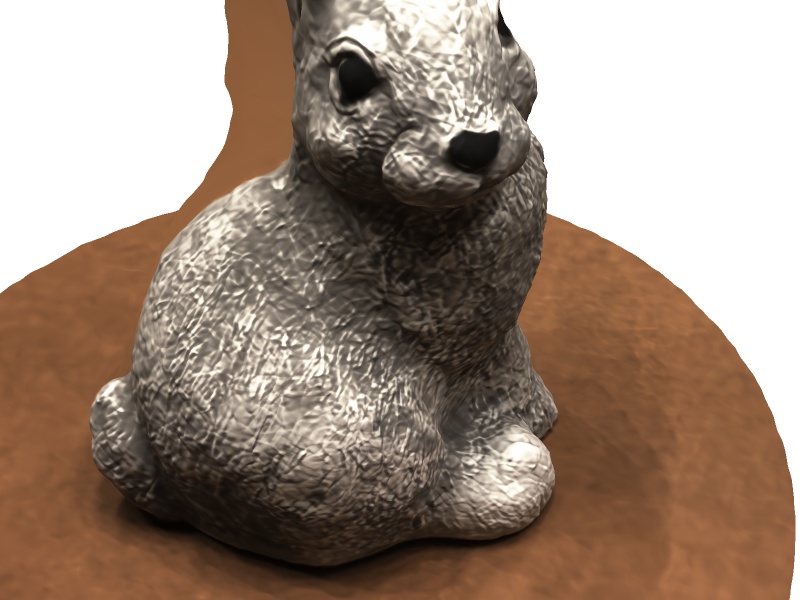} \\
		
		\includegraphics[width=0.19\linewidth]{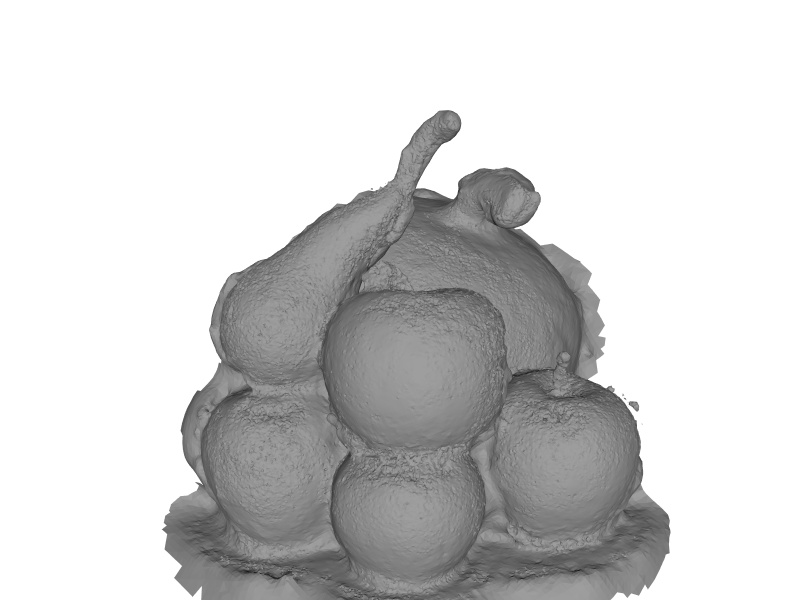} &
		\includegraphics[width=0.19\linewidth]{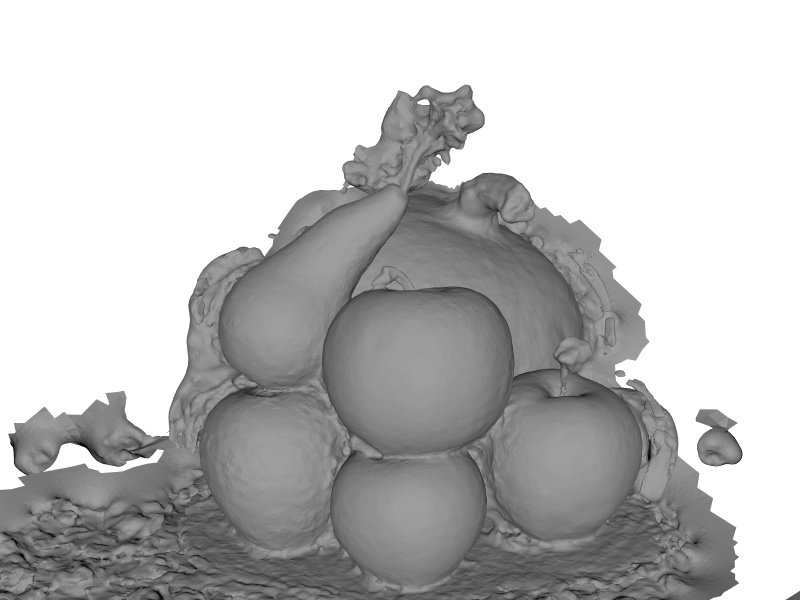} &
		\includegraphics[width=0.19\linewidth]{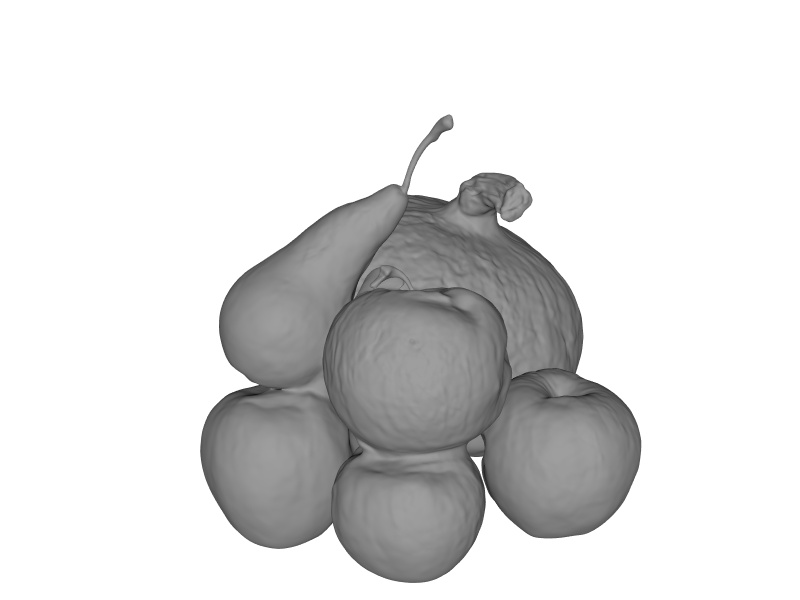} &
		\includegraphics[width=0.19\linewidth]{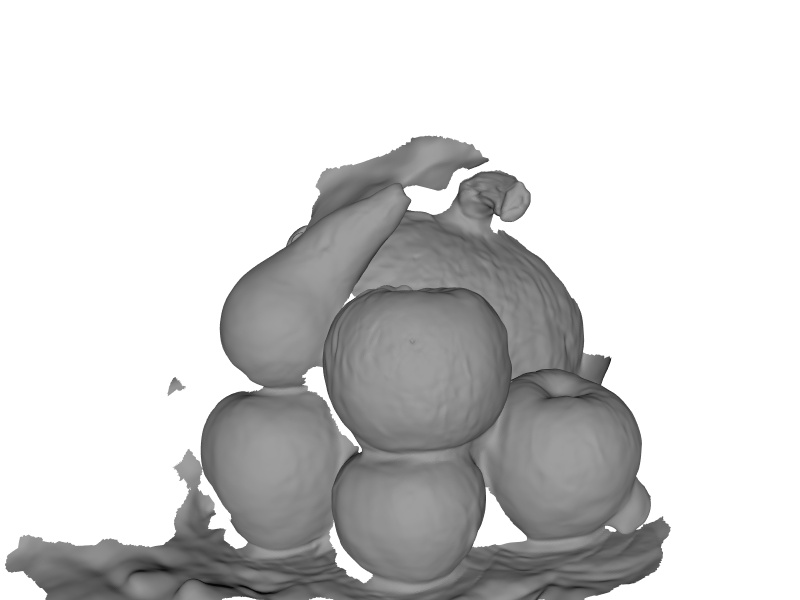} & 
		\includegraphics[width=0.19\linewidth]{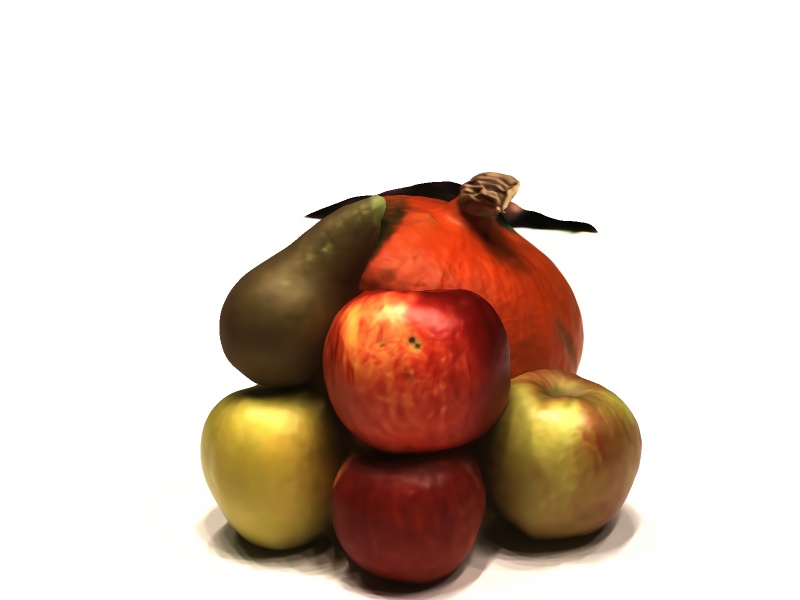} \\
		
		\includegraphics[width=0.19\linewidth]{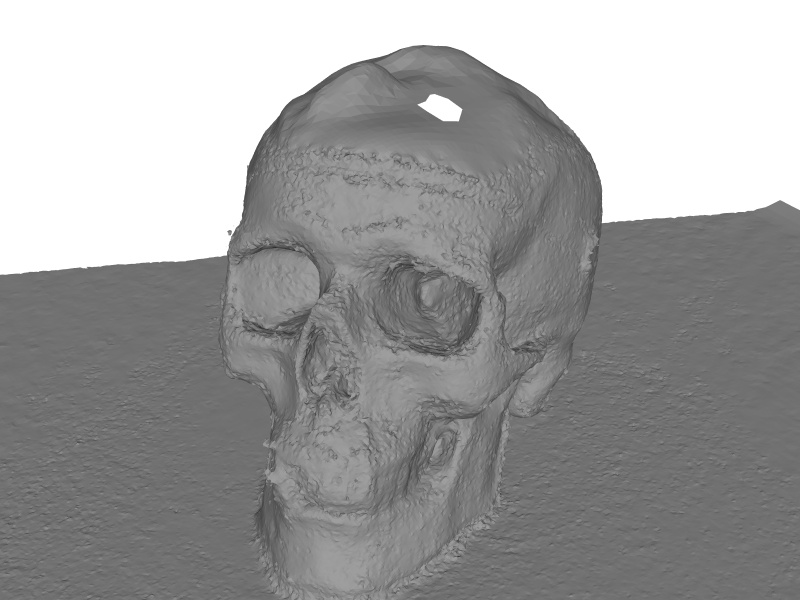} &
		\includegraphics[width=0.19\linewidth]{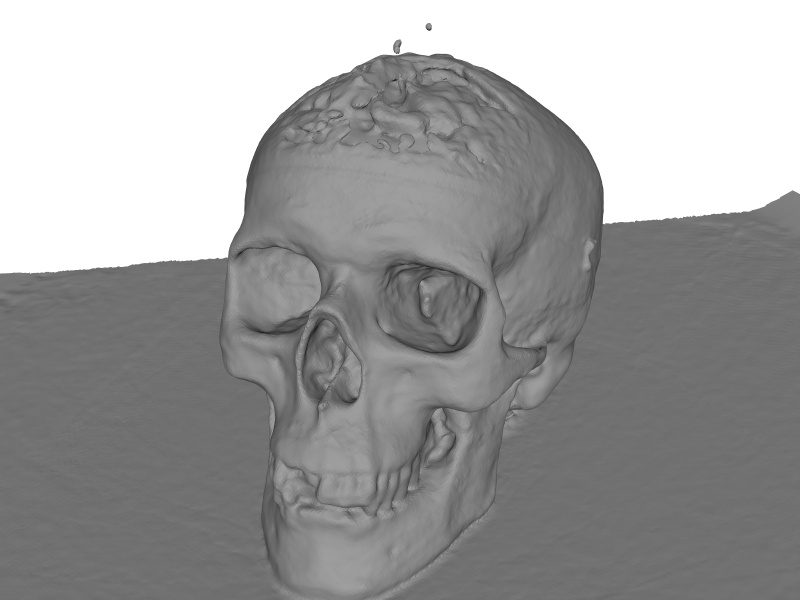} &
		\includegraphics[width=0.19\linewidth]{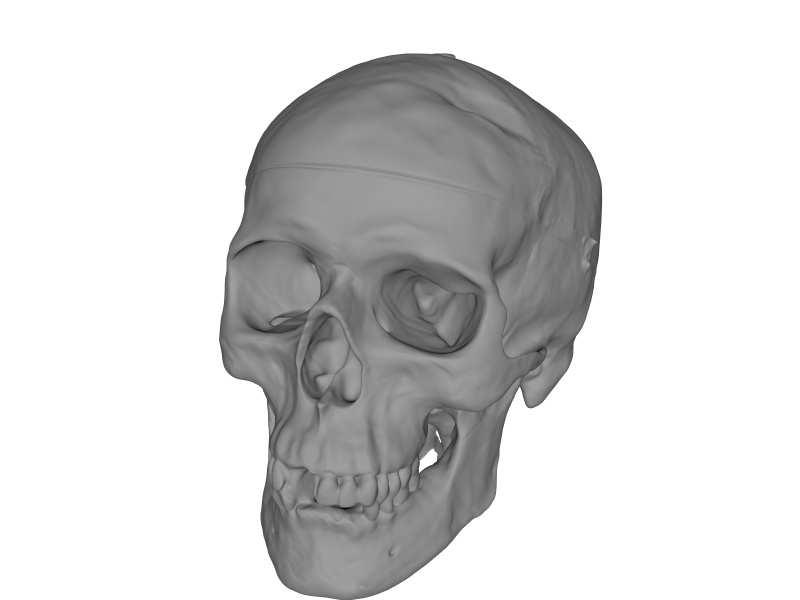} &
		\includegraphics[width=0.19\linewidth]{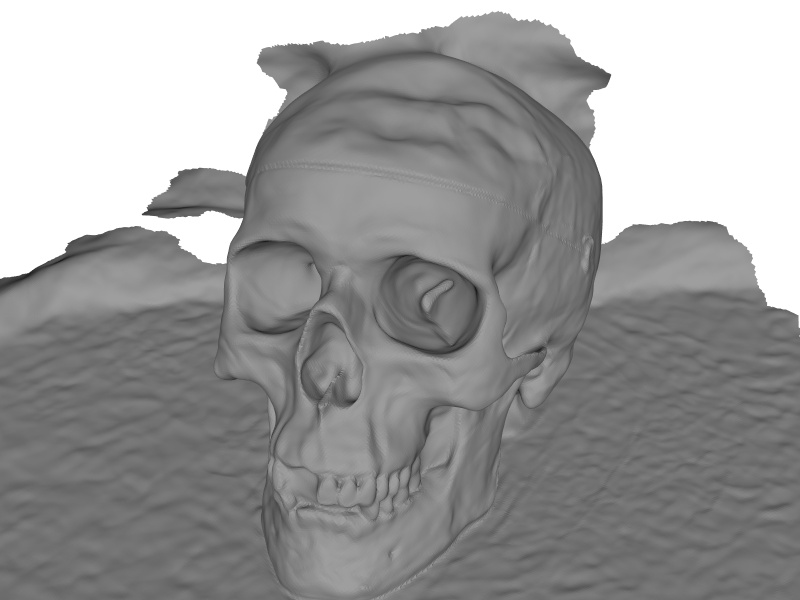} & 
		\includegraphics[width=0.19\linewidth]{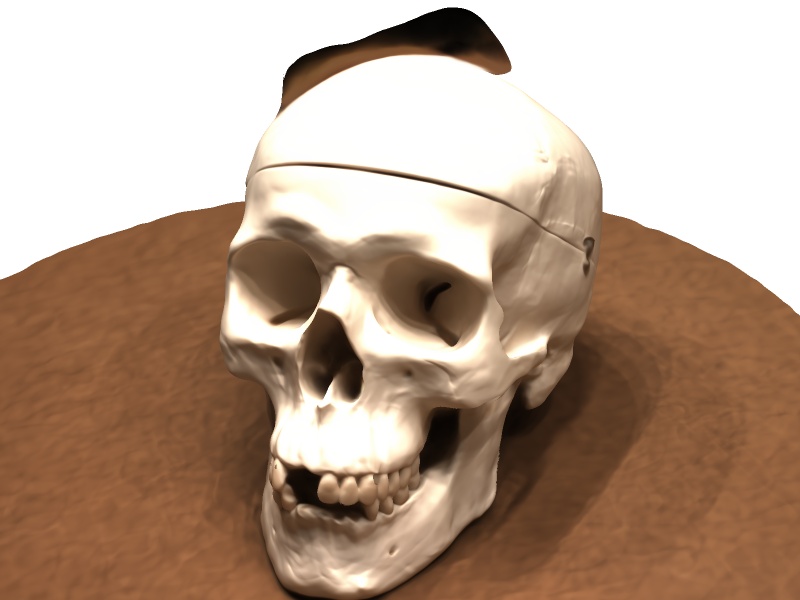} \\
		
		\includegraphics[width=0.19\linewidth]{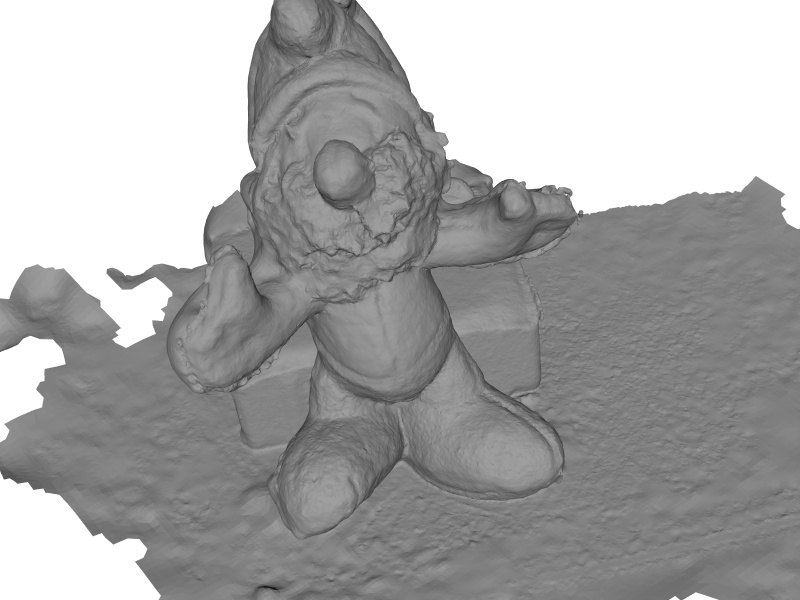} &
		\includegraphics[width=0.19\linewidth]{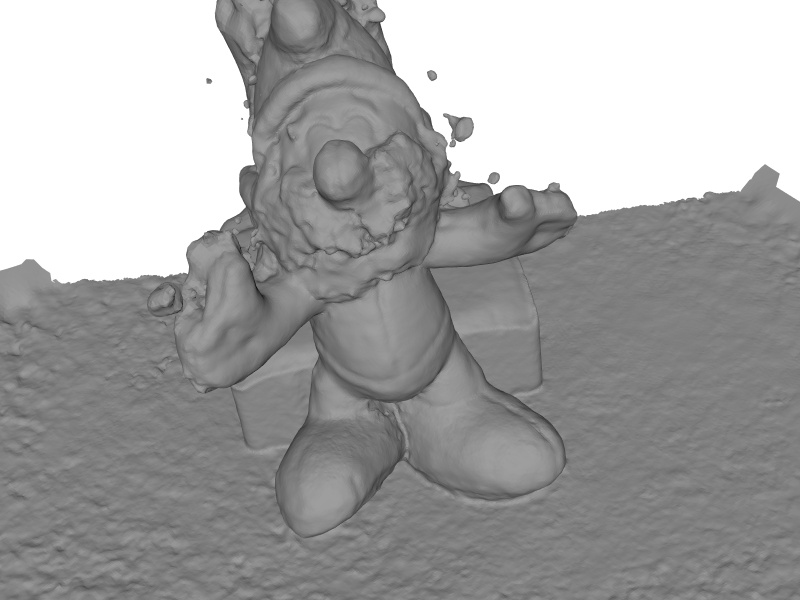} &
		\includegraphics[width=0.19\linewidth]{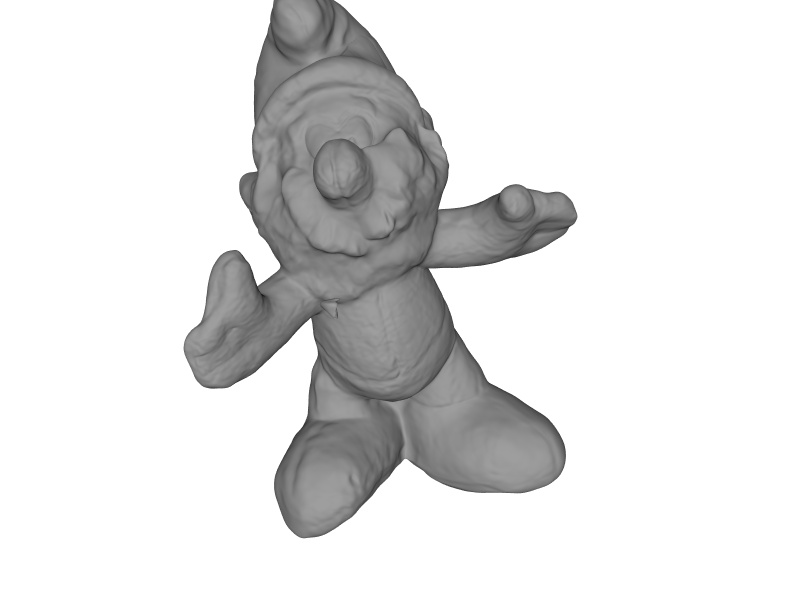} &
		\includegraphics[width=0.19\linewidth]{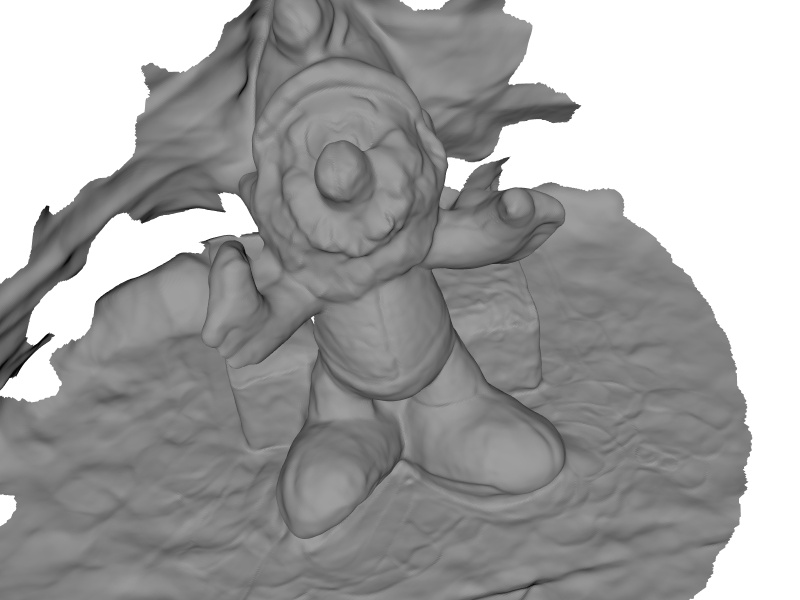} & 
		\includegraphics[width=0.19\linewidth]{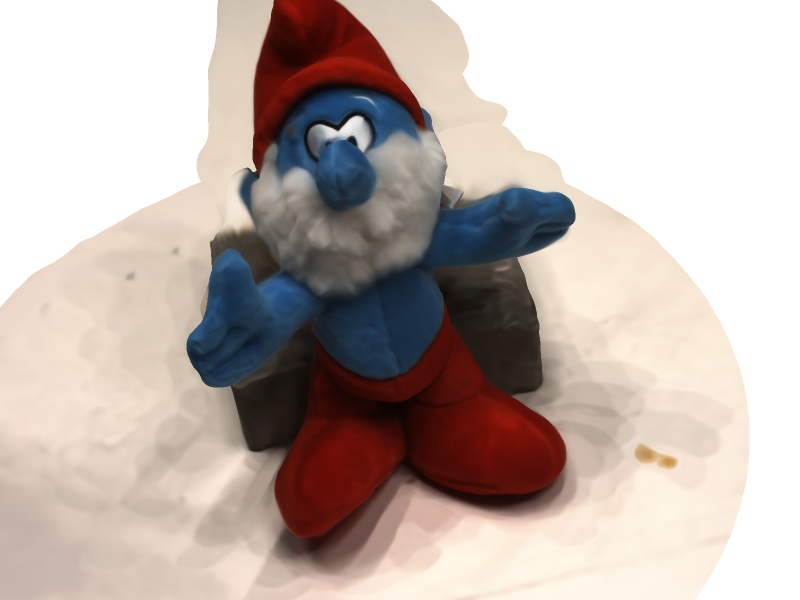} \\
		
		\includegraphics[width=0.19\linewidth]{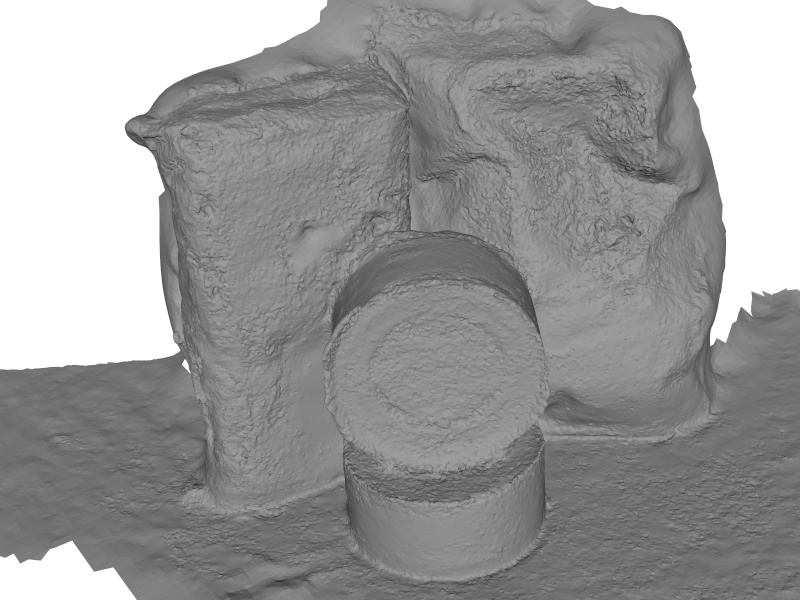} &
		\includegraphics[width=0.19\linewidth]{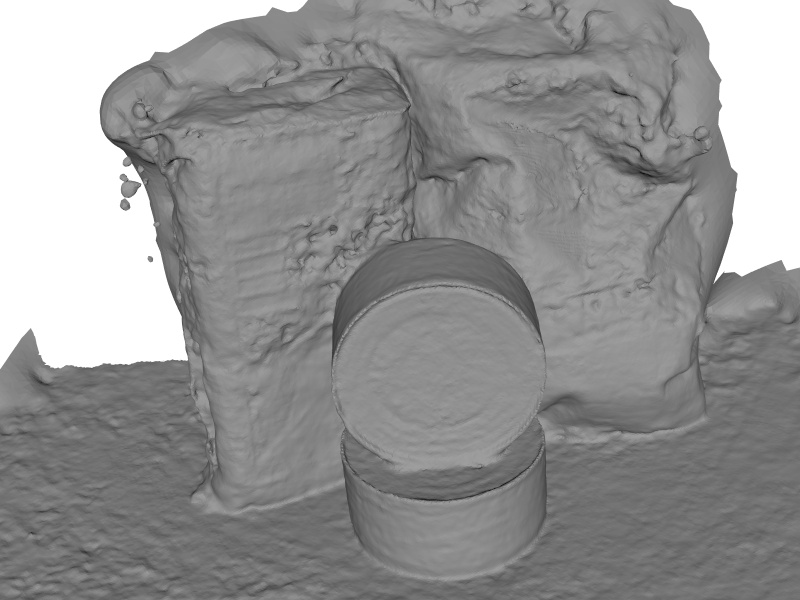} &
		\includegraphics[width=0.19\linewidth]{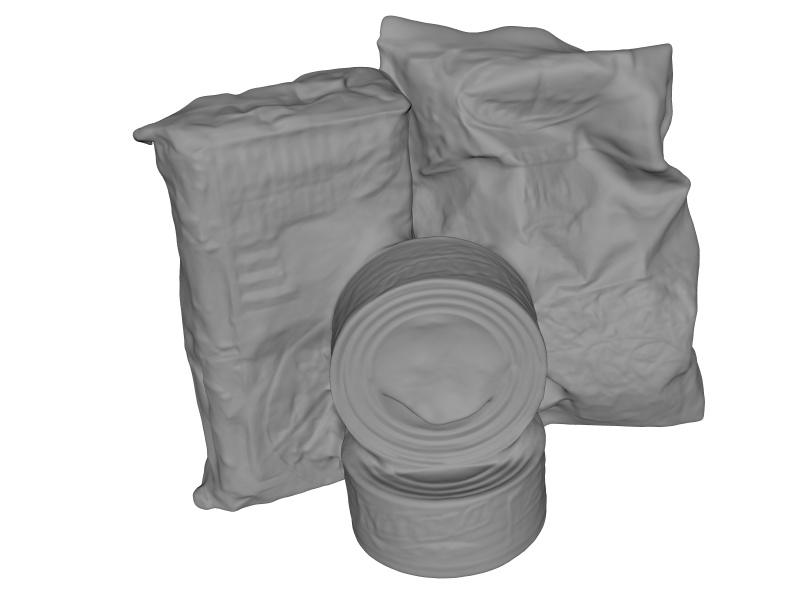} &
		\includegraphics[width=0.19\linewidth]{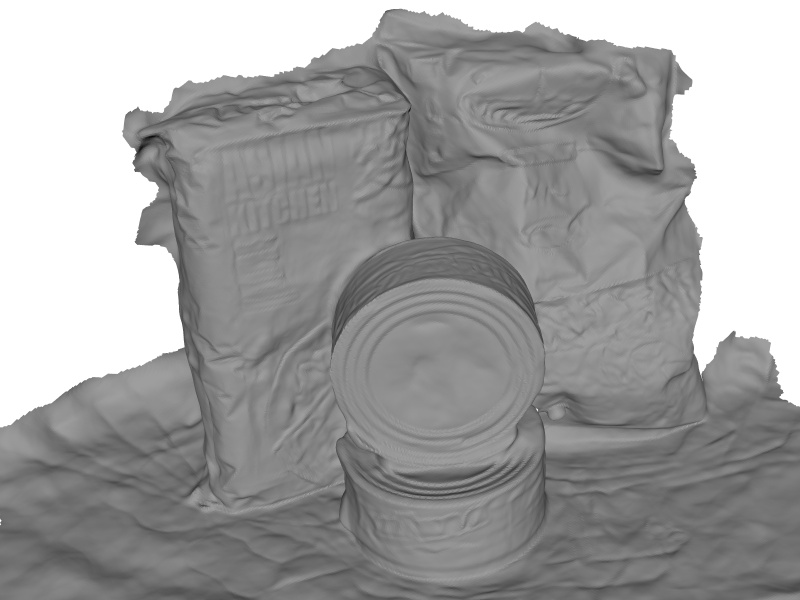} & 
		\includegraphics[width=0.19\linewidth]{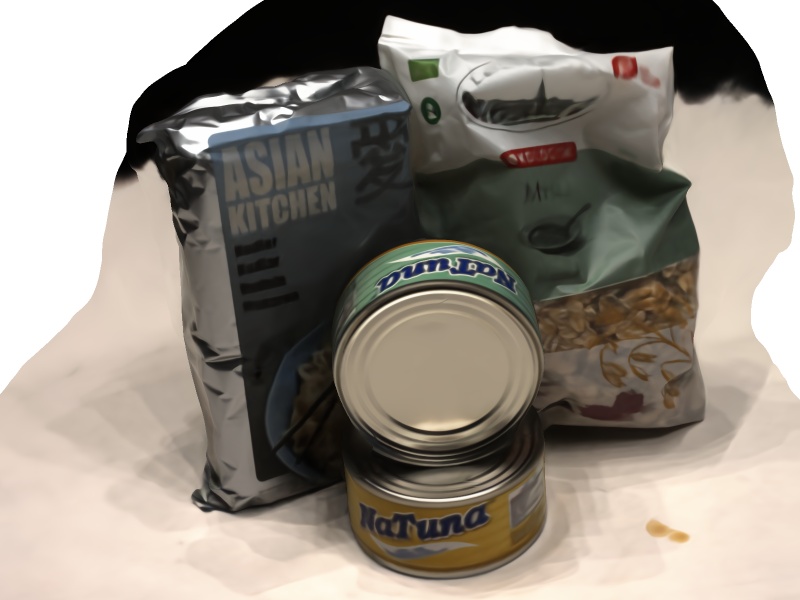} \\
		
		\includegraphics[width=0.19\linewidth]{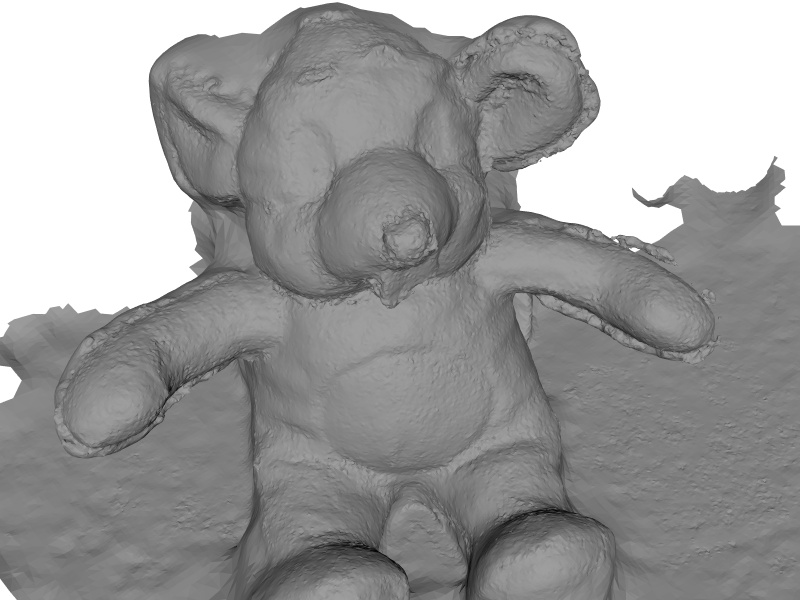} &
		\includegraphics[width=0.19\linewidth]{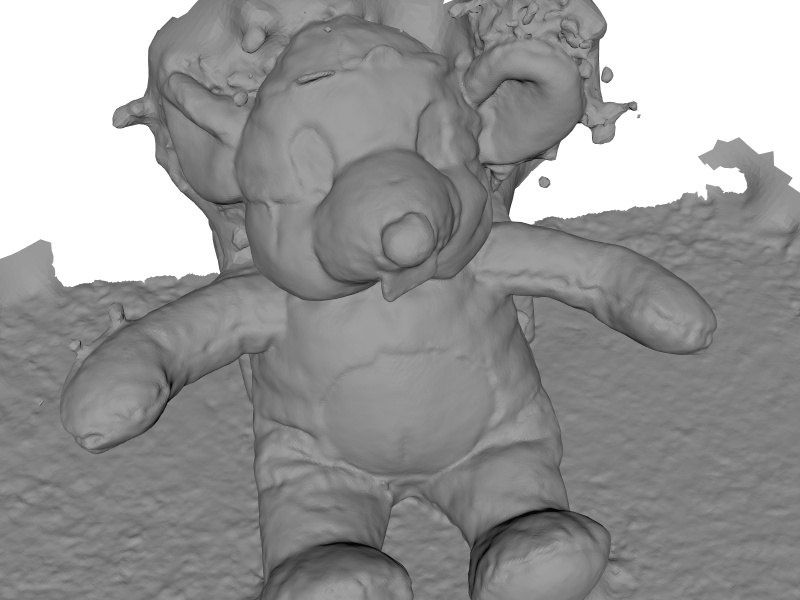} &
		\includegraphics[width=0.19\linewidth]{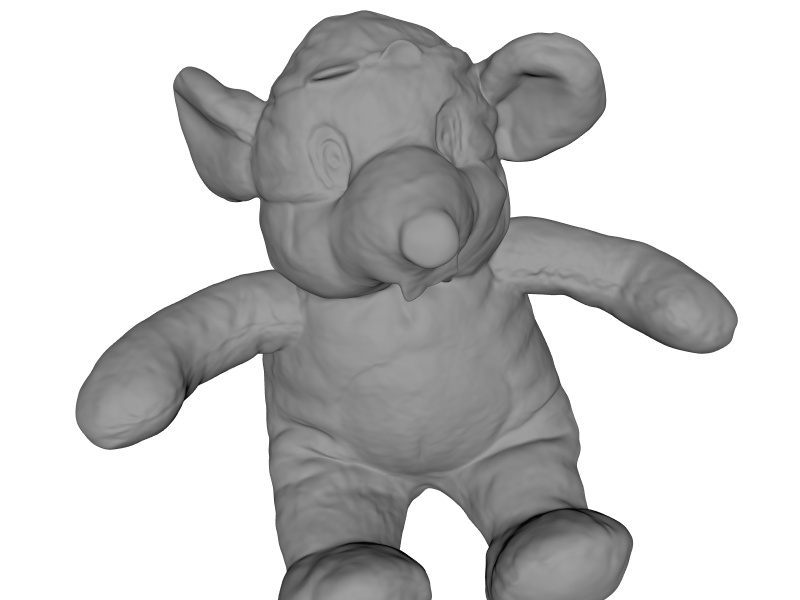} &
		\includegraphics[width=0.19\linewidth]{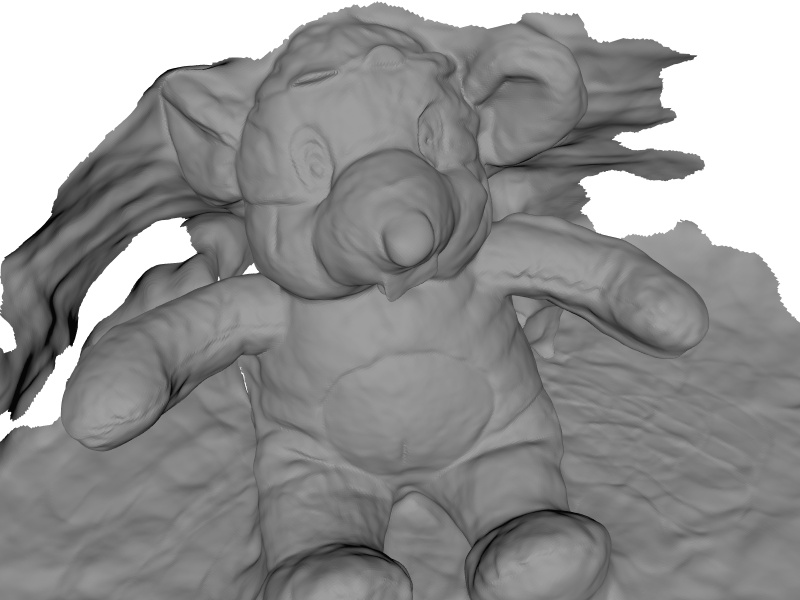} & 
		\includegraphics[width=0.19\linewidth]{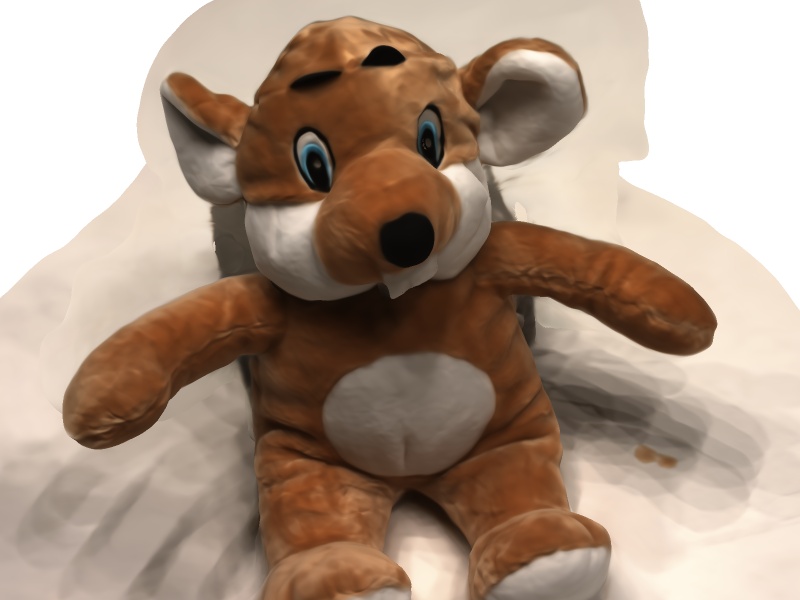} \\
		
		Colmap & Vis-MVSNet & IDR (perfect mask) & MVSDF (Ours) & MVSDF (Ours) Render
	\end{tabular}
	\caption{Qualitative Results on DTU dataset. }
	\label{fig:dtu1_supp}
\end{figure*}

\begin{figure*}
	\centering
	\begin{tabular}{@{\hskip2pt}c@{\hskip2pt}@{\hskip2pt}c@{\hskip2pt}@{\hskip2pt}c@{\hskip2pt}@{\hskip2pt}c@{\hskip2pt}@{\hskip2pt}c@{\hskip2pt}}
		
		\includegraphics[width=0.19\linewidth]{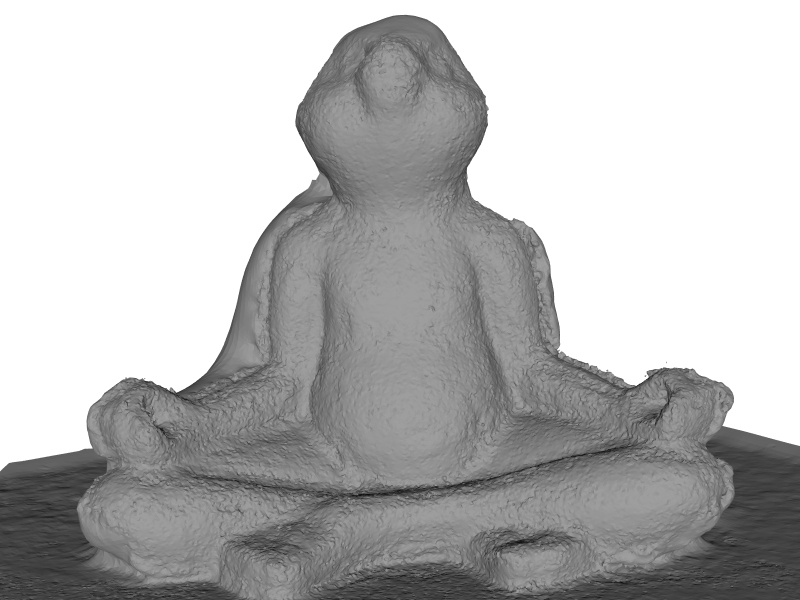} &
		\includegraphics[width=0.19\linewidth]{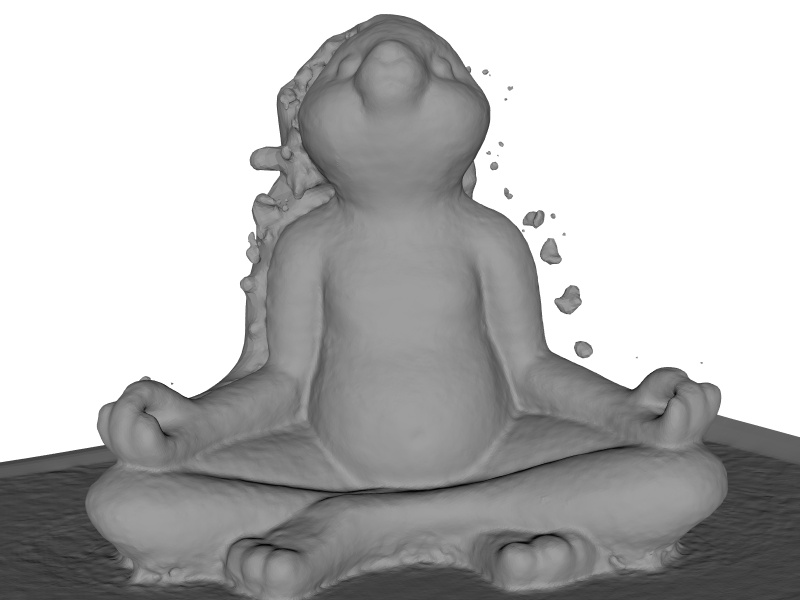} &
		\includegraphics[width=0.19\linewidth]{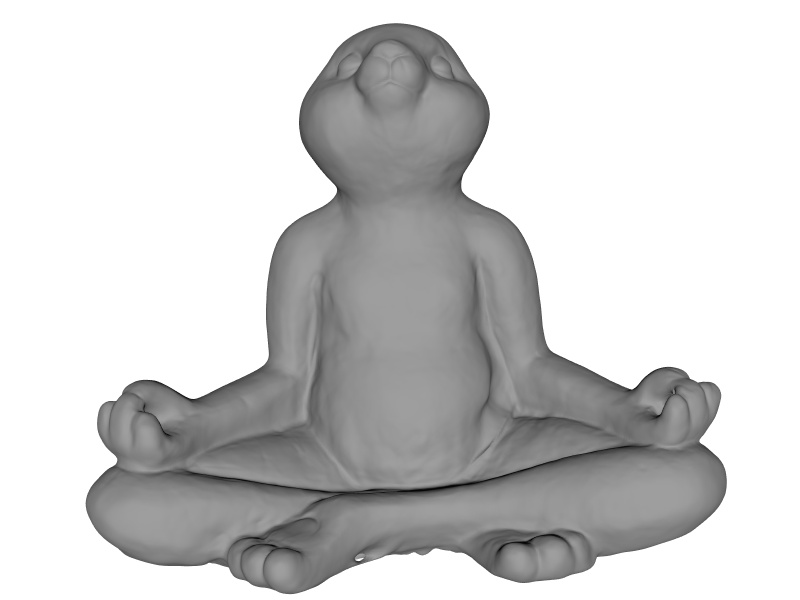} &
		\includegraphics[width=0.19\linewidth]{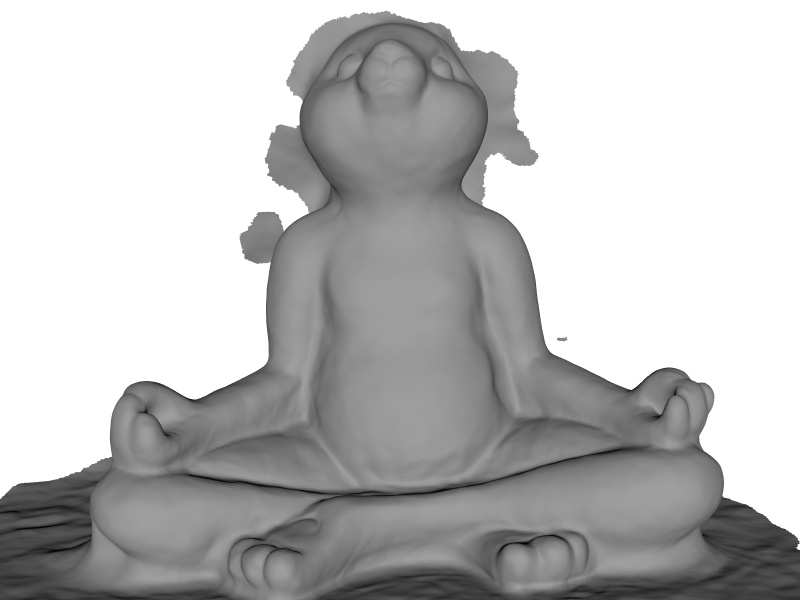} & 
		\includegraphics[width=0.19\linewidth]{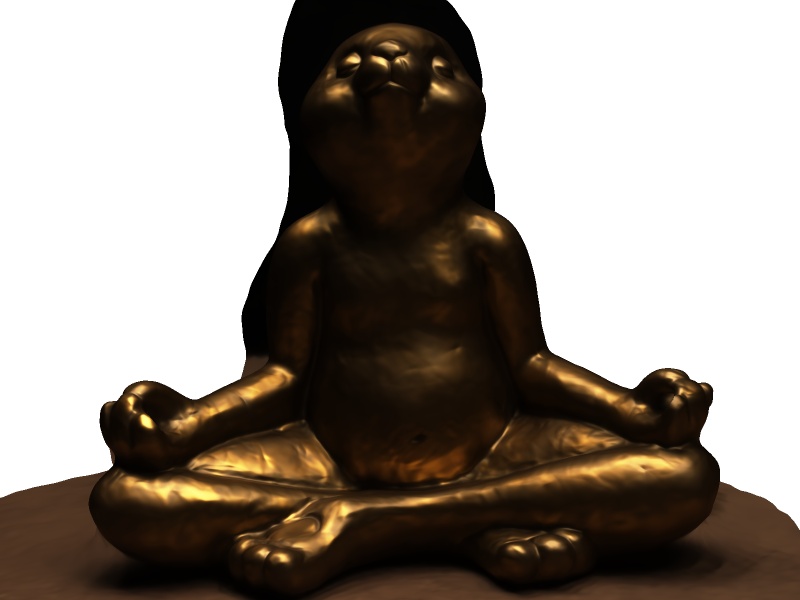} \\
		
		\includegraphics[width=0.19\linewidth]{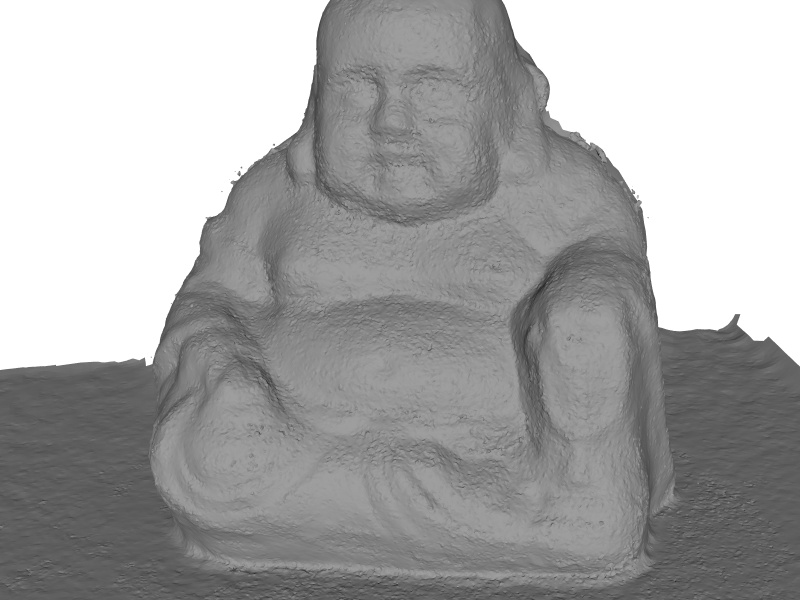} &
		\includegraphics[width=0.19\linewidth]{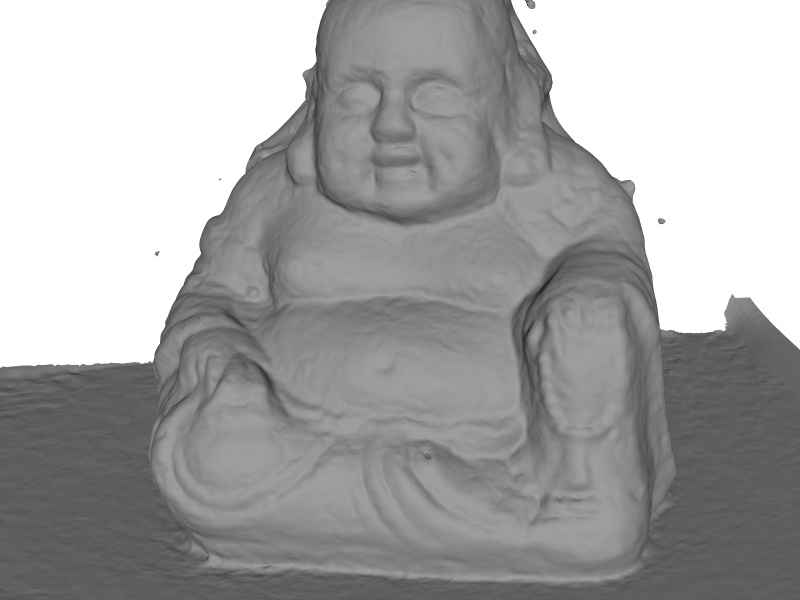} &
		\includegraphics[width=0.19\linewidth]{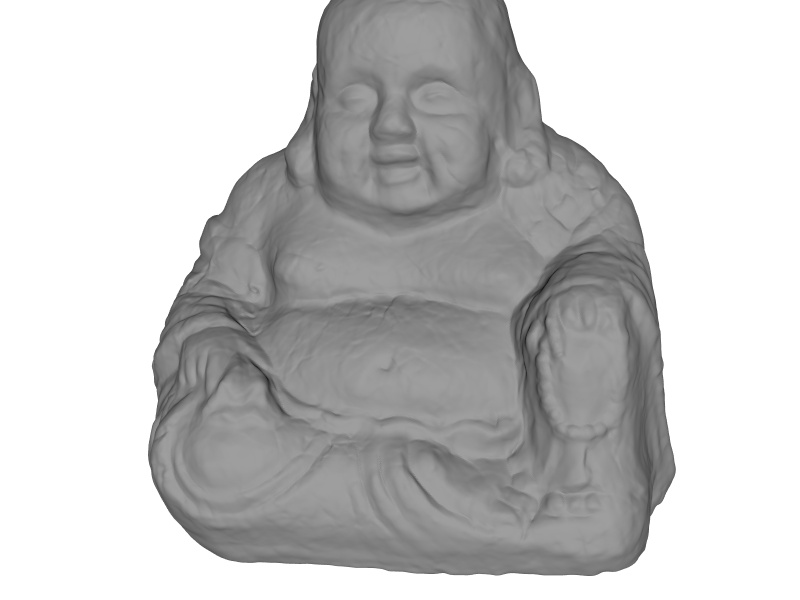} &
		\includegraphics[width=0.19\linewidth]{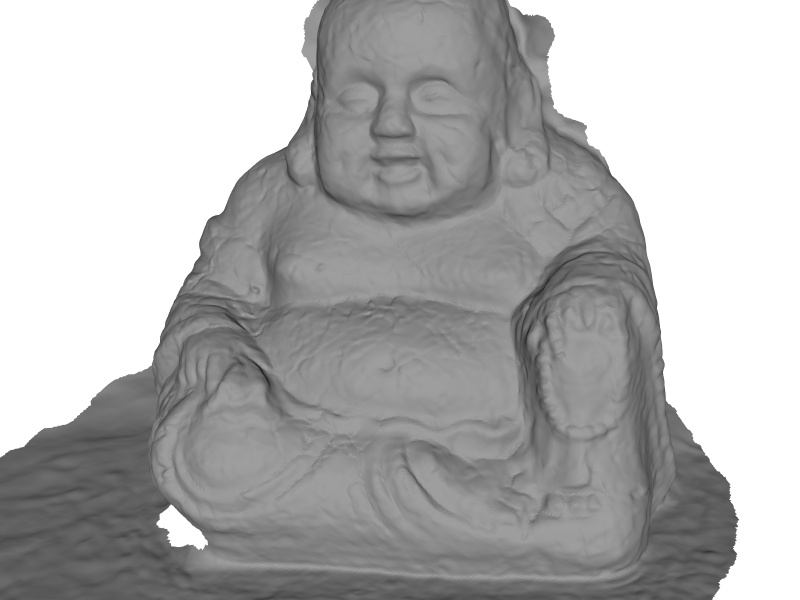} & 
		\includegraphics[width=0.19\linewidth]{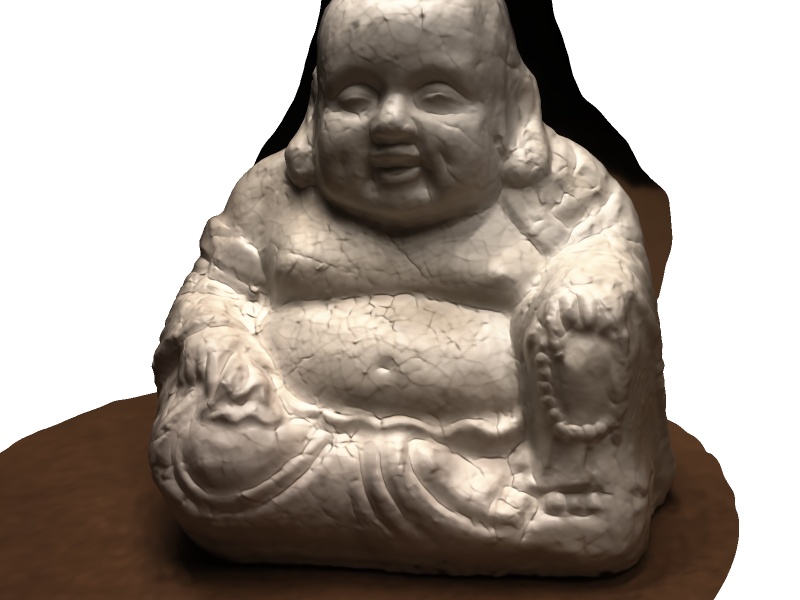} \\
		
		\includegraphics[width=0.19\linewidth]{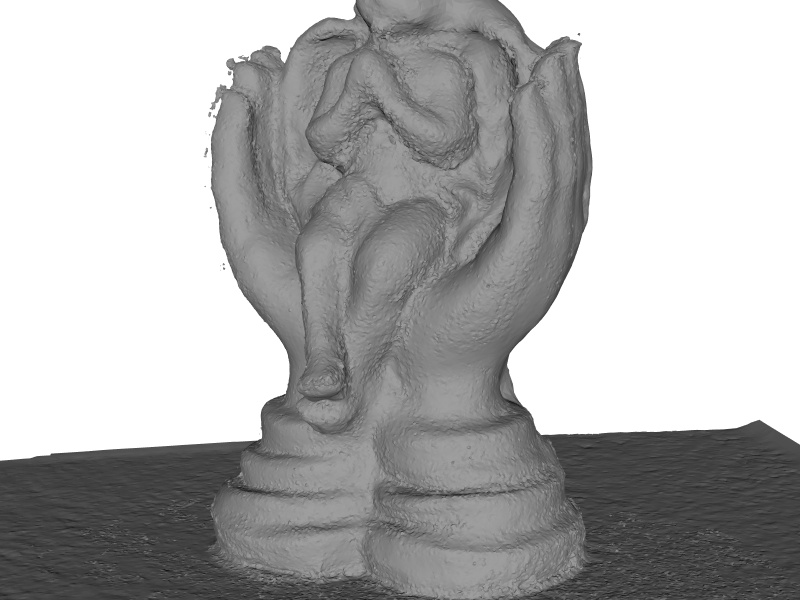} &
		\includegraphics[width=0.19\linewidth]{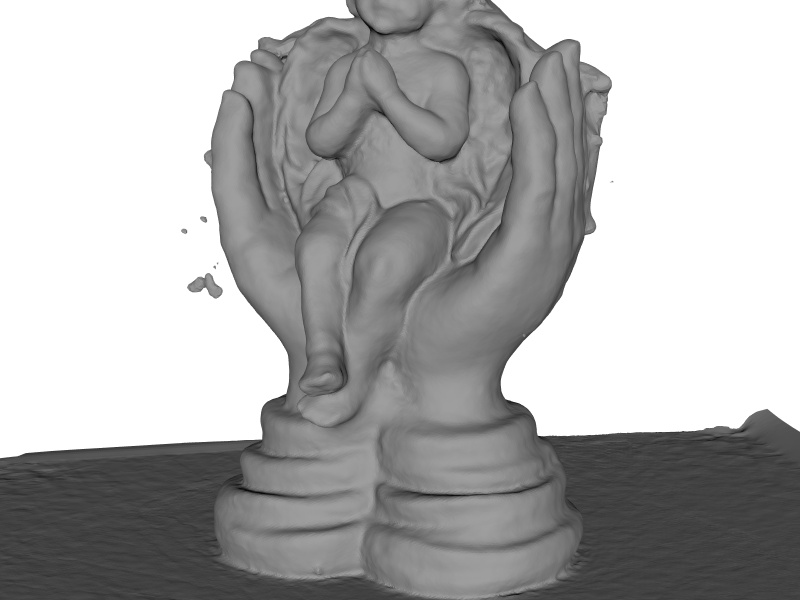} &
		\includegraphics[width=0.19\linewidth]{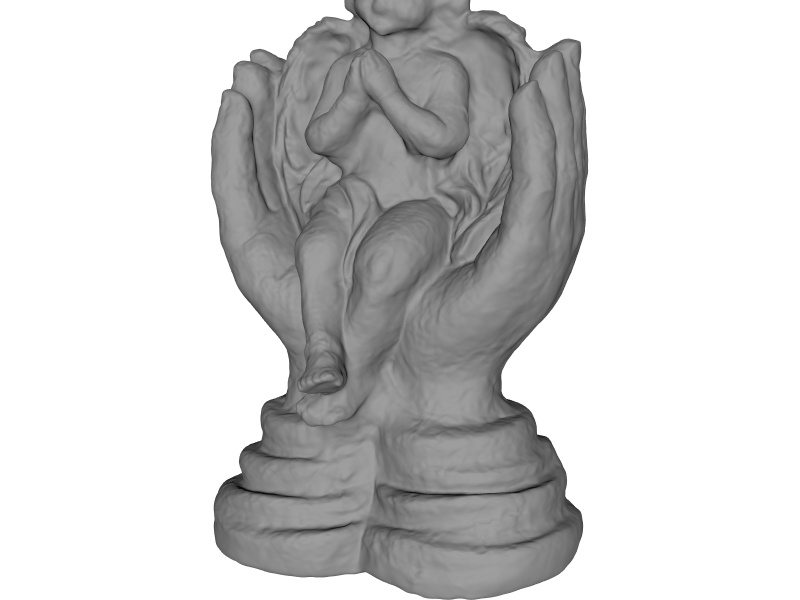} &
		\includegraphics[width=0.19\linewidth]{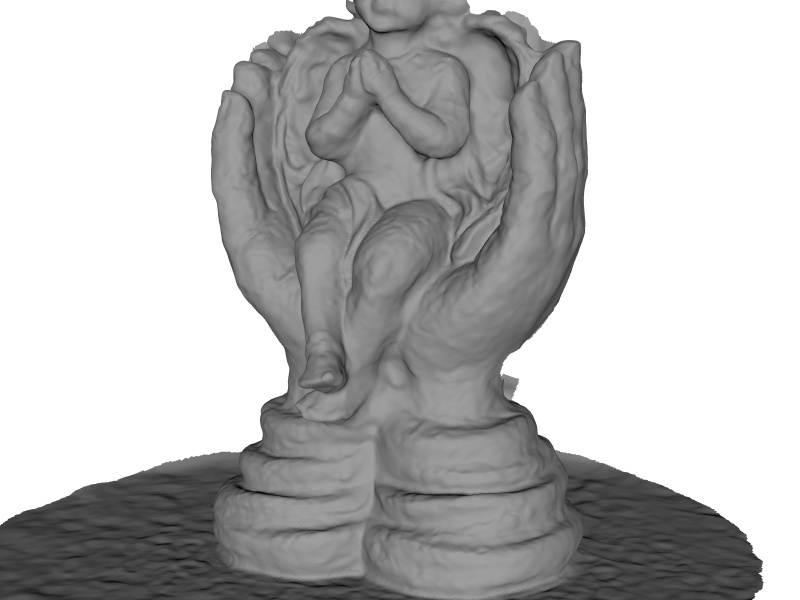} & 
		\includegraphics[width=0.19\linewidth]{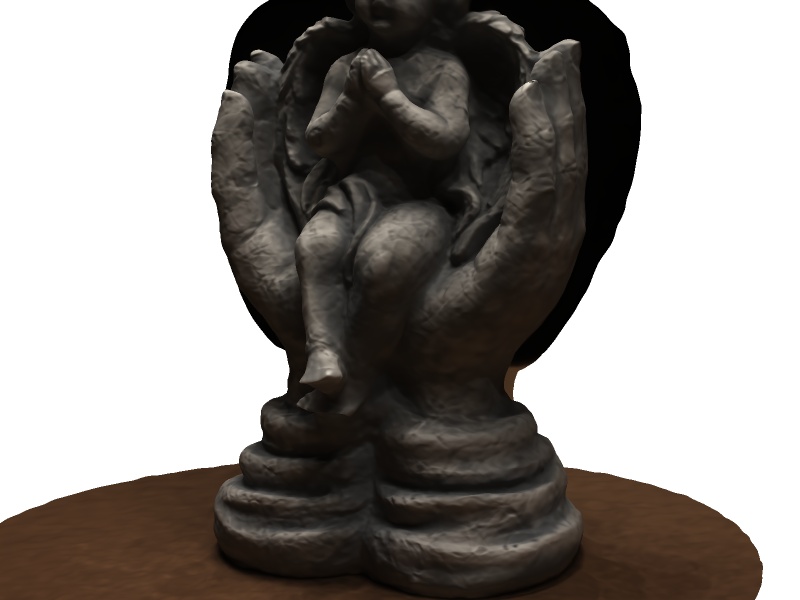} \\
		
		\includegraphics[width=0.19\linewidth]{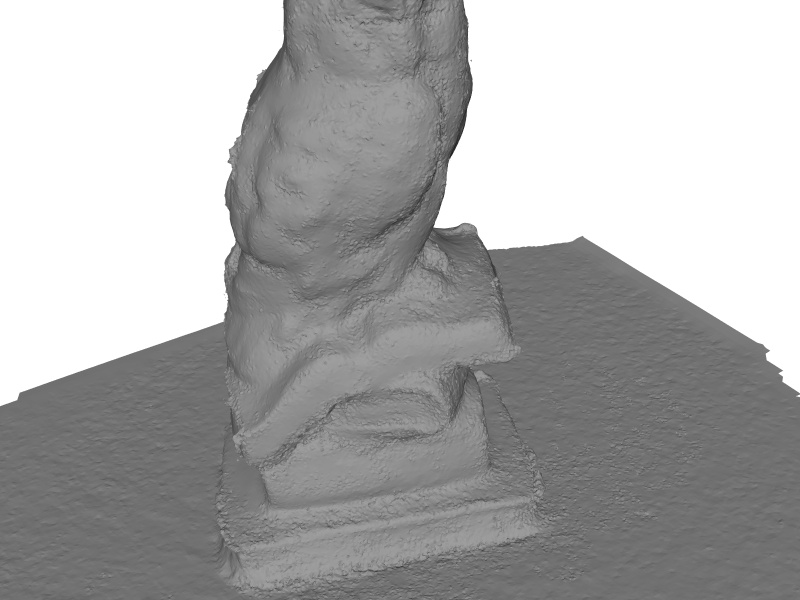} &
		\includegraphics[width=0.19\linewidth]{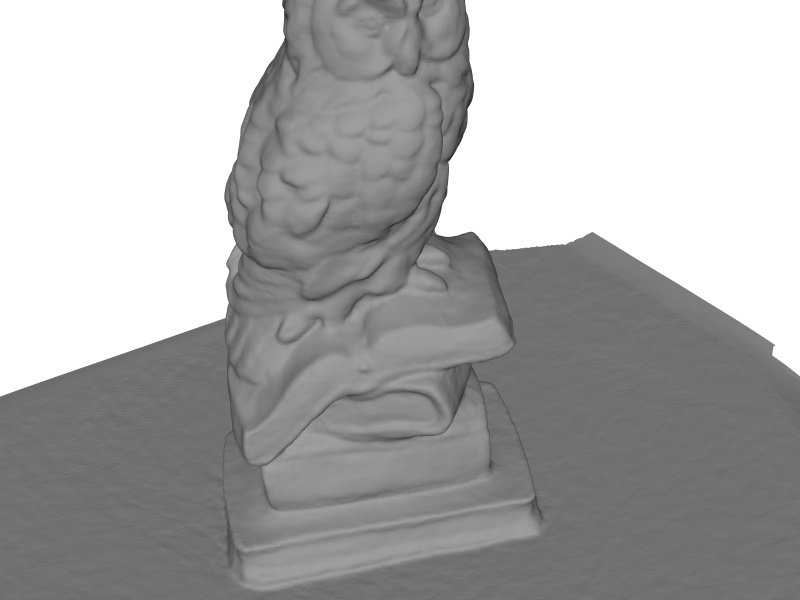} &
		\includegraphics[width=0.19\linewidth]{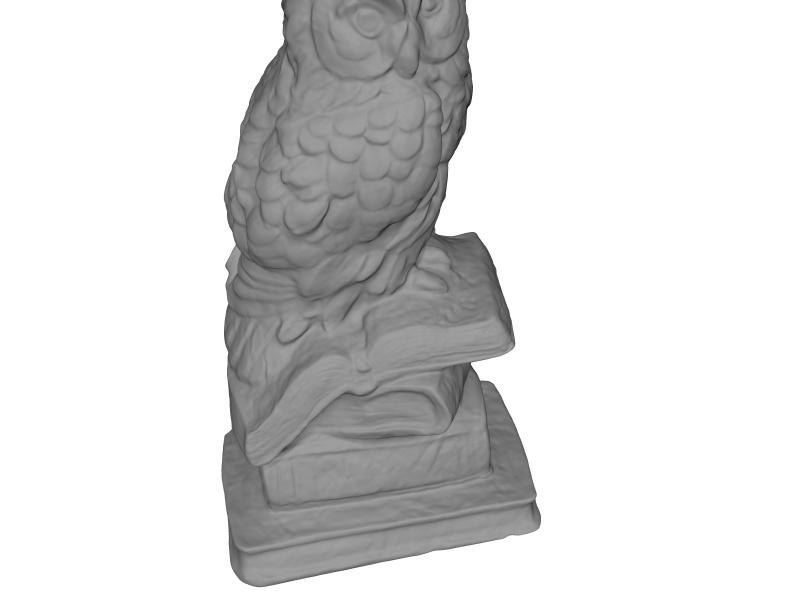} &
		\includegraphics[width=0.19\linewidth]{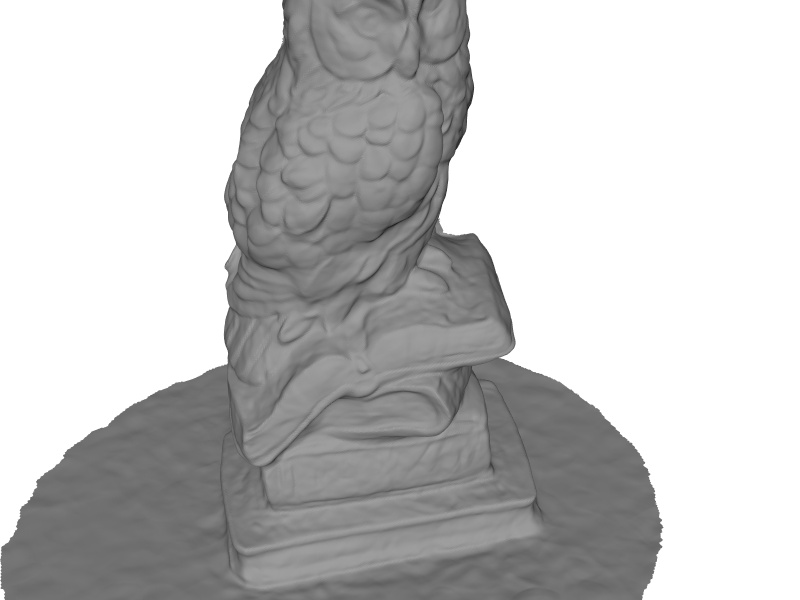} & 
		\includegraphics[width=0.19\linewidth]{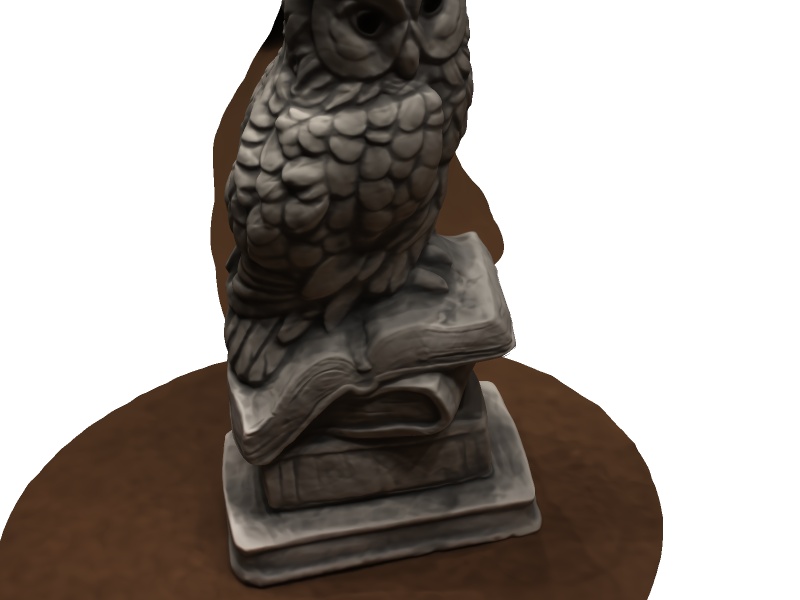} \\
		
		Colmap & Vis-MVSNet & IDR (perfect mask) & MVSDF (Ours) & MVSDF (Ours) Render
	\end{tabular}
	\caption{Qualitative Results on DTU dataset. }
	\label{fig:dtu2_supp}
\end{figure*}

\begin{figure*}
	\centering
	\begin{tabular}{@{\hskip2pt}c@{\hskip2pt}@{\hskip2pt}c@{\hskip2pt}@{\hskip2pt}c@{\hskip2pt}@{\hskip2pt}c@{\hskip2pt}}
		\includegraphics[width=0.24\linewidth]{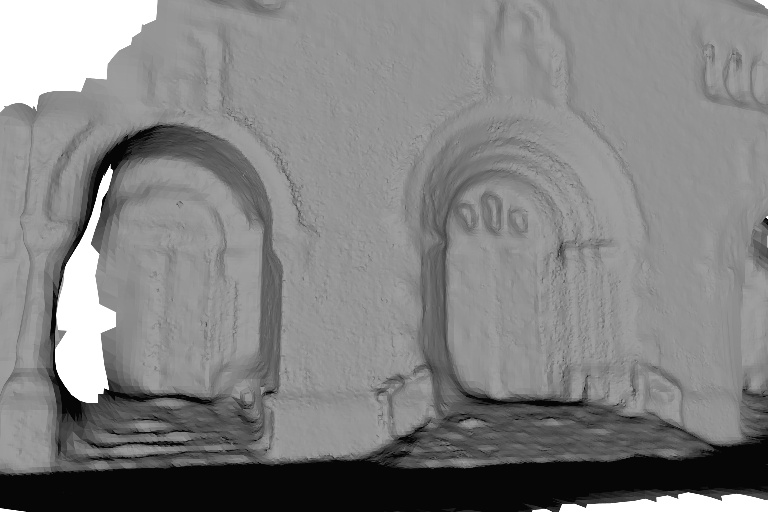} &
		\includegraphics[width=0.24\linewidth]{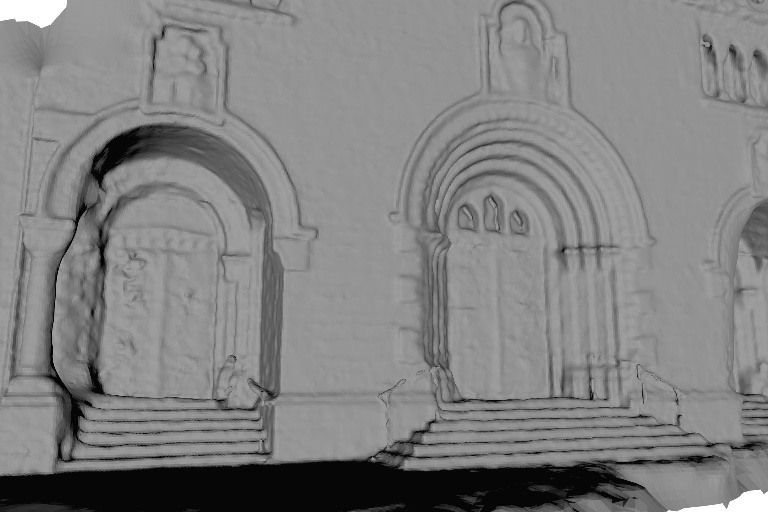} &
		\includegraphics[width=0.24\linewidth]{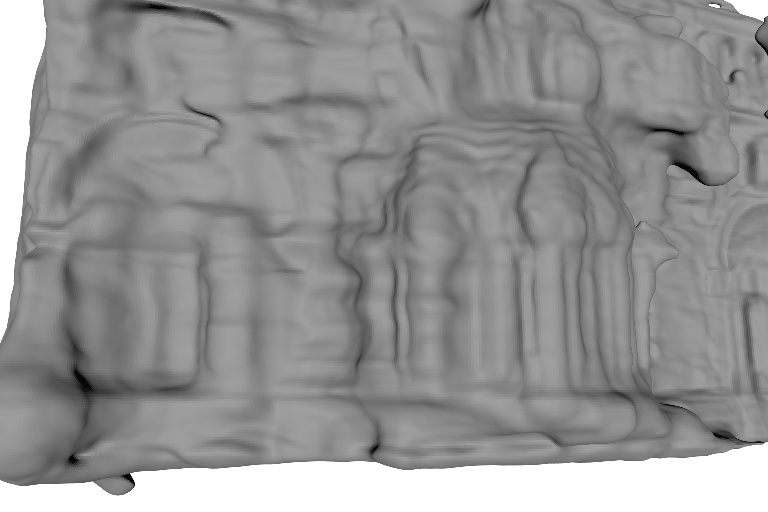} &
		\includegraphics[width=0.24\linewidth]{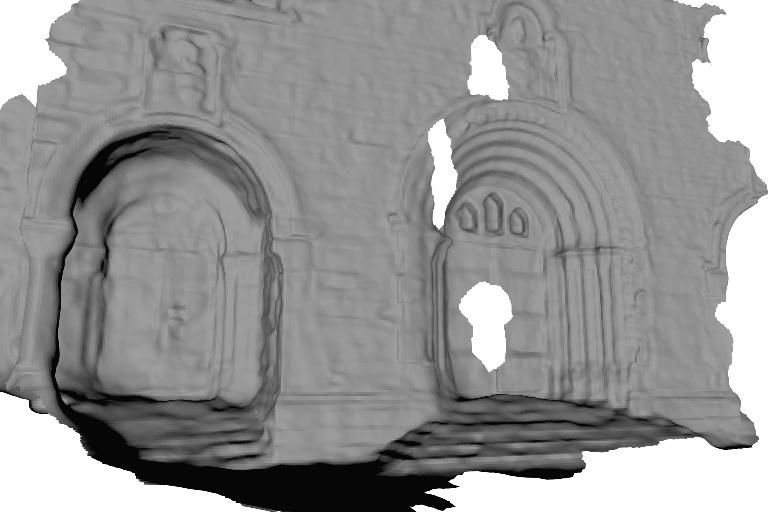} \\
		
		\includegraphics[width=0.24\linewidth]{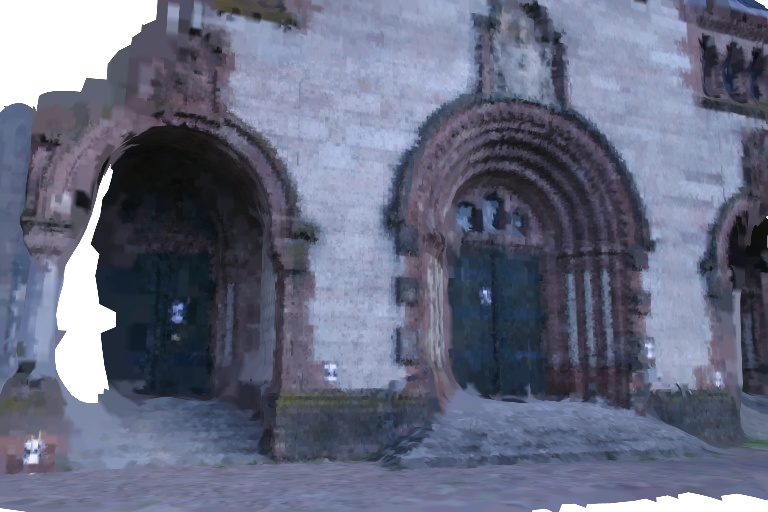} &
		\includegraphics[width=0.24\linewidth]{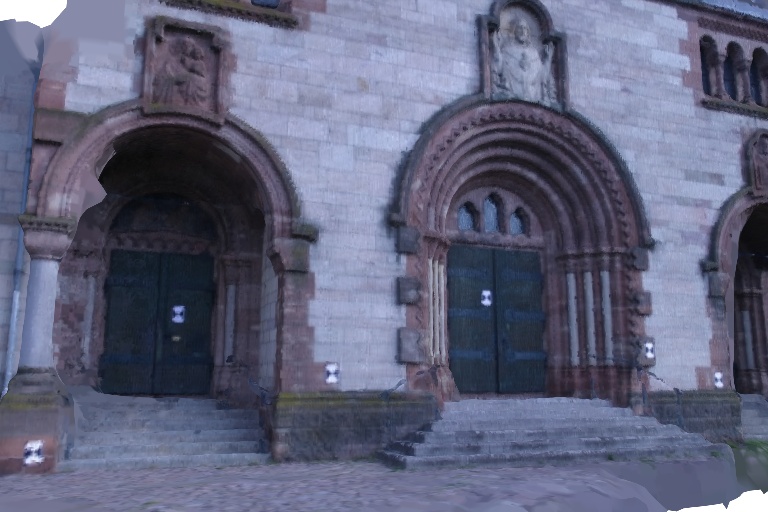} &
		\includegraphics[width=0.24\linewidth]{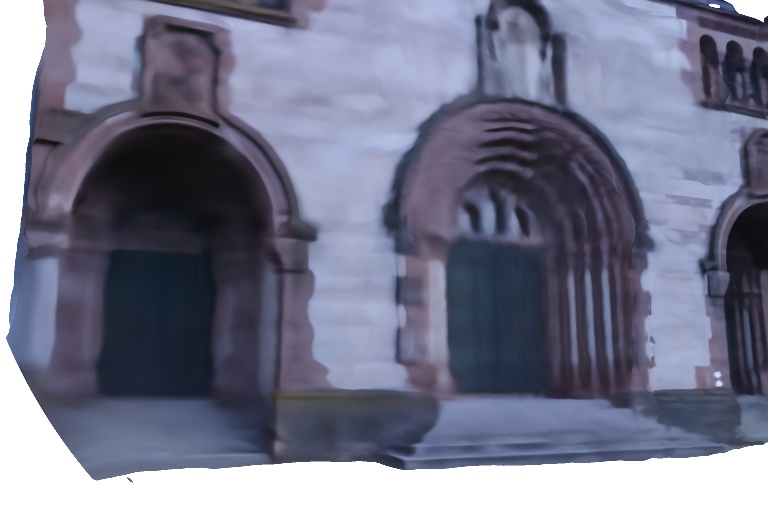} &
		\includegraphics[width=0.24\linewidth]{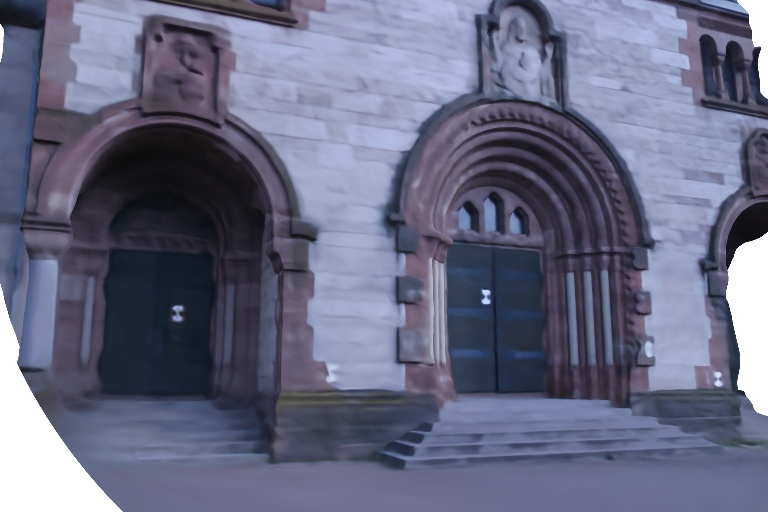} \\
		
		Colmap & Vis-MVSNet & IDR & MVSDF (Ours)
	\end{tabular}
	\caption{Qualitative results on EPFL dataset.}
	\label{fig:epfl_supp}
\end{figure*}

\begin{figure*}
	\centering
	\begin{tabular}{@{\hskip2pt}c@{\hskip2pt}@{\hskip2pt}c@{\hskip2pt}@{\hskip2pt}c@{\hskip2pt}}
		
		\includegraphics[width=0.3\linewidth,trim={150 0 300 0},clip]{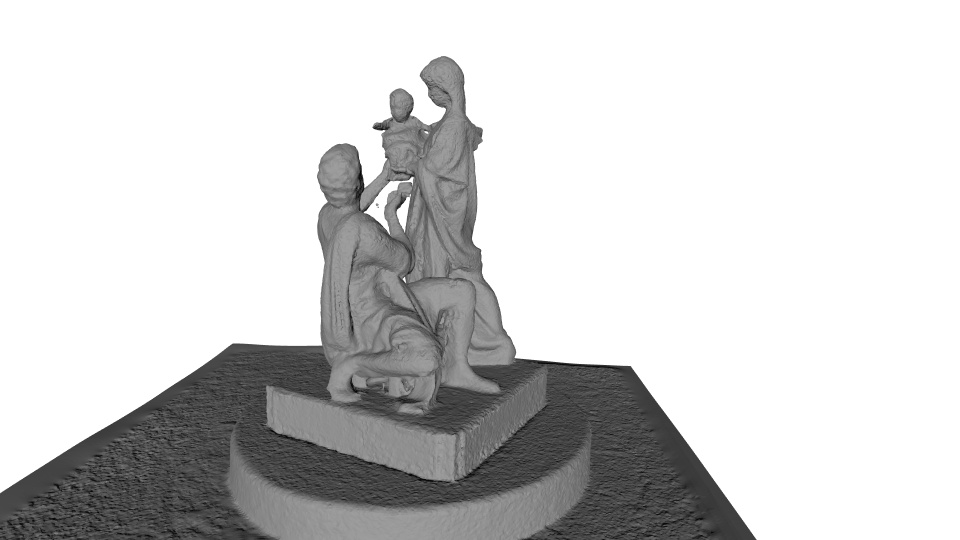} & 
		\includegraphics[width=0.3\linewidth,trim={150 0 300 0},clip]{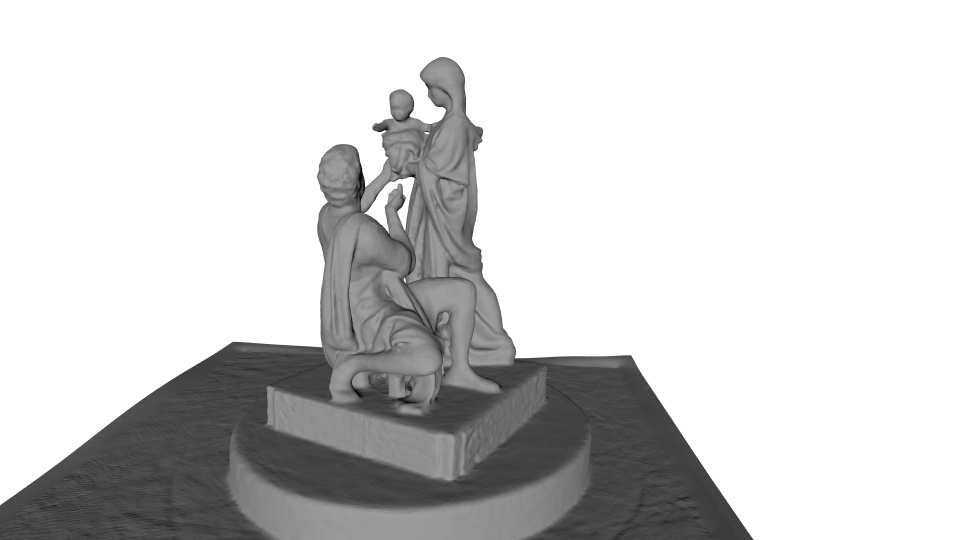} & 
		\includegraphics[width=0.3\linewidth,trim={150 0 300 0},clip]{images_tnt_idrx_mesh_Family_00000022_render.jpg} \\
		
		\includegraphics[width=0.3\linewidth,trim={150 0 300 0},clip]{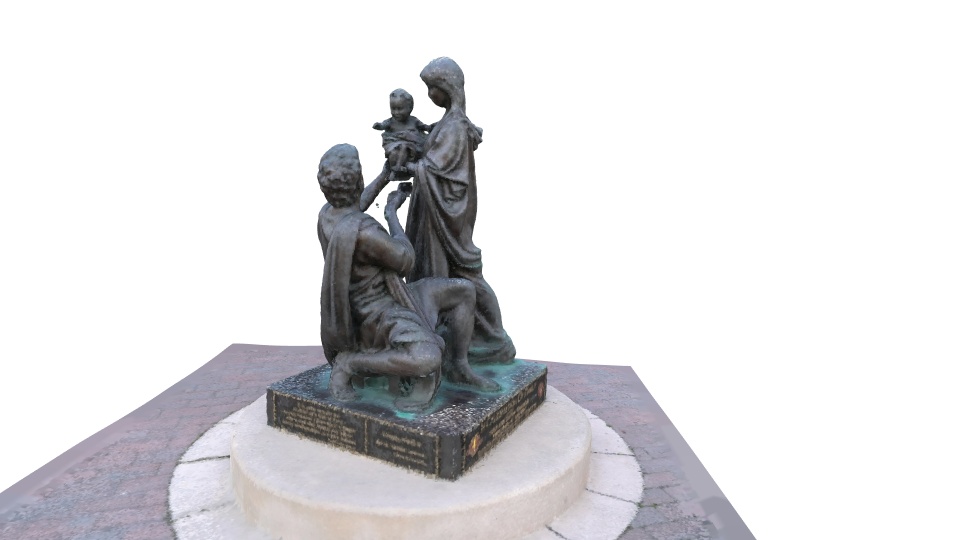} &
		\includegraphics[width=0.3\linewidth,trim={150 0 300 0},clip]{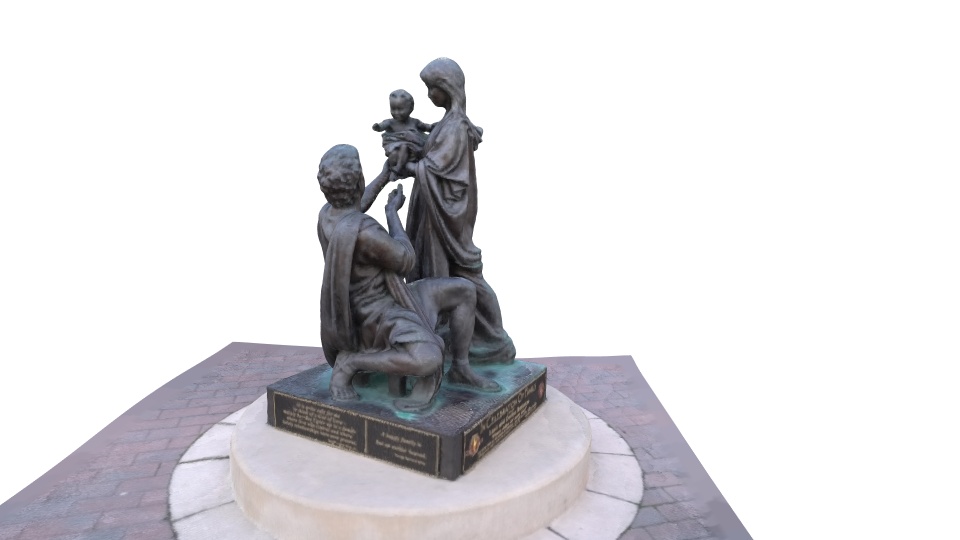} &
		\includegraphics[width=0.3\linewidth,trim={100 0 200 0},clip]{images_tnt_idrx_Family_eval_022.jpg} \\

		\includegraphics[width=0.3\linewidth,trim={180 0 210 0},clip]{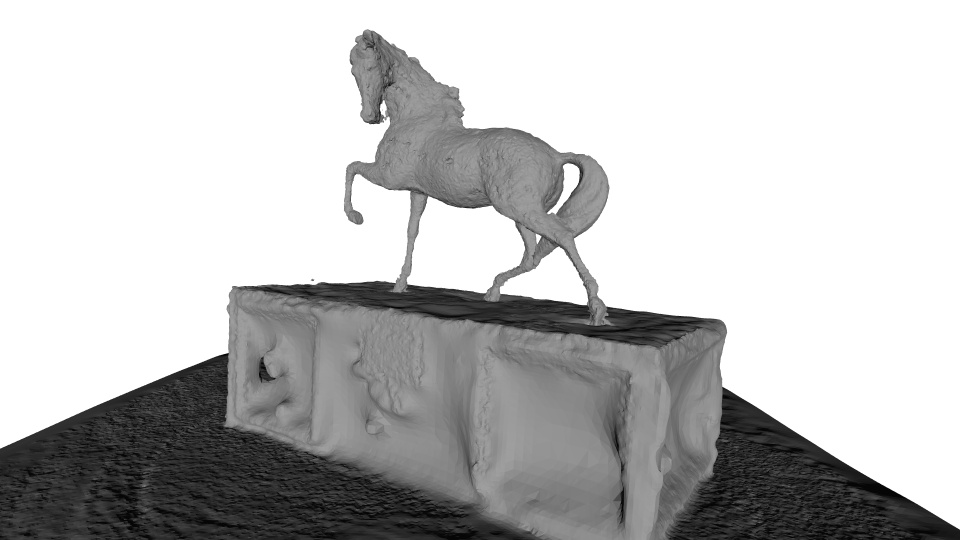} & 
		\includegraphics[width=0.3\linewidth,trim={180 0 210 0},clip]{images_tnt_vismvsnet_mesh_Horse_00000063_render.jpg} & 
		\includegraphics[width=0.3\linewidth,trim={180 0 210 0},clip]{images_tnt_idrx_mesh_Horse_00000063_render.jpg} \\
		
		\includegraphics[width=0.3\linewidth,trim={180 0 210 0},clip]{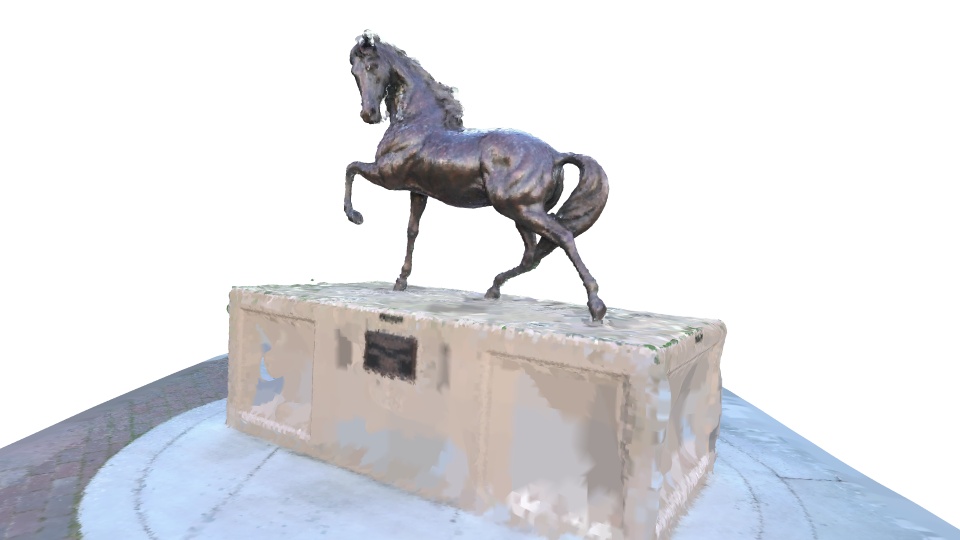} &
		\includegraphics[width=0.3\linewidth,trim={180 0 210 0},clip]{images_tnt_vismvsnet_Horse_00000063_render.jpg} &
		\includegraphics[width=0.3\linewidth,trim={120 0 140 0},clip]{images_tnt_idrx_Horse_eval_063.jpg} \\

		Colmap & Vis-MVSNet & MVSDF (Ours) 
	\end{tabular}
	\caption{Qualitative results on Tanks and Temples dataset.}
	\label{fig:tnt_supp}
\end{figure*}

Here we show additional qualitative results on \textit{DTU} (Fig.~\ref{fig:dtu1_supp},\ref{fig:dtu2_supp}), \textit{EPFL} (Fig.~\ref{fig:epfl_supp}) and \textit{Tanks and Temples} (Fig.~\ref{fig:tnt_supp}) datasets. 

\end{document}